%% file: arxiv.tex
\documentclass[10pt,twocolumn,letterpaper]{article}

\usepackage[pagenumbers]{cvpr} 

\usepackage{graphicx}
\usepackage{amsmath}
\usepackage{amssymb}
\usepackage{booktabs}

\usepackage{listings}
\usepackage{color}
\usepackage{subcaption}
\usepackage{mathtools}
\usepackage{paralist} 
\usepackage{colortbl}
\usepackage{nicefrac}
\usepackage{multirow}
\usepackage[table,dvipsnames]{xcolor}
\usepackage{makecell}
\usepackage[outline]{contour}
\usepackage{soul}
\usepackage{bbm}
\usepackage{appendix}
\usepackage{xfrac}
\usepackage[percent]{overpic}
\usepackage{xcolor}

\input{macros.tex}

\usepackage[pagebackref,breaklinks]{hyperref}

\usepackage[capitalize]{cleveref}
\crefname{section}{Sec.}{Secs.}
\Crefname{section}{Section}{Sections}
\Crefname{table}{Table}{Tables}
\crefname{table}{Tab.}{Tabs.}

\def\cvprPaperID{9422} 
\def\confName{CVPR}
\def\confYear{2022}

\begin{document}

\title{Mip-NeRF 360: Unbounded Anti-Aliased Neural Radiance Fields}

\author{
Jonathan T. Barron$^1$
\quad
Ben Mildenhall$^1$
\quad
Dor Verbin$^{1, 2}$
\\
Pratul P. Srinivasan$^1$
\quad
Peter Hedman$^1$
\\
\vspace{2mm}
{$^1$Google Research \quad\quad $^2$Harvard University}
}
%

\newcommand{\suppname}{appendix\,}

\maketitle

\input{main_content}

{\small
\bibliographystyle{ieee_fullname}
\bibliography{egbib}
}

\clearpage

\appendix

\input{supp_content}

\end{document}

%% file: macros.tex
\definecolor{yellow}{rgb}{1, 1, 0.7}
\definecolor{orange}{rgb}{1, 0.85, 0.7}
\definecolor{red}{rgb}{1, 0.7, 0.7}

\definecolor{lightyellow}{rgb}{1,1, 0.8}

\definecolor{wincolor}{rgb}{0.85, 0.0, 0.0}

\definecolor{darkyellow}{rgb}{0.8, 0.8, 0.5}
\definecolor{darkred}{rgb}{0.7, 0.3, 0.3}
\definecolor{darkgreen}{rgb}{0.3, 0.7, 0.3}
\definecolor{blue}{rgb}{0, 0, 1.0}
\definecolor{green}{rgb}{0, 1.0, 0}
\definecolor{pink}{rgb}{1, 0.4, 0.7}

\let\originalleft\left
\let\originalright\right
\renewcommand{\left}{\mathopen{}\mathclose\bgroup\originalleft}
\renewcommand{\right}{\aftergroup\egroup\originalright}

\newcommand{\norm}[1]{\left\lVert#1\right\rVert}

\newcommand{\distance}{t}
\newcommand{\sdistance}{s}

\newcommand{\tinterval}{T}

\newcommand{\transpose}{{\operatorname{T}}}

\newcommand{\ray}{\mathbf{r}}

\newcommand{\col}{\mathbf{c}}
\newcommand{\Col}{\mathbf{C}}
\newcommand{\trueCol}{\mathbf{C}^*}
\newcommand{\mlp}{\operatorname{MLP}}
\newcommand{\modelweights}{\Theta}

\newcommand{\basis}{\mathbf{P}}

\newcommand{\density}{\tau}

\newcommand{\zval}{t}
\newcommand{\zvec}{\mathbf{\zval}}

\newcommand{\proposal}[1]{\hat{#1}}

\newcommand{\covmat}{\boldsymbol{\Sigma}}

\newcommand{\distloss}{\mathcal{L}_{\mathrm{dist}}}
\newcommand{\proploss}{\mathcal{L}_{\mathrm{prop}}}
\newcommand{\reconloss}{\mathcal{L}_{\mathrm{recon}}}

%% file: main_content.tex
\begin{abstract}

Though neural radiance fields (NeRF) have demonstrated impressive view synthesis results on objects and small bounded regions of space, they struggle on ``unbounded'' scenes, where the camera may point in any direction and content may exist at any distance. In this setting, existing NeRF-like models often produce blurry or low-resolution renderings (due to the unbalanced detail and scale of nearby and distant objects), are slow to train, and may exhibit artifacts due to the inherent ambiguity of the task of reconstructing a large scene from a small set of images. We present an extension of mip-NeRF (a NeRF variant that addresses sampling and aliasing) that uses a non-linear scene parameterization, online distillation, and a novel distortion-based regularizer to overcome the challenges presented by unbounded scenes. Our model, which we dub ``mip-NeRF 360'' as we target scenes in which the camera rotates 360 degrees around a point, reduces mean-squared error by 57\% compared to mip-NeRF, and is able to produce realistic synthesized views and detailed depth maps for highly intricate, unbounded real-world scenes.

\end{abstract}


\begin{figure}[t]
    \centering
    \begin{tabular}{@{}c@{\,\,}c@{}}
        \includegraphics[width=0.49\linewidth]{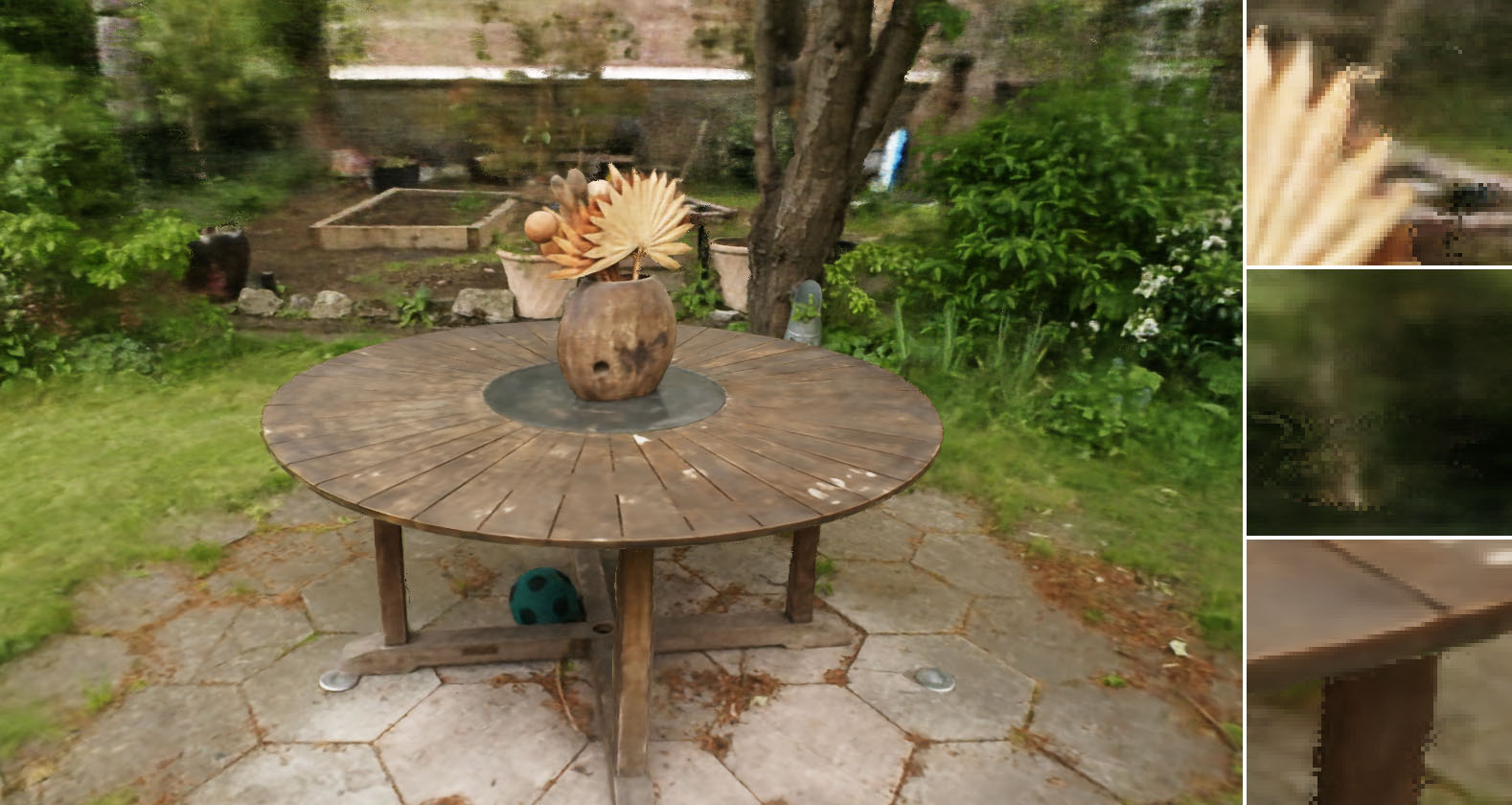} &
        \includegraphics[width=0.49\linewidth]{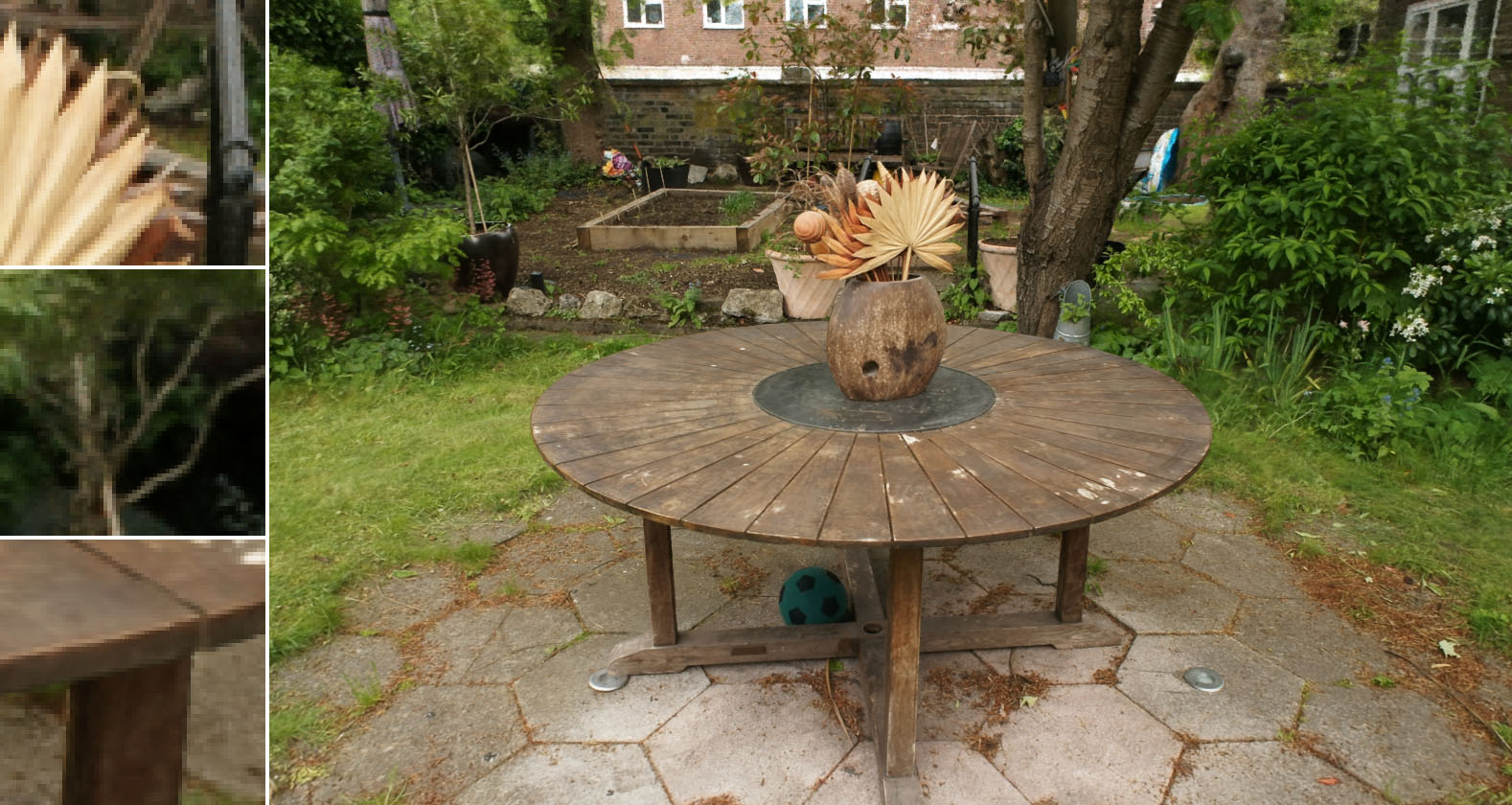} \\
        \includegraphics[width=0.49\linewidth]{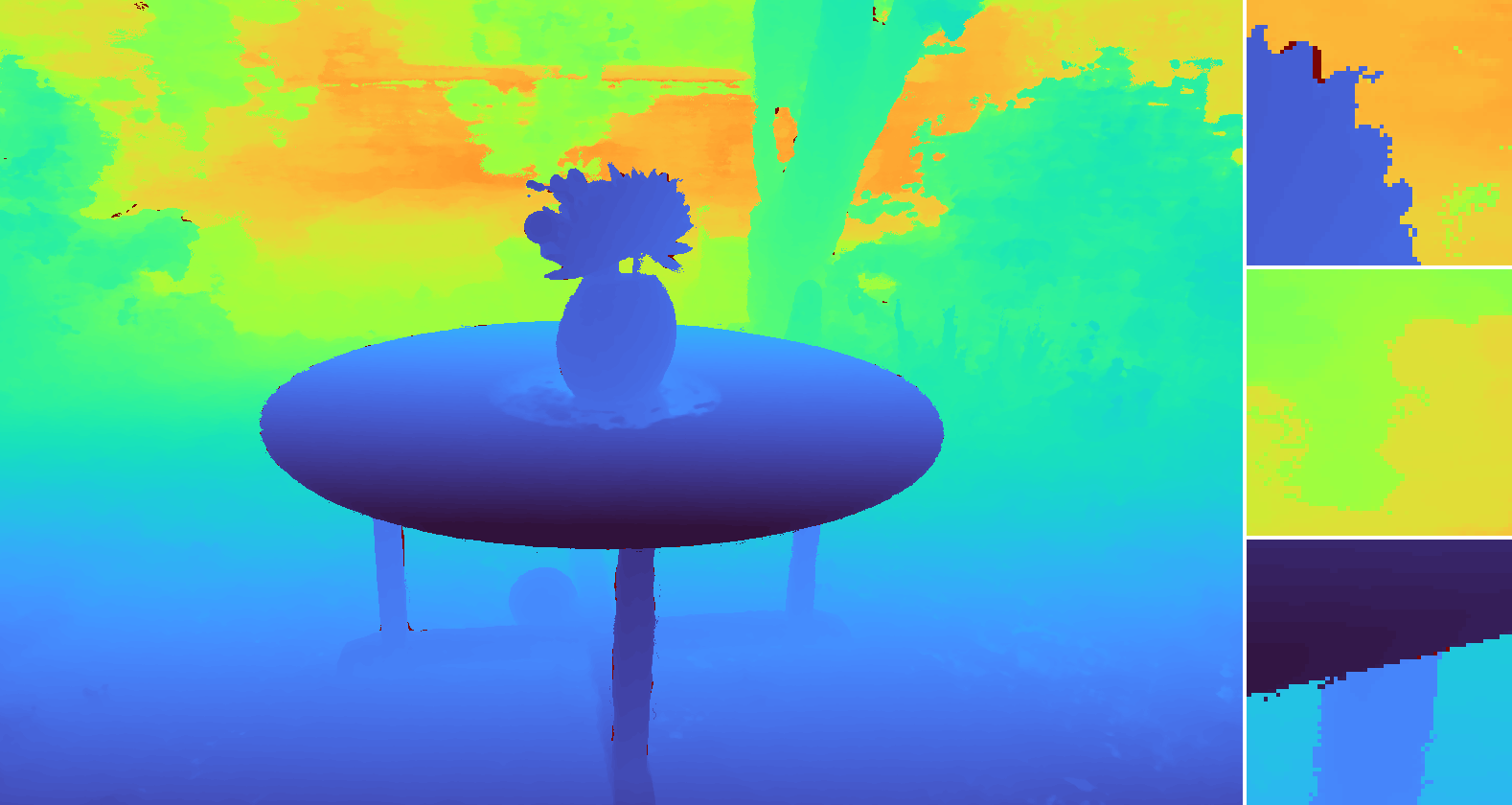} &
        \includegraphics[width=0.49\linewidth]{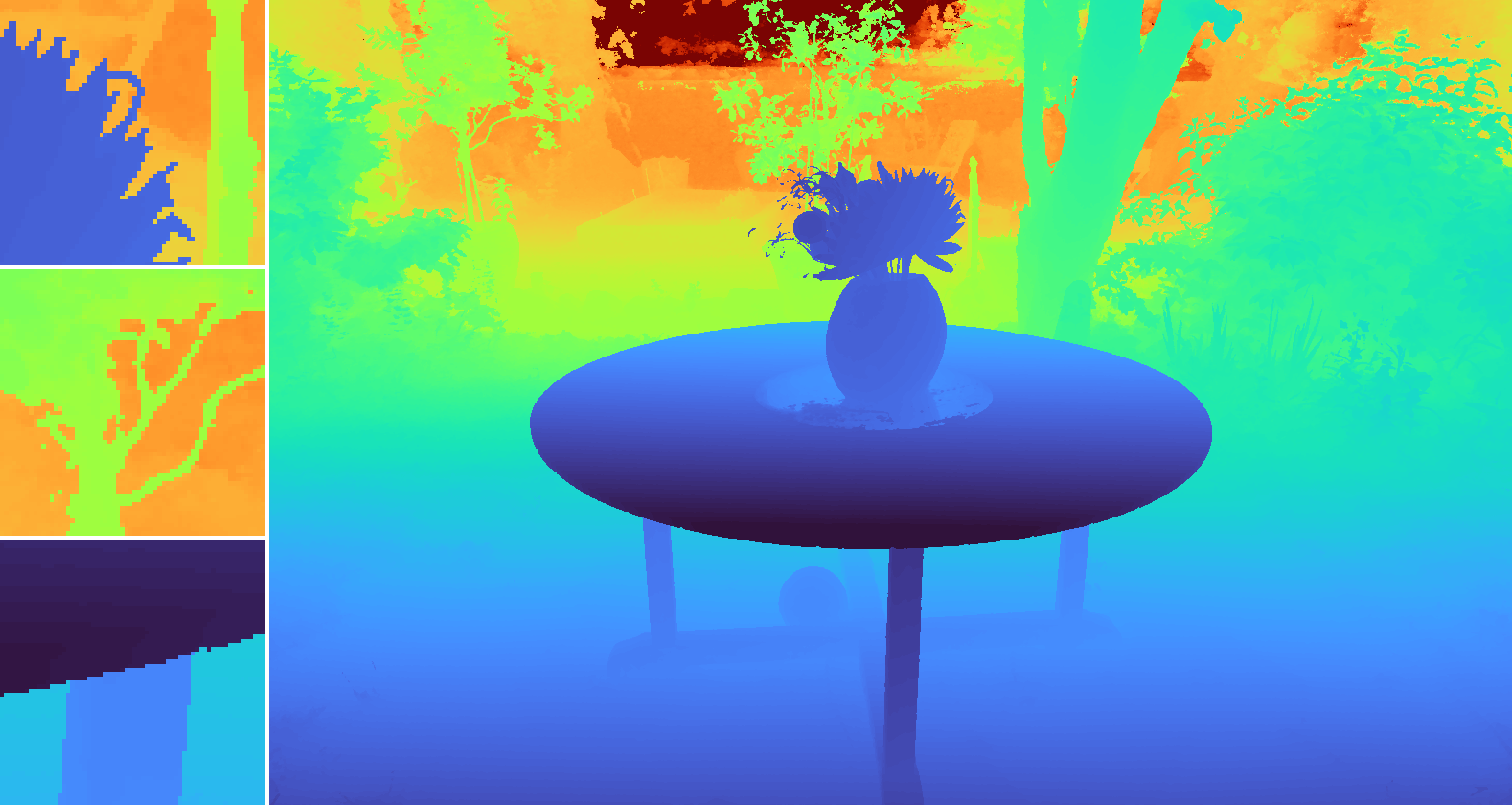} \\
        \scriptsize (a) mip-NeRF~\cite{barron2021mipnerf}, SSIM=0.526 & \scriptsize (b) Our Model, SSIM=0.804
    \end{tabular}
    \caption{
    (a) Though mip-NeRF is able to produce accurate renderings of objects, for unbounded scenes it often generates blurry backgrounds and low-detail foregrounds. (b) Our model produces detailed realistic renderings of these unbounded scenes, as evidenced by the renderings (top) and depth maps (bottom) from both models.
    See the supplemental video for additional results.
    }
    \label{fig:teaser}
\end{figure}

Neural Radiance Fields (NeRF) synthesize highly realistic renderings of scenes by encoding the volumetric density and color of a scene within the weights of a coordinate-based multi-layer perceptron (MLP). This approach has enabled significant progress towards photorealistic view synthesis~\cite{mildenhall2020}. However, NeRF models the input to the MLP using infinitesimally small 3D points along a ray, which causes aliasing when rendering views of varying resolutions. Mip-NeRF rectified this problem by extending NeRF to instead reason about volumetric frustums along a cone~\cite{barron2021mipnerf}. Though this improves quality, both NeRF and mip-NeRF struggle when dealing with \emph{unbounded} scenes, where the camera may face any direction and scene content may exist at any distance. In this work, we present an extension to mip-NeRF we call ``mip-NeRF 360'' that is capable of producing realistic renderings of these unbounded scenes, as shown in Figure~\ref{fig:teaser}.
 
Applying NeRF-like models to large unbounded scenes raises three critical issues:
\begin{compactenum}
    \item Parameterization. Unbounded 360 degree scenes can occupy an arbitrarily large region of Euclidean space, but mip-NeRF requires that 3D scene coordinates lie in a bounded domain.
    \item Efficiency. Large and detailed scenes require more network capacity, but densely querying a large MLP along each ray during training is expensive.
    \item Ambiguity. The content of unbounded scenes may lie at any distance and will be observed by only a small number of rays, exacerbating the inherent ambiguity of reconstructing 3D content from 2D images.
\end{compactenum}

\paragraph{Parameterization.} Due to perspective projection, an object placed far from the camera will occupy a small portion of the image plane, but will occupy more of the image and be visible in detail if placed nearby.
Therefore, an ideal parameterization of a 3D scene should allocate more capacity to nearby content and less capacity to distant content.
Outside of NeRF,
traditional view-synthesis methods address this by parameterizing the scene in projective panoramic space~\cite{shum1999concentric, zheng2007layered, hedman2017casual3d, overbeck2018system, broxton2020immersive, lin2020deep, attal2020, OmniPhotos, Khademi2021} or by embedding scene content within some proxy geometry~\cite{riegler2021svs, hedman2018deep, kopanas2021point} that has been recovered using multi-view stereo.

One aspect of NeRF's success is its pairing of specific scene types with their appropriate 3D parameterizations. The original NeRF paper~\cite{mildenhall2020} focused on 360 degree captures of objects with masked backgrounds and on front-facing scenes where all images face roughly the same direction.
For masked objects NeRF directly parameterized the scene in 3D Euclidean space, but for front-facing scenes NeRF used coordinates defined in projective space (normalized device coordinates, or ``NDC''~\cite{Blinn1991}).
By warping an infinitely deep camera frustum into a bounded cube where distance along the $z$-axis corresponds to disparity (inverse distance), NDC effectively reallocates the NeRF MLP's capacity in a way that is consistent with the geometry of perspective projection.

However, scenes that are unbounded in \emph{all} directions, not just in a single direction, require a different parameterization.
This idea was explored by NeRF++~\cite{kaizhang2020}, which used an additional network to model distant objects, and by DONeRF~\cite{donerf}, which proposed a space-warping procedure to shrink distant points towards the origin.
Both of these approaches behave somewhat analogously to NDC but in \emph{every} direction, rather than just along the $z$-axis. 
In this work, we extend this idea to mip-NeRF 
and present a method for applying any smooth parameterization to \emph{volumes} (rather than points), and also present our own parameterization for unbounded scenes.

\paragraph{Efficiency.} One fundamental challenge in dealing with unbounded scenes is that such scenes are often \emph{large} and \emph{detailed}. Though NeRF-like models can accurately reproduce objects or regions of scenes using a surprisingly small number of weights, the capacity of the NeRF MLP saturates when faced with increasingly intricate scene content. Additionally, larger scenes require significantly more samples along each ray to accurately localize surfaces.
For example, when scaling NeRF from objects to buildings, Martin-Brualla~\etal~\cite{martinbrualla2020nerfw} doubled the number of MLP hidden units and increased the number of MLP evaluations by 8$\times$. This increase in model capacity is expensive --- a NeRF already takes multiple hours to train, and multiplying this time by an additional $\sim$40$\times$ is prohibitively slow for most uses.

This training cost is exacerbated by the coarse-to-fine resampling strategy used by NeRF and mip-NeRF: MLPs are evaluated multiple times using ``coarse'' and ``fine'' ray intervals, and are supervised using an image reconstruction loss on both passes. This approach is wasteful, as the ``coarse'' rendering of the scene does not contribute to the final image.
Instead of training a single NeRF MLP that is supervised at multiple scales, we will instead train two MLPs: a ``proposal MLP'' and a ``NeRF MLP''. The proposal MLP predicts volumetric density (but not color) and those densities are used to resample new intervals that are provided to the NeRF MLP, which then renders the image. Crucially, the weights produced by the proposal MLP are not supervised using the input image, but are instead supervised with the histogram weights generated by the NeRF MLP.
This allows us to use a large NeRF MLP that is evaluated relatively few times, alongside a small proposal MLP that is evaluated many more times.
As a result, our whole model's total capacity is significantly larger than mip-NeRF's ($\sim$15$\times$), resulting in greatly improved rendering quality, but our training time only increases modestly ($\sim$2$\times$).

We can think of this approach as a kind of ``online distillation'': while ``distillation'' commonly refers to training a small network to match the output of an already-trained large network~\cite{hinton2015distilling}, here we distill the structure of the outputs predicted by the NeRF MLP into the proposal MLP ``online'' by training both networks simultaneously.
NeRV~\cite{nerv2021} performs a similar kind of online distillation for an entirely different task: training MLPs to approximate rendering integrals for the purpose of modeling visibility and indirect illumination.
Our online distillation approach is similar in spirit to the ``sampling oracle networks'' used in DONeRF, though that approach uses ground-truth depth for supervision~\cite{donerf}. A related idea was used in TermiNeRF~\cite{piala2021terminerf}, though that approach only accelerates inference and actually \emph{slows} training (a NeRF is trained to convergence, and an additional model is trained afterwards). A learned ``proposer'' network was explored in NeRF in Detail~\cite{arandjelovic2021nerf} but only achieves a speedup of 25\%, while our approach accelerates training by 300\%. 

Several works have attempted to distill or ``bake'' a trained NeRF into a format that can be \emph{rendered} quickly\cite{hedman2021snerg, reiser2021kilonerf, yu2021plenoctrees}, but these techniques do not accelerate training. The idea of accelerating ray-tracing through a hierarchical data structure such as octrees~\cite{samet1990design} or bounding volume hierarchies~\cite{rubin1980bvh} is well-explored in the rendering literature, though these approaches assume a-priori knowledge of the geometry of the scene and therefore do not naturally generalize to an inverse rendering context in which the geometry of the scene is unknown and must be recovered. Indeed, despite building an octree acceleration structure while optimizing a NeRF-like model, the Neural Sparse Voxel Fields approach does not significantly reduce training time~\cite{liu2020nsvf}.

\paragraph{Ambiguity.} Though NeRFs are traditionally optimized using many input images of a scene, the problem of recovering a NeRF that produces realistic synthesized views from novel camera angles is still fundamentally underconstrained --- an infinite family of NeRFs can explain away the input images, but only a small subset produces acceptable results for novel views.
For example, a NeRF could recreate all input images by simply reconstructing each image as a textured plane immediately in front of its respective camera.
The original NeRF paper regularized ambiguous scenes by injecting Gaussian noise into the density head of the NeRF MLP before the rectifier~\cite{mildenhall2020}, which encourages densities to gravitate towards either zero or infinity. Though this reduces some ``floaters'' by discouraging semi-transparent densities, we will show that it is insufficient for our more challenging task. Other regularizers for NeRF have been proposed, such as a robust loss on density~\cite{hedman2021snerg} or smoothness penalties on surfaces~\cite{nerfactor, Oechsle2021ICCV}, but these solutions address different problems than ours (slow rendering and non-smooth surfaces, respectively). 
Additionally, these regularizers are designed for the point samples used by NeRF, while our approach is designed to work with the continuous weights defined along each mip-NeRF ray.

These three issues will be addressed in Sections~\ref{sec:param}, \ref{sec:distil}, and \ref{sec:regularize} respectively, after a review of mip-NeRF. We will demonstrate our improvement over prior work using a new dataset consisting of challenging indoor and outdoor scenes. We urge the reader to view our supplemental video, as our results are best appreciated when animated.

\section{Preliminaries: mip-NeRF}
\label{sec:prelim}

Let us first describe how a fully-trained mip-NeRF~\cite{barron2021mipnerf} renders the color of a single ray cast into the scene $\mathbf{r}(t) = \mathbf{o} + t \mathbf{d}$, where $\mathbf{o}$ and $\mathbf{d}$ are the origin and direction of the ray respectively, and $t$ denotes distance along the ray. In mip-NeRF, a sorted vector of distances $\mathbf{t}$ is defined and the ray is split into a set of intervals $\tinterval_i = [\distance_i, \distance_{i+1})$. For each interval $i$ we compute the mean and covariance $(\boldsymbol{\mu}, \covmat) = \mathbf{r}(T_i)$ of the conical frustum corresponding to the interval (the radii of which are determined by the ray's focal length and pixel size on the image plane), and featurize those values using an integrated positional encoding:
\begin{equation}
\resizebox{2.95in}{!}{$
\gamma(\boldsymbol{\mu}, \covmat) 
= \Bigg\{ \begin{bmatrix} \sin(2^\ell \boldsymbol{\mu}) \exp\left(-2^{2\ell - 1} \operatorname{diag}\left(\covmat\right)\right)
\\ \cos(2^\ell \boldsymbol{\mu}) \exp\left(-2^{2\ell - 1} \operatorname{diag}\left(\covmat\right)\right)\end{bmatrix} \Bigg\}_{\ell=0}^{L-1}\label{eq:ipe}
$}
\end{equation}
This is the expectation of the encodings used by NeRF with respect to a Gaussian approximating the conical frustum. These features are used as input to an MLP parameterized by weights $\modelweights_{\mathrm{NeRF}}$ that outputs a density $\density$ and color $\col$:
\begin{equation}
    \forall T_i \in \mathbf{t}, \quad (\density_i, \, \col_i) = \mlp\left( \gamma\left(\ray(T_i)\right); \, \modelweights_{\mathrm{NeRF}} \right) \, .\label{eq:nerf_mlp}
\end{equation}
The view direction $\mathbf{d}$ is also provided as input to the MLP, but we omit this for simplicity.
With these densities and colors we approximate the volume rendering integral using numerical quadrature\cite{max1995optical}:
\begin{gather}
\Col(\ray, \zvec) = \sum_{i} w_i \col_i \,, \quad\quad\quad\quad\quad \label{eq:nerf_render} \\
w_i = \left(1-e^{-\density_i (t_{i+1} - t_i)} \right) e^{-\sum_{i' < i} \density_{i'} \left(t_{i'+1} - t_{i'} \right)} \label{eq:nerf_weights}
\end{gather}
where $\Col(\ray, \zvec)$ is the final rendered pixel color. By construction, the alpha compositing weights $\mathbf{w}$ are guaranteed to sum to less than or equal to 1.

The ray is first rendered using evenly-spaced ``coarse'' distances $\mathbf{t}^c$, which are sorted samples from a uniform distribution spanning $[t_n, t_f]$, the camera's near and far planes:
\begin{equation}
    t^c \sim \mathcal{U}[t_n, t_f]\,, \quad \mathbf{t}^c = \operatorname{sort}( \{ t^c \} )\,. \label{eq:samplecoarse}
\end{equation}
During training this sampling is stochastic, but during evaluation samples are evenly spaced from $t_n$ to $t_f$.
After the MLP generates a vector of ``coarse'' weights $\mathbf{w}^c$, 
``fine'' distances $\mathbf{t}^f$ are sampled from the histogram defined by $\mathbf{t}^c$ and $\mathbf{w}^c$ using inverse transform sampling:
\begin{equation}
    t^f \sim \operatorname{hist}(\mathbf{t}^c, \mathbf{w}^c)\,, \quad \mathbf{t}^f = \operatorname{sort}( \{ t^f \} )\,. \label{eq:samplefine}
\end{equation}
Because the coarse weights $\mathbf{w}^c$ tend to concentrate around scene content, this strategy improves sampling efficiency.

A mip-NeRF is recovered by optimizing MLP parameters $\modelweights_{\mathrm{NeRF}}$ via gradient descent to minimize a weighted combination of coarse and fine reconstruction losses:
\begin{equation}
\resizebox{2.9in}{!}{$
\displaystyle \sum_{\ray \in \mathcal{R}} \frac{1}{10} \reconloss(\Col(\ray, \zvec^{c}), \trueCol(\ray)) + \reconloss(\Col(\ray, \zvec^{f}), \trueCol(\ray))
$}
\end{equation}
where $\mathcal{R}$ is the set of rays in our training data, $\trueCol(\ray)$ is the ground truth color corresponding to ray $\ray$ taken from an input image, and $\reconloss$ is mean squared error.

\section{Scene and Ray Parameterization}
\label{sec:param}

\begin{figure}[t]
    \centering
    \includegraphics[width=\linewidth]{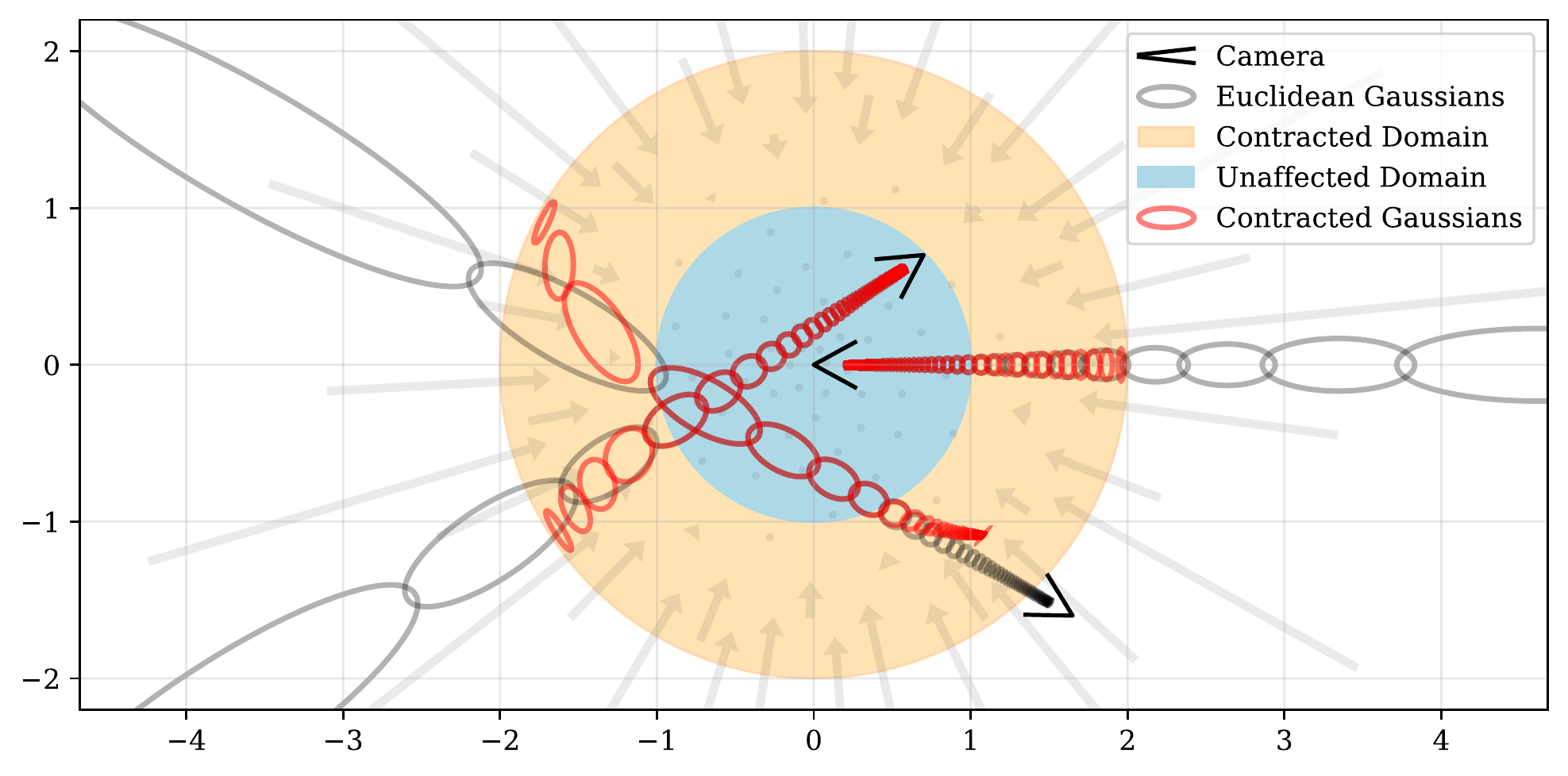}
    \vspace{-0.15in}
    \caption{
    A 2D visualization of our scene parameterization. We define a $\operatorname{contract}(\cdot)$ operator (Equation~\ref{eq:contract}, shown as arrows) that maps coordinates onto a ball of radius 2 (orange), where points within a radius of 1 (blue) are unaffected. 
    We apply this contraction to mip-NeRF Gaussians in Euclidean 3D space (gray ellipses) similarly to a Kalman filter to produce our contracted Gaussians (red ellipses), whose centers are guaranteed to lie within a ball of radius 2.
    The design of $\operatorname{contract}(\cdot)$ combined with our choice to space ray intervals linearly according to disparity means that rays cast from a camera located at the origin of the scene will have equidistant intervals in the orange region, as demonstrated here.
    }
    \label{fig:contraction}
\end{figure}

Though there exists prior work on the parameterization of \emph{points} for unbounded scenes, this does not provide a solution for the mip-NeRF context, in which we must re-parameterize \emph{Gaussians}. To do this, first let us define $f(\mathbf{x})$ as some smooth coordinate transformation that maps from $\mathbb{R}^n \rightarrow \mathbb{R}^n$ (in our case, $n=3$). We can compute the linear approximation of this function:
\begin{equation}
     f(\mathbf{x}) \approx f(\mathbf{\boldsymbol{\mu}}) + \mathbf{J}_{f}(\mathbf{\boldsymbol{\mu}})(\mathbf{x} - \mathbf{\boldsymbol{\mu}})
\end{equation}
Where $\mathbf{J}_{f}(\mathbf{\boldsymbol{\mu}})$ is the Jacobian of $f$ at $\boldsymbol{\mu}$. With this, we can apply $f$ to $(\boldsymbol{\mu}, \covmat)$ as follows:
\begin{equation}
    f(\boldsymbol{\mu}, \covmat) = \left( f(\boldsymbol{\mu}), \, \mathbf{J}_{f}(\mathbf{\boldsymbol{\mu}})  \covmat \mathbf{J}_{f}(\mathbf{\boldsymbol{\mu}})^\mathrm{T} \right)
\end{equation}
This is functionally equivalent to the classic Extended Kalman filter~\cite{kalman1960new}, where $f$ is the state transition model.
Our choice for $f$ is the following contraction:
\begin{equation}
\operatorname{contract}(\mathbf{x})  = \begin{cases}
\mathbf{x} & \norm{\mathbf{x}} \leq 1\\
\left(2  - \frac{1}{\norm{\mathbf{x}}}\right)\left(\frac{\mathbf{x}}{\norm{\mathbf{x}}}\right) & \norm{\mathbf{x}} > 1 \label{eq:contract}
\end{cases}
\end{equation}
This design shares the same motivation as NDC: distant points should be distributed proportionally to disparity (inverse distance) rather than distance. In our model, instead of using mip-NeRF's IPE features in Euclidean space as per Equation~\ref{eq:ipe} we use similar features (see \suppname\!) in this contracted space: $\gamma( \operatorname{contract}(\boldsymbol{\mu}, \covmat))$. See Figure~\ref{fig:contraction} for a visualization of this parameterization. 

In addition to the question of how 3D coordinates should be parameterized, there is the question of how ray distances $\mathbf{t}$ should be selected. In NeRF this is usually done by sampling uniformly from the near and far plane as per Equation~\ref{eq:samplecoarse}. However, if an NDC parameterization is used, this uniformly-spaced series of samples is actually uniformly spaced in \emph{inverse} depth (disparity). This design decision is well-suited to unbounded scenes when the camera faces in only one direction, but is not applicable to scenes that are unbounded in all directions. We will therefore explicitly sample our distances $\mathbf{t}$ linearly in disparity (see \cite{mildenhall2019llff} for a detailed motivation of this spacing).

To parameterize a ray in terms of disparity we define an invertible mapping between Euclidean ray distance $t$ and a ``normalized'' ray distance $s$:
\begin{equation}
\resizebox{2.9in}{!}{$
\displaystyle s \triangleq \frac{g(\distance) - g(t_n)}{g(t_f) - g(t_n)}\,, \,\,\, \distance \triangleq g^{-1}\left(s \cdot g(t_f) + (1 - s) \cdot g(t_n) \right)\,,
$}
\end{equation}
where $g(\cdot)$ is some invertible scalar function. This gives us ``normalized'' ray distances $s \in [0, 1]$ that map to $[t_n, t_f]$. Throughout this paper we will refer to distances along a ray in either $t$-space or $s$-space, depending on which is more convenient or intuitive. 
By setting $g(x) = 1/x$ and constructing ray samples that are uniformly distributed in $s$-space, we produce ray samples whose $t$-distances are distributed linearly in disparity (additionally, setting $g(x) = \log(x)$ yields DONeRF's logarithmic spacing~\cite{donerf}). In our model, instead of performing the sampling in Equations~\ref{eq:samplecoarse} and \ref{eq:samplefine} using $t$ distances, we do so with $s$ distances. This means that, not only are our initial samples spaced linearly in disparity, but subsequent resamplings from individual intervals of the weights $\mathbf{w}$ will also be distributed similarly. As can be seen from the camera in the center of Figure~\ref{fig:contraction}, this linear-in-disparity spacing of ray samples counter-balances $\operatorname{contract}(\cdot)$. Effectively, we have co-designed our scene coordinate space with our inverse-depth spacing, which gives us a parameterization of unbounded scenes that closely resembles the highly-effective setting of the original NeRF paper: evenly-spaced ray intervals within a bounded space.

\begin{figure}[b!]
    \centering
    \includegraphics[width=\linewidth]{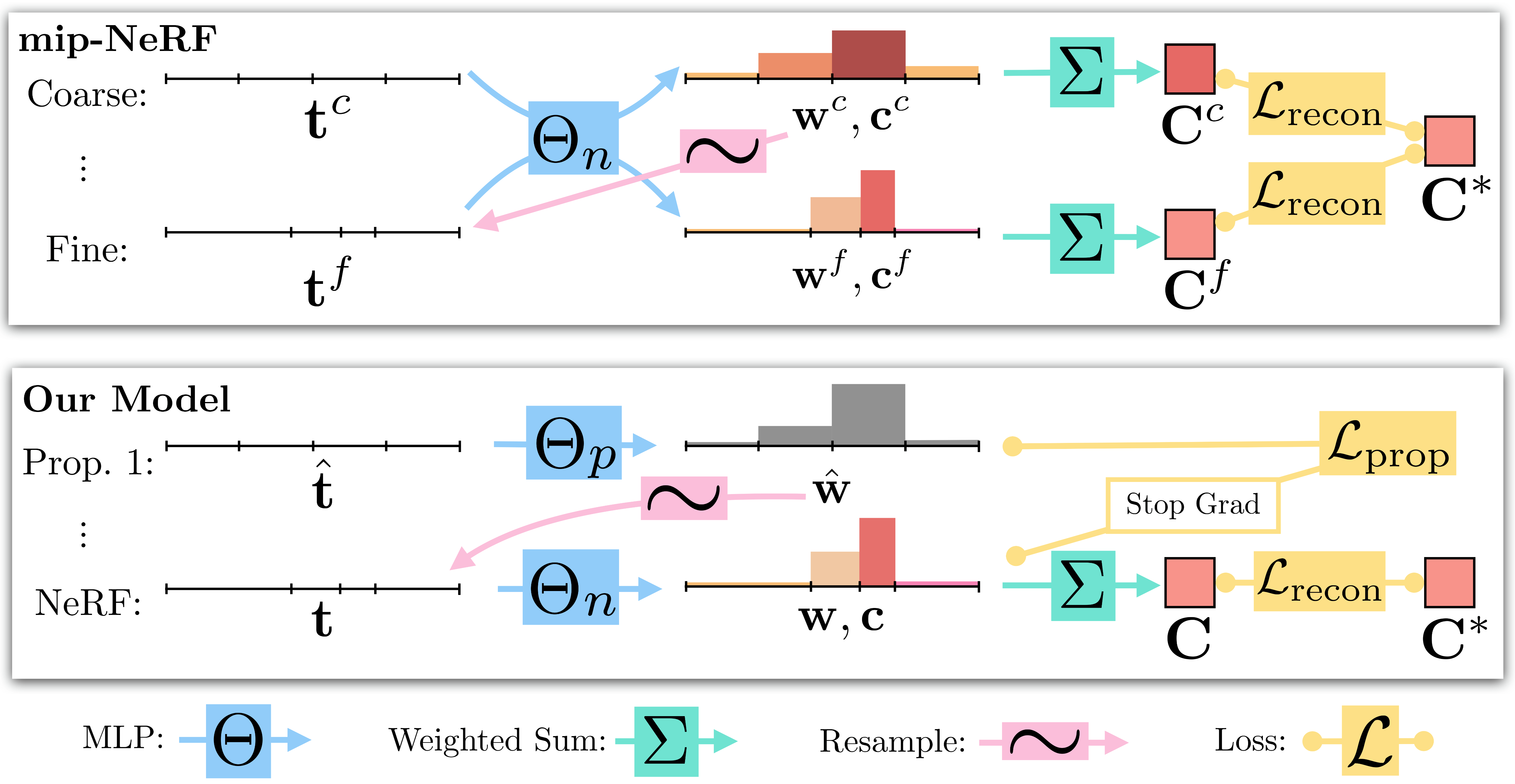}
    \vspace{-0.2in}
    \caption{
    A comparison of our model's architecture with mip-NeRF's.
    Mip-NeRF uses one multi-scale MLP that is repeatedly queried (only two repetitions shown here) for weights that are resampled into intervals for the next stage, and supervises the renderings produced at all scales.
    We use a ``proposal MLP'' that emits weights (but not color) that are resampled, and in the final stage we use a ``NeRF MLP'' to produce weights and colors that result in the rendered image, which we supervise. The proposal MLP is trained to produce proposal weights $\hat{\mathbf{w}}$ that are consistent with the NeRF MLP's $\mathbf{w}$ output.
    By using a small proposal MLP and a large NeRF MLP we obtain a combined model with a high capacity that is still tractable to train.
    }
    \label{fig:sketch}
\end{figure}

\section{Coarse-to-Fine Online Distillation}
\label{sec:distil}

As discussed, mip-NeRF uses a coarse-to-fine resampling strategy (Figure~\ref{fig:sketch}) in which the MLP is evaluated once using ``coarse'' ray intervals and again using ``fine'' ray intervals, and is supervised using an image reconstruction loss at both levels. We instead train two MLPs, a ``NeRF MLP'' $\modelweights_{\mathrm{NeRF}}$ (which behaves similarly to the MLPs used by NeRF and mip-NeRF) and a ``proposal MLP'' $\modelweights_{\mathrm{prop}}$. The proposal MLP predicts volumetric density, which is converted into a proposal weight vector $\hat{\mathbf{w}}$ according to Equation~\ref{eq:nerf_weights}, but does not predict color. 
These proposal weights $\hat{\mathbf{w}}$ are used to sample $s$-intervals that are then provided to the NeRF MLP, which predicts its own weight vector $\mathbf{w}$ (and color estimates, for use in rendering an image). Critically, the proposal MLP is not trained to reproduce the input image, but is instead trained to bound the weights $\mathbf{w}$ produced by the NeRF MLP. Both MLPs are initialized randomly and trained jointly, so this supervision can be thought of as a kind of ``online distillation'' of the NeRF MLP's knowledge into the proposal MLP.
We use a large NeRF MLP and a small proposal MLP, and repeatedly evaluate and resample from the proposal MLP with many samples (some figures and discussion illustrate only a single resampling for clarity) but evaluate the NeRF MLP only once with a smaller set of samples. This gives us a model that behaves as though it has a much higher capacity than mip-NeRF but is only moderately more expensive to train. As we will show, using a small MLP to model the proposal distribution does not reduce accuracy, which suggests that distilling the NeRF MLP is an easier task than view synthesis.

This online distillation requires a loss function that encourages the histograms emitted by the proposal MLP $(\proposal{\mathbf{\distance}}, \proposal{\mathbf{w}})$ and the NeRF MLP $(\mathbf{\distance}, \mathbf{w})$ to be consistent. At first this problem may seem trivial, as minimizing the dissimilarity between two histograms is a well-established task, but recall that the ``bins'' of those histograms $\mathbf{t}$ and $\hat{\mathbf{t}}$ need not be similar --- indeed, if the proposal MLP successfully culls the set of distances where scene content exists, $\proposal{\mathbf{\distance}}$ and $\mathbf{\distance}$ will be highly dissimilar. Though the literature contains numerous approaches for measuring the difference between two histograms with identical bins~\cite{dalal2005histograms, maji2008classification, pele2010quadratic}, our case is relatively underexplored. This problem is challenging because we cannot assume anything about the distribution of contents within one histogram bin: an interval with non-zero weight may indicate a uniform distribution of weight over that entire interval, a delta function located \emph{anywhere} in that interval, or myriad other distributions. We therefore construct our loss under the following assumption: If it is \emph{in any way possible} that both histograms can be explained using any single distribution of mass, then the loss must be zero. A non-zero loss can only be incurred if it is \emph{impossible} that both histograms are reflections of the same ``true'' continuous underlying distribution of mass. See the \suppname for visualizations of this concept.

\begin{figure}[b!]
    \centering
    \begin{tabular}{@{}c@{\,\,}ccc@{}}
        \rotatebox{90}{\!\!\!\!\!\!\! \small $w_i / (t_{i+1} - t_i)$} & \includegraphics[width=0.28\linewidth, trim=0.1in 0.1in 0.1in 0.1in, clip]{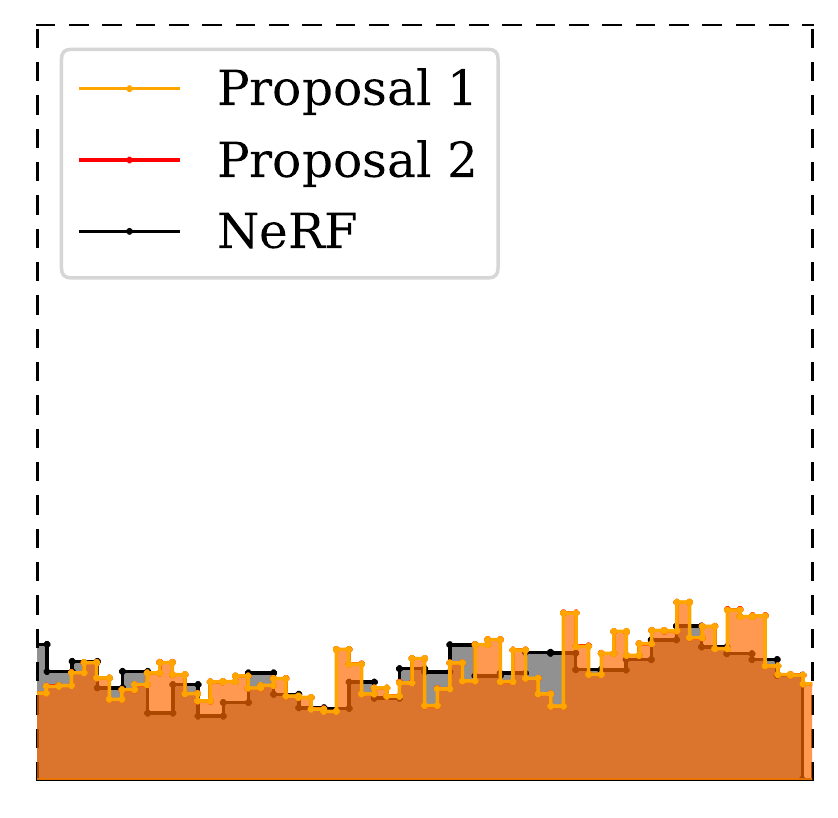} & 
        \includegraphics[width=0.28\linewidth, trim=0.1in 0.1in 0.1in 0.1in, clip]{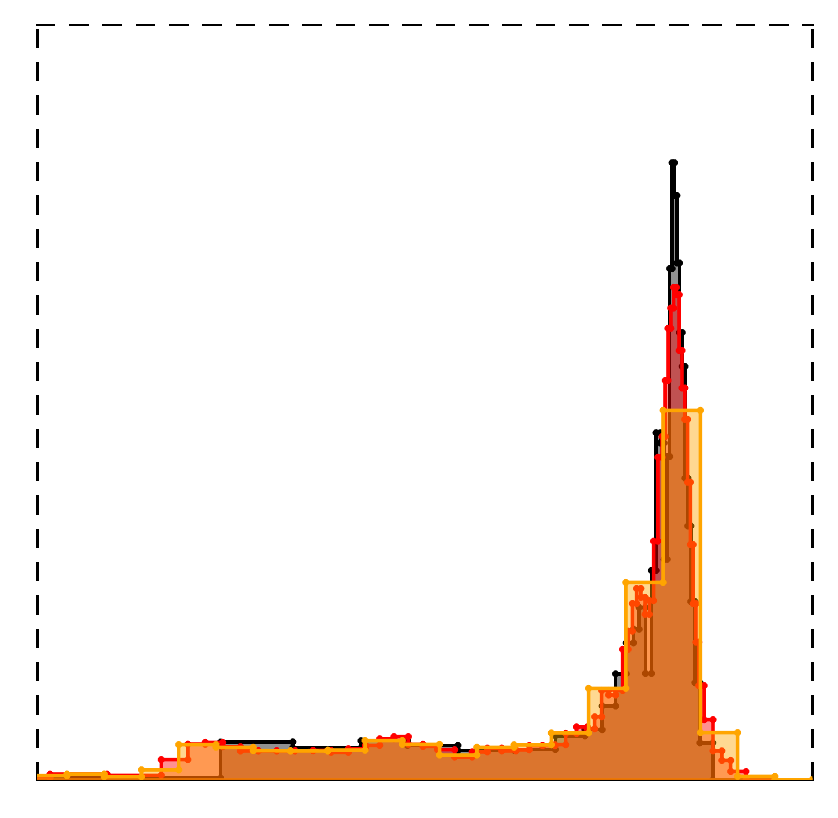} & 
        \includegraphics[width=0.28\linewidth, trim=0.1in 0.1in 0.1in 0.1in, clip]{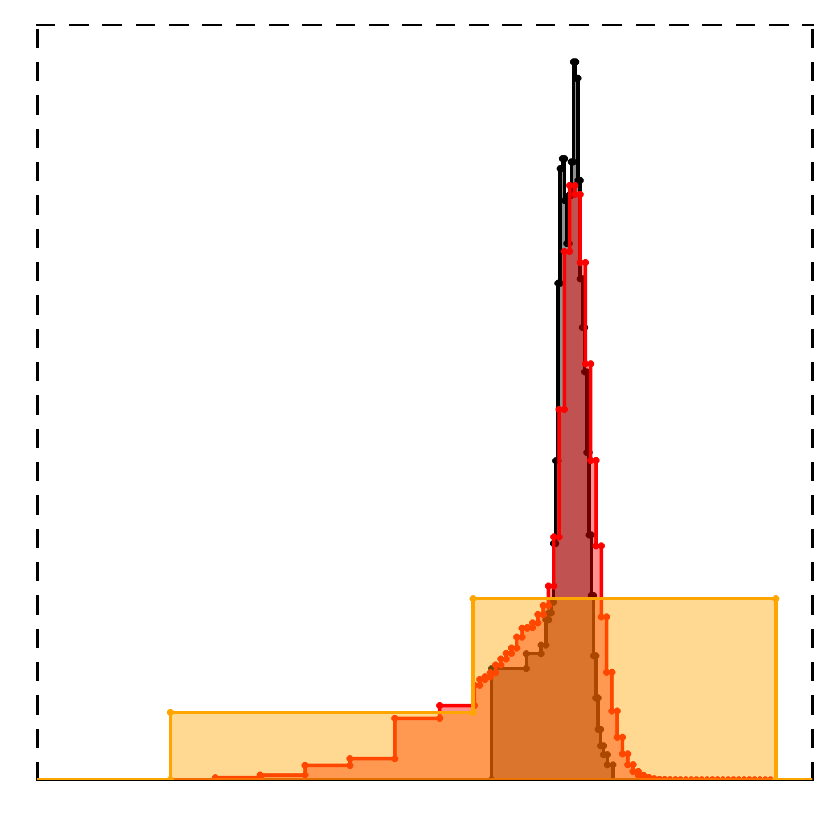} \\
        & \includegraphics[width=0.28\linewidth, trim=0.1in 0.1in 0.1in 0.1in, clip]{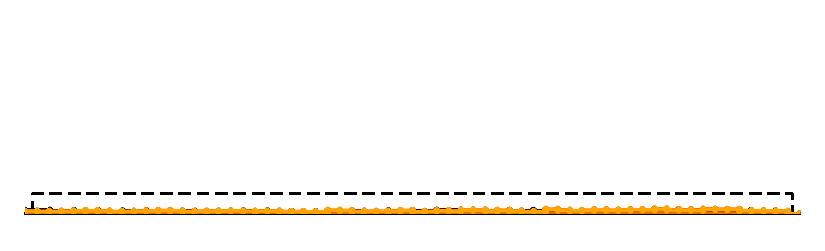} & 
        \includegraphics[width=0.28\linewidth, trim=0.1in 0.1in 0.1in 0.1in, clip]{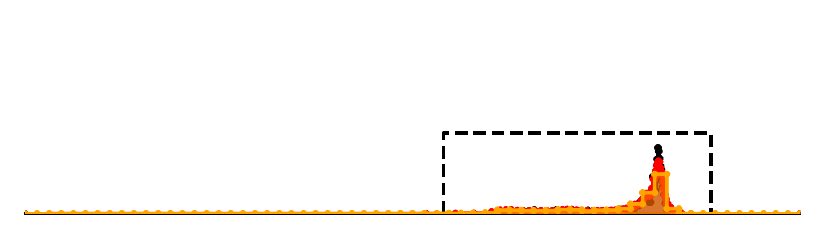} & 
        \includegraphics[width=0.28\linewidth, trim=0.1in 0.1in 0.1in 0.1in, clip]{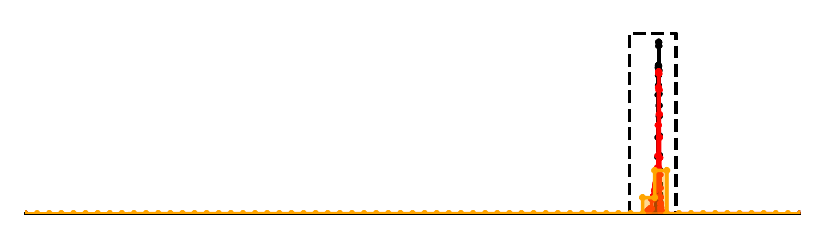} \\
        & \multicolumn{3}{c}{\small $s$} \\
        & \scriptsize (a) 0\% optimized & \scriptsize (b) 4\% optimized & \scriptsize (c) 100\% optimized
    \end{tabular}
    \caption{
    A visualization of the histograms $(\mathbf{t}, \mathbf{w})$ emitted from the NeRF MLP (black) and the two sets of histograms $(\hat{\mathbf{t}}, \hat{\mathbf{w}})$ emitted by the proposal MLP (yellow and orange) for a single ray from our dataset's \textit{bicycle} scene over the course of training.
    Below we visualize the entire ray with fixed x and y axes, but above we crop both axes to better visualize details near scene content. Histogram weights are plotted as distributions that integrate to $1$.
    (a) When training begins, all weights are uniformly distributed with respect to ray distance $t$. (b, c) As training progresses, the NeRF weights begin to concentrate around a surface and the proposal weights form a kind of envelope around those NeRF weights.}
    \label{fig:traces}
\end{figure}

To do this, we first define a function that computes the sum of all proposal weights that overlap with interval $T$:
\begin{gather}
\operatorname{bound}\left( \proposal{\mathbf{\distance}}, \proposal{\mathbf{w}}, T \right) = \sum_{j: \, \tinterval  \cap \hat{\tinterval}_j \neq \varnothing} \hat w_j\,. \label{eq:bound}
\end{gather}
If the two histograms are consistent with each other, then it must hold that $w_i \leq \operatorname{bound}\left( \proposal{\mathbf{\distance}}, \proposal{\mathbf{w}}, T_i \right)$  for all intervals $(T_i, w_i)$ in $(\mathbf{\distance}, \mathbf{w})$.
This property is similar to the additivity property of an outer measure in measure theory~\cite{evans2018measure}.
Our loss penalizes any surplus histogram mass that violates this inequality and exceeds this bound:
\begin{equation}
\resizebox{2.91in}{!}{$
\!\!\displaystyle \proploss\left(\mathbf{\distance}, \mathbf{w}, \proposal{\mathbf{\distance}}, \proposal{\mathbf{w}} \right)\!=\! \sum_{i}\frac{1}{w_{i}}\max\left( 0, w_{i} - \operatorname{bound}\left( \proposal{\mathbf{\distance}}, \proposal{\mathbf{w}}, T_i \right) \right)^{2}\,, \label{eq:proploss}
$}
\end{equation}
This loss resembles a half-quadratic version of the chi-squared histogram distance that is often used in statistics and computer vision~\cite{pele2010quadratic}. 
This loss is asymmetric because we only want to penalize the proposal weights for \emph{underestimating} the distribution implied by the NeRF MLP --- overestimates are to be expected, as the proposal weights will likely be more coarse than the NeRF weights, and will therefore form an upper envelope over it.
The division by $w_i$ guarantees that the gradient of this loss with respect to the bound is a constant value when the bound is zero, which leads to well-behaved optimization.
Because $\mathbf{t}$ and $\hat{\mathbf{t}}$ are sorted, Equation~\ref{eq:proploss} can be computed efficiently through the use of summed-area tables~\cite{crow1984summed}. Note that this loss is invariant to monotonic transformations of distance $t$ (assuming that $\mathbf{w}$ and $\proposal{\mathbf{w}}$ have already been computed in $t$-space) so it behaves identically whether applied to Euclidean ray $\distance$-distances or to normalized ray $\sdistance$-distances.

We impose this loss between the NeRF histogram $(\mathbf{t}, \mathbf{w})$ and all proposal histograms $(\hat{\mathbf{t}}^k, \hat{\mathbf{w}}^k)$. The NeRF MLP is supervised using a reconstruction loss with the input image $\reconloss$, as in mip-NeRF. We place a stop-gradient on the NeRF MLP's outputs $\mathbf{t}$ and $\mathbf{w}$ when computing $\proploss$  so that the NeRF MLP ``leads'' and the proposal MLP ``follows'' --- otherwise the NeRF may be encouraged to produce a worse reconstruction of the scene so as to make the proposal MLP's job less difficult.
The effect of this proposal supervision can be seen in Figure~\ref{fig:traces}, where the NeRF MLP gradually localizes its weights $\mathbf{w}$ around a surface in the scene, while the proposal MLP ``catches up'' and predicts coarse proposal histograms that envelope the NeRF weights.

\section{Regularization for Interval-Based Models}
\label{sec:regularize}

\begin{figure}[b!]
    \centering
    \begin{tabular}{@{}c@{\,}|@{\,}c@{\,}|@{\,}c@{}}
        \includegraphics[width=0.32\linewidth]{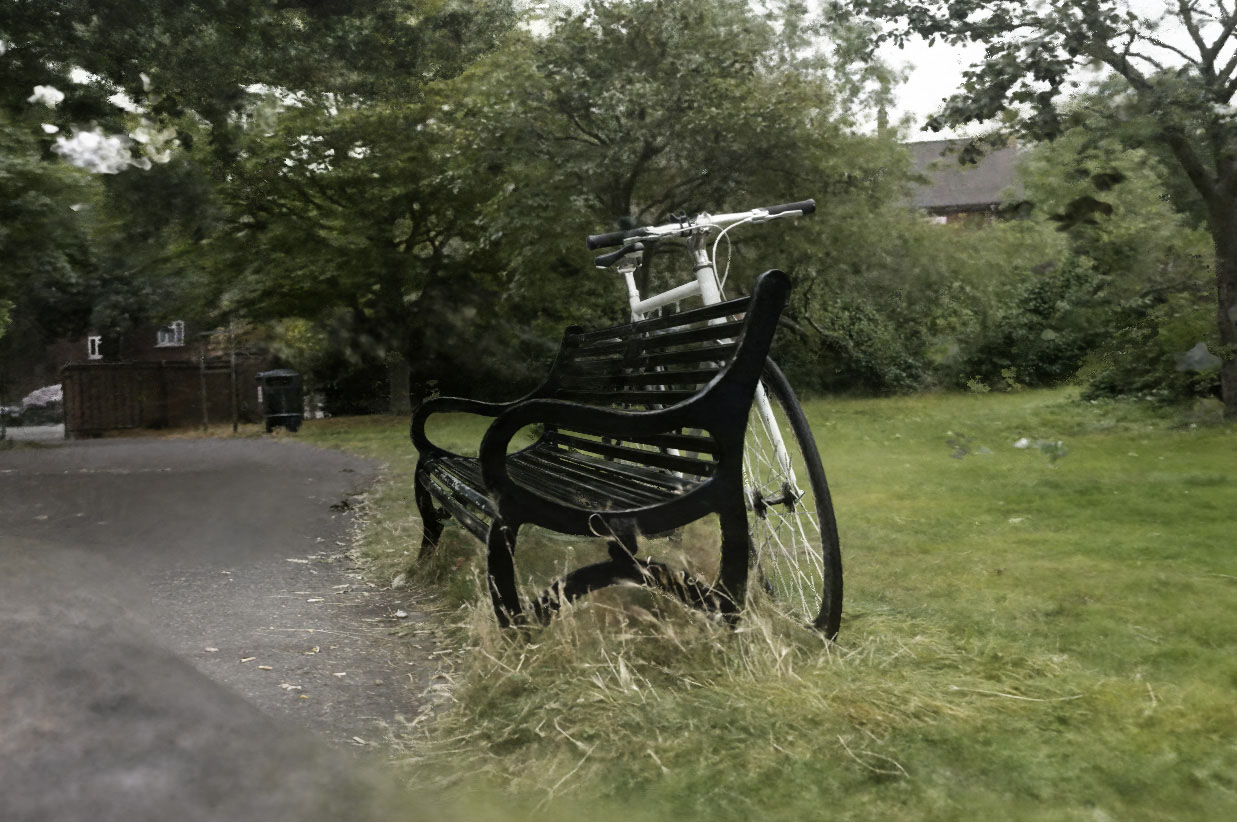} & 
        \includegraphics[width=0.32\linewidth]{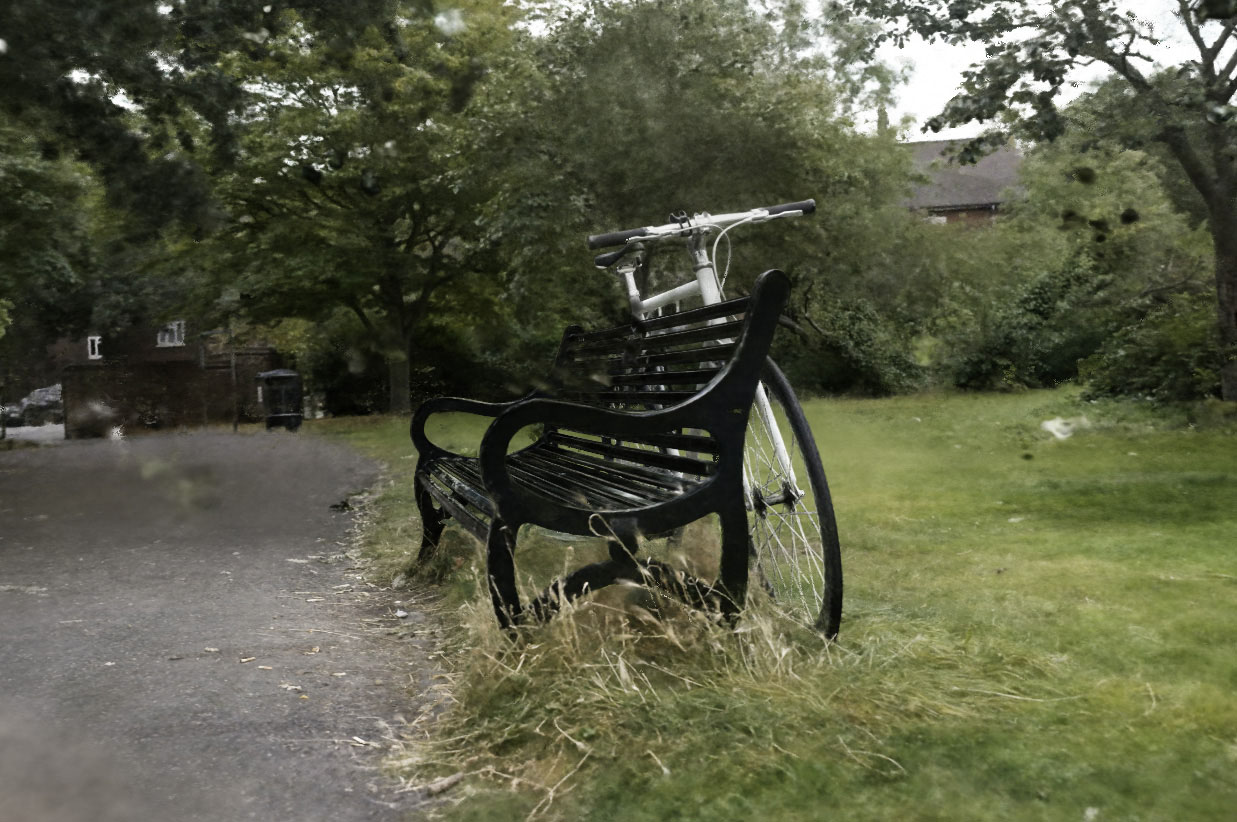} & 
        \includegraphics[width=0.32\linewidth]{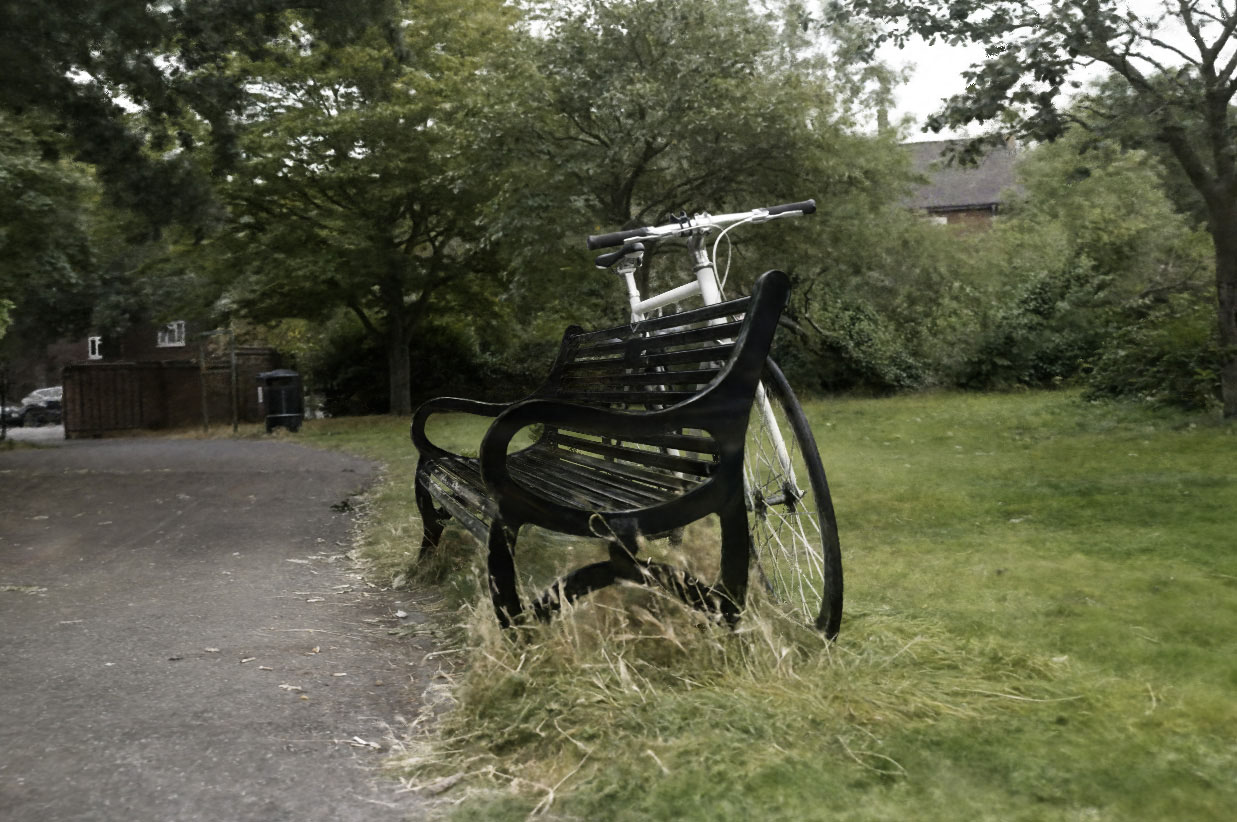} \\
        \includegraphics[width=0.32\linewidth]{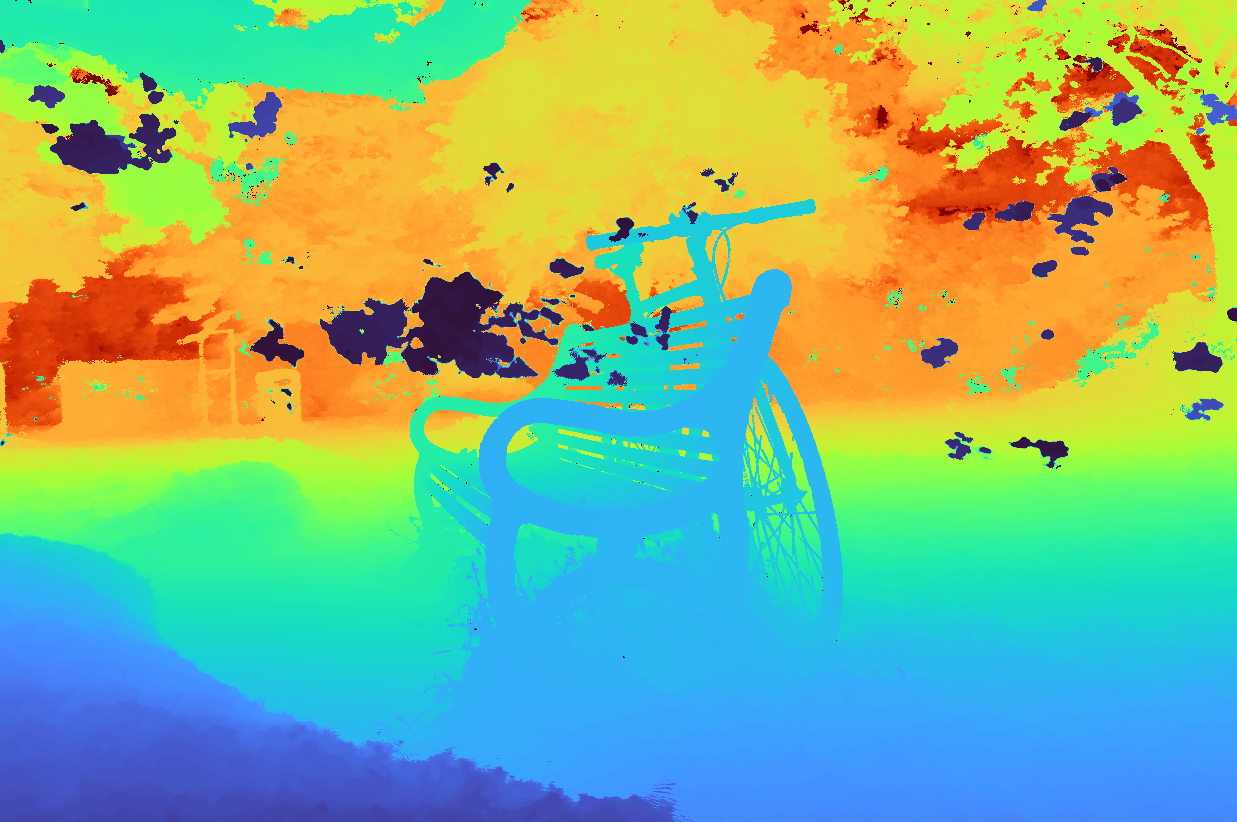} & 
        \includegraphics[width=0.32\linewidth]{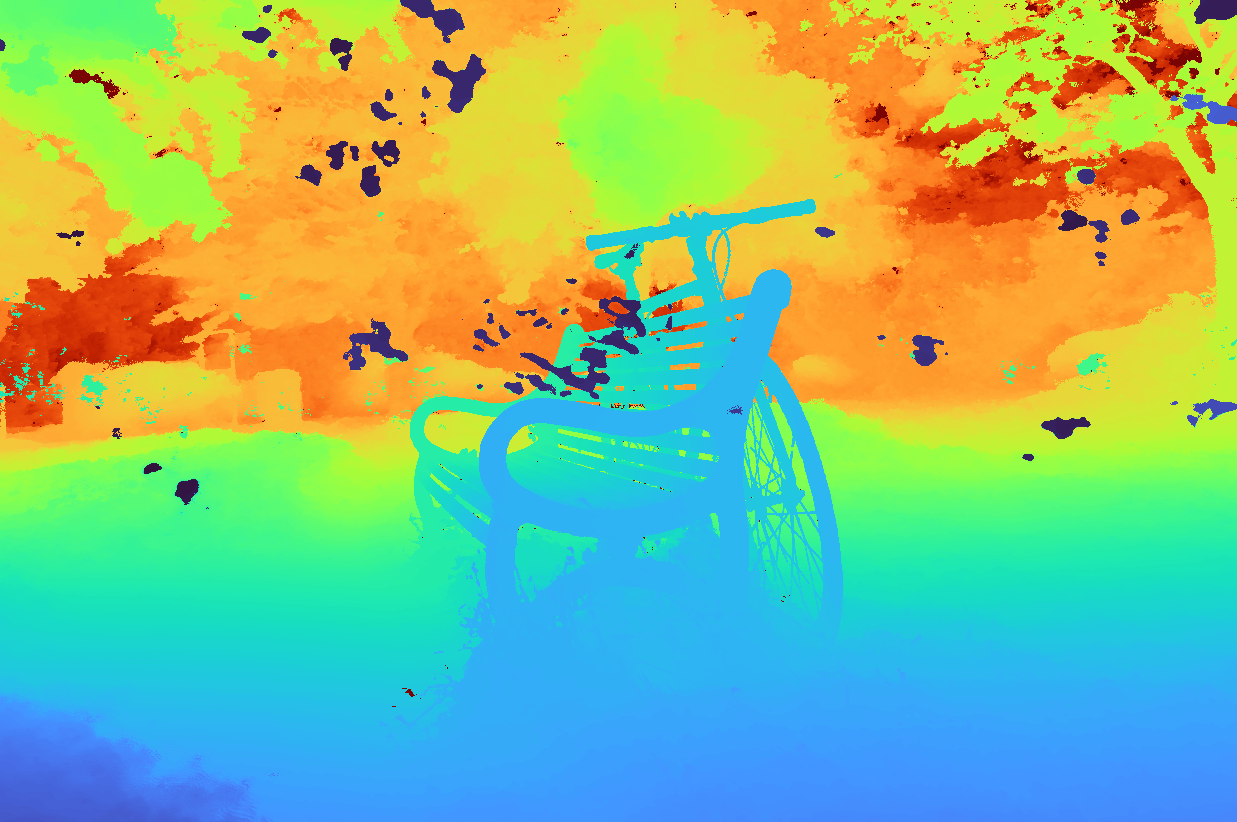} & 
        \includegraphics[width=0.32\linewidth]{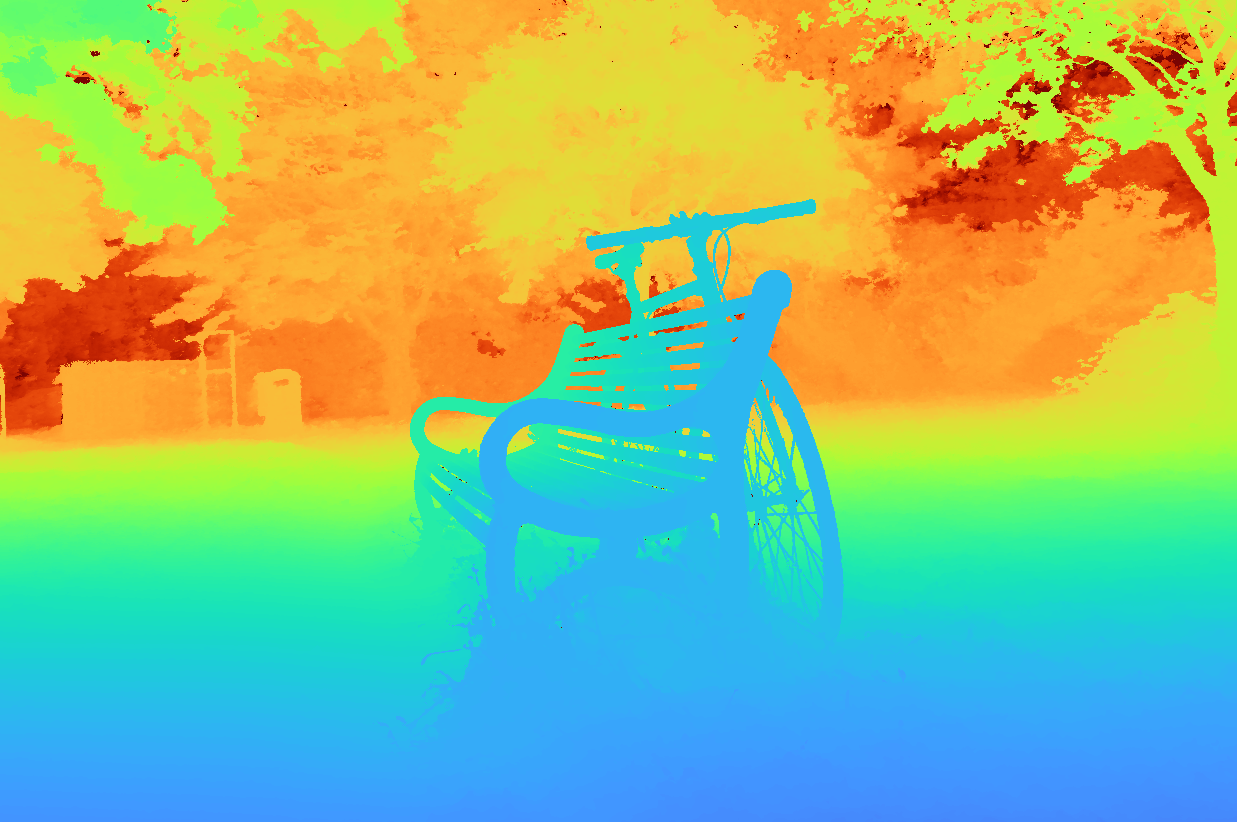} \\
        \scriptsize (a) no $\distloss$ & \scriptsize (b) no $\distloss$, w/noise\cite{mildenhall2020} & \scriptsize (c) with $\distloss$
    \end{tabular}
    \vspace{-0.05in}
    \caption{
    Our regularizer suppresses ``floaters'' (pieces of semi-transparent material floating in space, which are easy to identify in the depth map) and prevents a phenomenon in which surfaces in the background ``collapse'' towards the camera (shown in the bottom left of (a)). The noise-injection approach of Mildenhall~\etal~\cite{mildenhall2020} only partially eliminates these artifacts, and reduces reconstruction quality (note the lack of detail in the depths of the distant trees). See the supplemental video for more visualizations.
    }
    \label{fig:distortion_results}
\end{figure}

Due to ill-posedness, trained NeRFs often exhibit two characteristic artifacts we will call ``floaters'' and ``background collapse'', both shown in Figure~\ref{fig:distortion_results}(a). By ``floaters'' we refer to small disconnected regions of volumetrically dense space which serve to explain away some aspect of a subset of the input views, but when viewed from another angle look like blurry clouds. By ``background collapse'' we mean a phenomenon in which distant surfaces are incorrectly modeled as semi-transparent clouds of dense content close to the camera. Here we presents a regularizer that, as shown in Figure~\ref{fig:distortion_results}, prevents floaters and background collapse more effectively than the approach used by NeRF of injecting noise into volumetric density~\cite{mildenhall2020}.

\newcommand{\distnum}{002}

\begin{figure}[t]
    \centering
    \includegraphics[width=3.2in,trim=0.15in 0.15in 0.05in 0, clip]{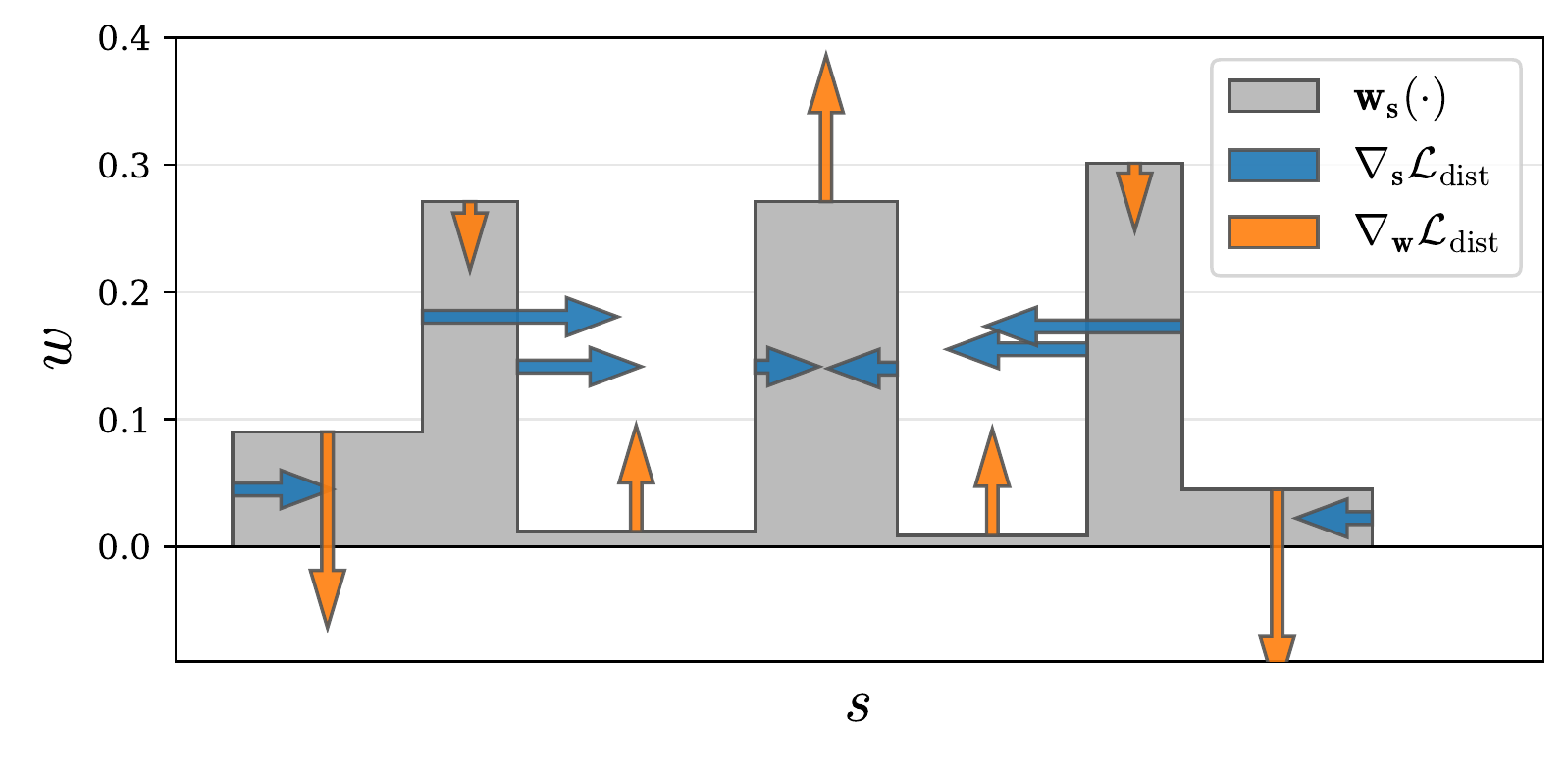}
    \vspace{-0.05in}
    \caption{
    A visualization of $\nabla \distloss$, the gradient of our regularizer, as a function of $\mathbf{s}$ and $\mathbf{w}$ on a toy step function. Our loss encourages each ray to be as compact as possible by 1) minimizing the width of each interval, 2) pulling distant intervals towards each other, 3) consolidating weight into a single interval or a small number of nearby intervals, and 4) driving all weights towards zero when possible (such as when the entire ray is unoccupied).
    }
    \label{fig:distortion}
\end{figure}

Our regularizer has a straightforward definition in terms of the step function defined by the set of (normalized) ray distances $\mathbf{\sdistance}$ and weights $\mathbf{w}$ that parameterize each ray:
\begin{equation}
\distloss(\mathbf{\sdistance}, \mathbf{w}) =\iint\limits_{-\infty }^{\,\,\,\infty }\mathbf{w}_\mathbf{\sdistance}(u)\mathbf{w}_\mathbf{\sdistance}(v)  |u - v|\,d_{u}\,d_{v}\,, \label{eq:def_distortion}
\end{equation}
where $\mathbf{w}_\mathbf{\sdistance}(u)$ is interpolation into the step function defined by $(\mathbf{\sdistance}, \mathbf{w})$ at $u$: 
$\mathbf{w}_\mathbf{\sdistance}(u) = \sum_i w_i \mathbbm{1}_{[\sdistance_i, \sdistance_{i+1})}(u)$.
We use normalized ray distances $\mathbf{\sdistance}$ because using $\mathbf{\distance}$ significantly upweights distant intervals and causes nearby intervals to be effectively ignored.
This loss is the integral of the distances between all pairs of points along this 1D step function, scaled by the weight $w$ assigned to each point by the NeRF MLP. We refer to this as ``distortion'', as it resembles a continuous version of the distortion minimized by k-means (though it could also be thought of as maximizing a kind of autocorrelation).
This loss is minimized by setting $\mathbf{w}=\mathbf{0}$ (recall that $\mathbf{w}$ sums to \emph{no more than} $1$, not exactly $1$). If that is not possible (i.e., if the ray is non-empty), it is minimized by consolidating weights into as small a region as possible.
Figure~\ref{fig:distortion} illustrates this behavior by showing the gradient of this loss on a toy histogram.

Though Equation~\ref{eq:def_distortion} is straightforward to define, it is non-trivial to compute.
But because $\mathbf{w}_\mathbf{\sdistance}(\cdot)$ has a constant value within each interval we can rewrite Equation~\ref{eq:def_distortion} as:
\begin{align}
\distloss(\mathbf{\sdistance}, \mathbf{w}) &=
\sum_{i,j} w_{i} w_{j} \left| \frac{\sdistance_{i} +\sdistance_{i+1}}{2} - \frac{\sdistance_{j} + \sdistance_{j+1}}{2} \right|  \nonumber \\
&+ \frac{1}{3}\sum _{i} w_{i}^{2}( \sdistance_{i+1} - \sdistance_{i}) 
\label{eq:alg_distortion} 
\end{align}
In this form, our distortion loss is trivial to compute. This reformulation also provides some intuition for how this loss behaves: the first term minimizes the weighted distances between all pairs of interval midpoints, and the second term minimizes the weighted size of each individual interval.

\section{Optimization}

Now that we have described our model components in general terms, we can detail the specific model used in all experiments.
We use a proposal MLP with $4$ layers and $256$ hidden units and a NeRF MLP with $8$ layers and $1024$ hidden units, both of which use ReLU internal activations and a softplus activation for density $\tau$. We do two stages of evaluation and resampling of the proposal MLP each using 64 samples to produce $(\hat{\mathbf{s}}^0, \hat{\mathbf{w}}^0)$ and $(\hat{\mathbf{s}}^1, \hat{\mathbf{w}}^1)$, and then one stage of evaluation of the NeRF MLP using 32 samples to produce $(\mathbf{s}, \mathbf{w})$.
We minimize the following loss:
\begin{equation}
\resizebox{2.9in}{!}{$
\!\!\!\displaystyle \reconloss(\Col(\zvec), \trueCol)
    + \lambda \distloss(\mathbf{\sdistance}, \mathbf{w}) +
    \sum_{k=0}^1 \proploss\left(\mathbf{s}, \mathbf{w}, \proposal{\mathbf{s}}^k, \proposal{\mathbf{w}}^k \right)\, ,
$}
\end{equation}
averaged over all rays in each batch (rays are not included in our notation).
The $\lambda$ hyperparameter balances our data term $\reconloss$ and our regularizer $\distloss$; we set $\lambda=0.01$ in all experiments. 
The stop-gradient used in $\proploss$ makes the optimization of $\modelweights_{\mathrm{prop}}$ independent from the optimization of $\modelweights_{\mathrm{NeRF}}$, and as such there is no need for a hyperparameter to scale the effect of $\proploss$.
For $\reconloss$ we use  Charbonnier loss~\cite{charbonnier1994two}: $\sqrt{(x - x^*)^2 + \epsilon^2}$ with $\epsilon=0.001$, which achieves slightly more stable optimization than the mean squared error used in mip-NeRF.
We train our model (and all reported NeRF-like baselines) using a slightly modified version of mip-NeRF's learning schedule: $250\mathrm{k}$ iterations of optimization with a batch size of $2^{14}$, using Adam~\cite{adam} with hyperparameters $\beta_1 = 0.9$, $\beta_2= 0.999$, $\epsilon=10^{-6}$, a learning rate that is annealed log-linearly from $2 \times 10^{-3}$ to $2 \times 10^{-5}$ with a warm-up phase of $512$ iterations, and gradient clipping to a norm of $10^{-3}$.

\section{Results}

\newcommand{\resultswidth}{0.184\linewidth}
\begin{figure*}[t!]
    \centering
    \begin{tabular}{@{}c@{\,}c@{\,}|@{\,}c@{\,\,}c@{\,\,}c@{\,\,}c@{\,\,}c@{\,\,}c@{}}
        \rotatebox{90}{\quad \quad \quad \scriptsize Depth Map} & \includegraphics[width=\resultswidth]{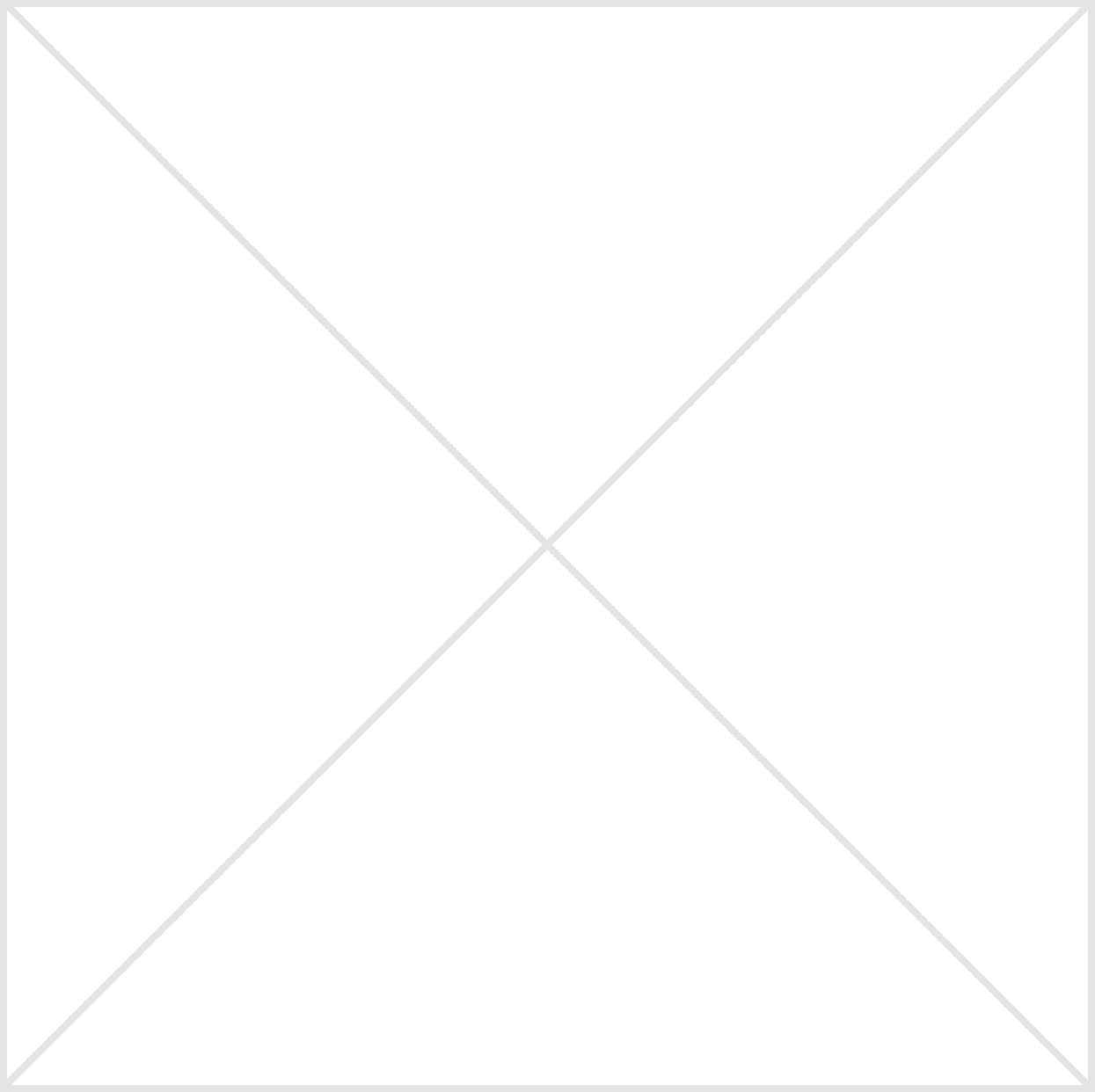} &
        \includegraphics[width=\resultswidth]{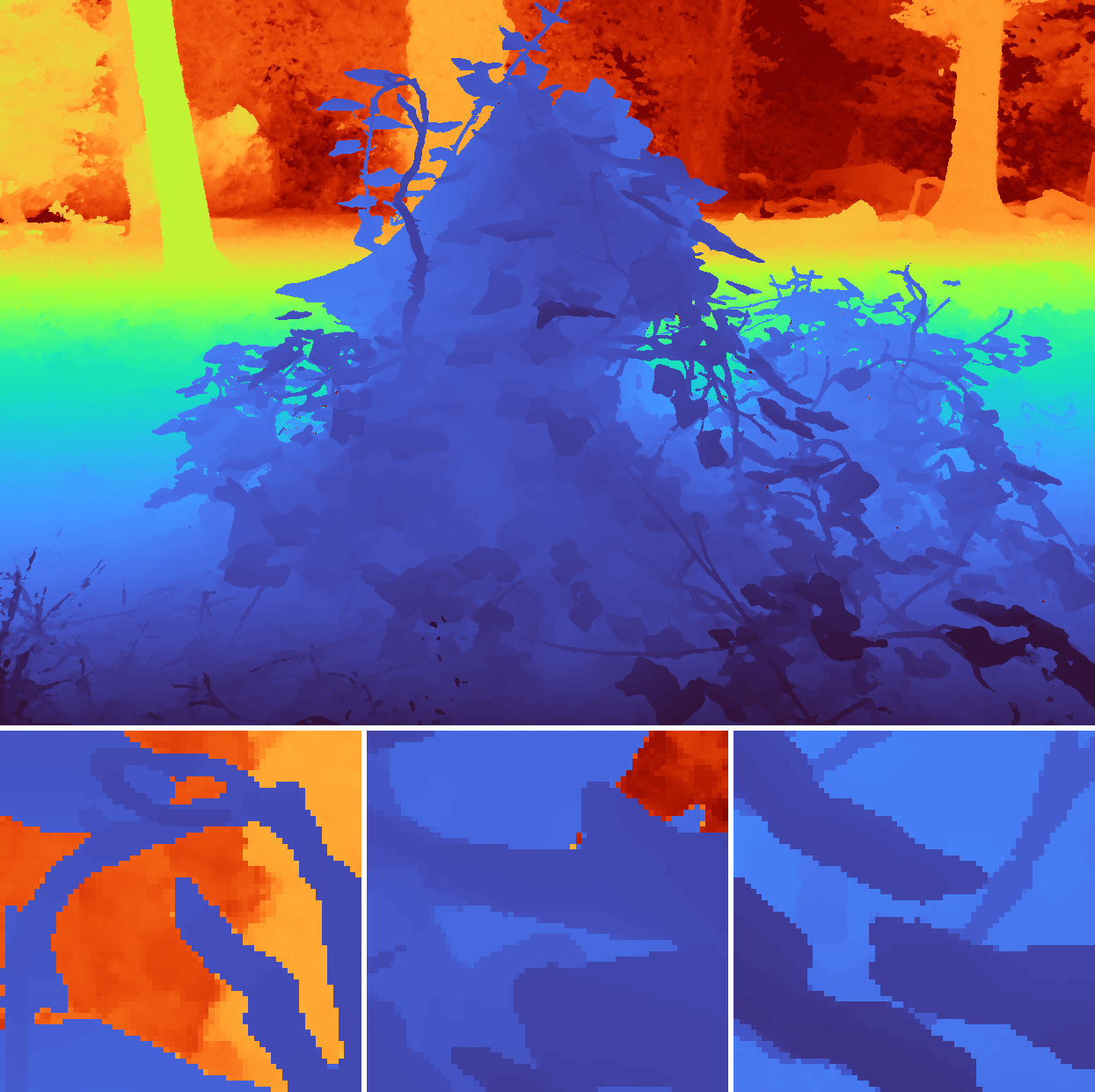} & 
        \includegraphics[width=\resultswidth]{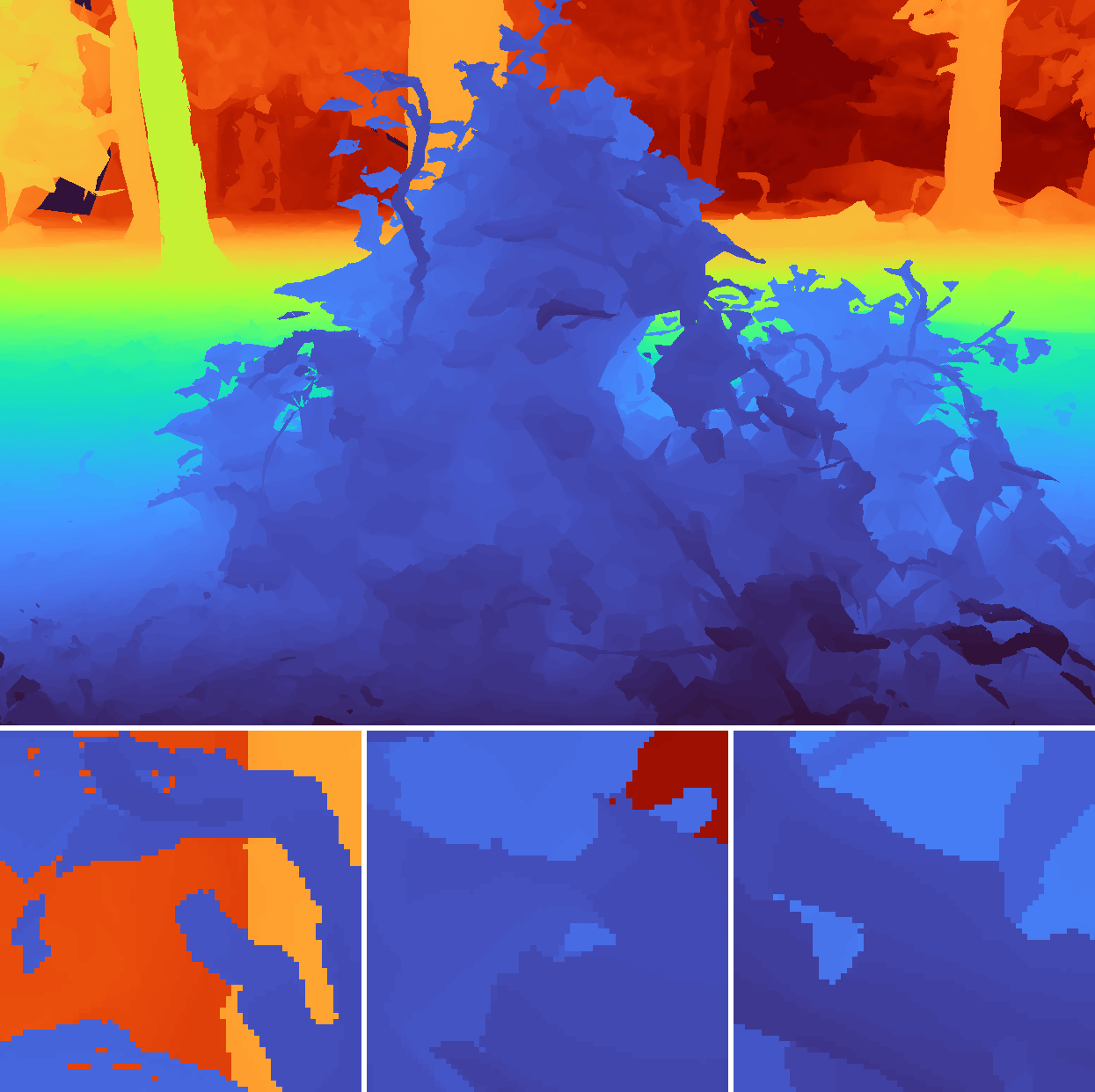} &
        \includegraphics[width=\resultswidth]{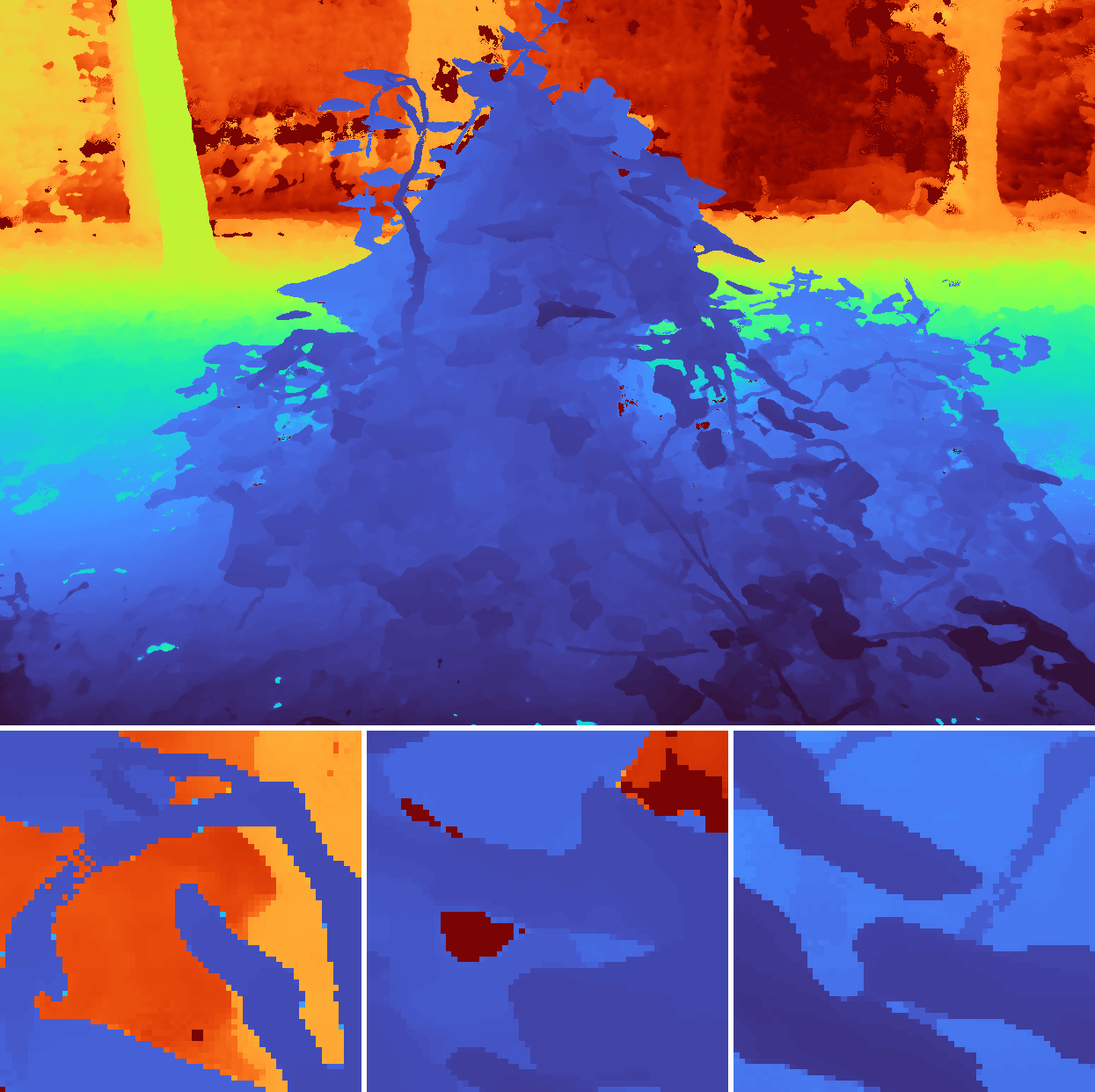} &
        \includegraphics[width=\resultswidth]{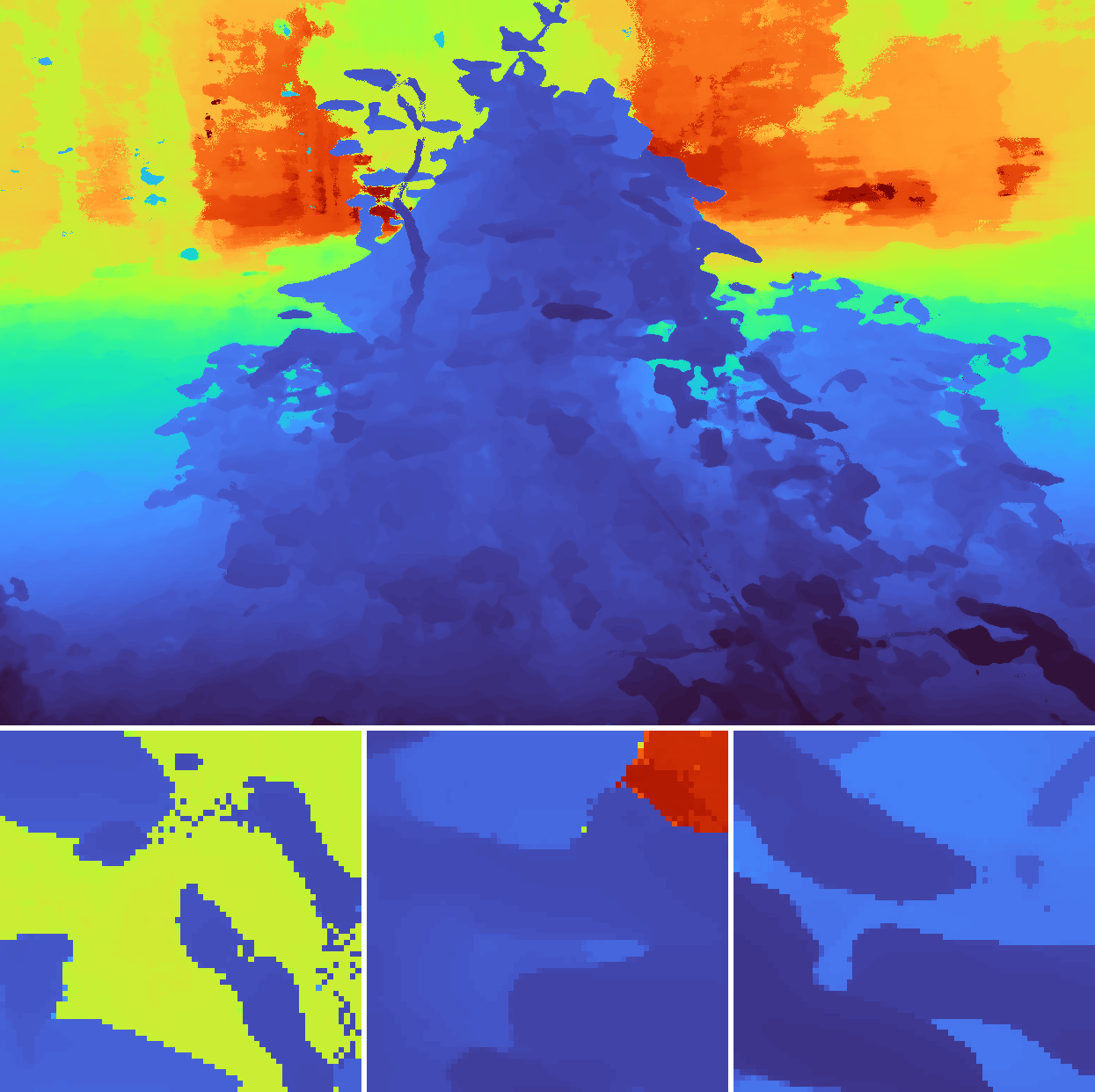} \\
        
        \rotatebox{90}{\quad \quad \,\,\, \scriptsize Color Image} & \includegraphics[width=\resultswidth]{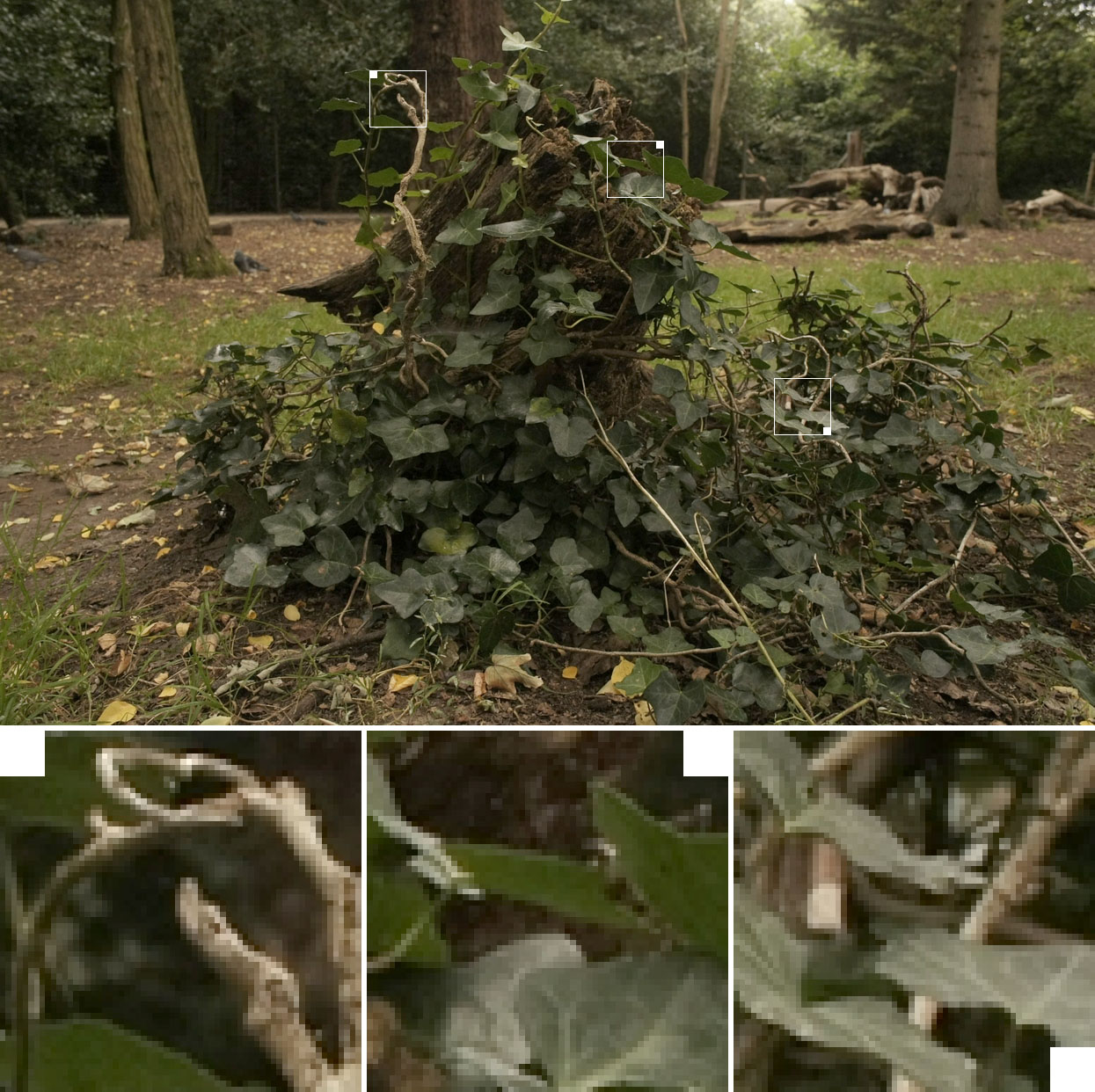} & 
        \includegraphics[width=\resultswidth]{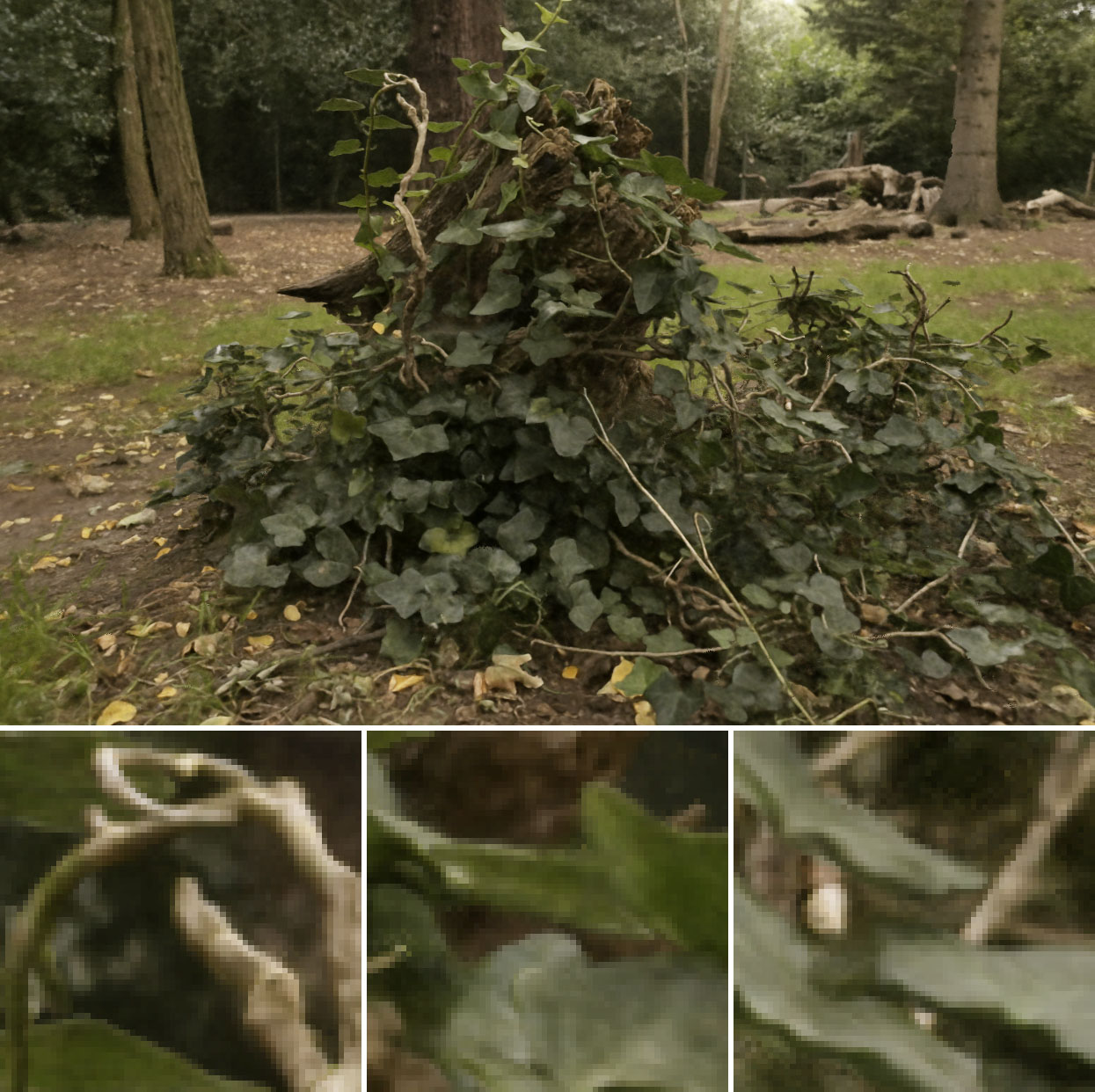} & 
        \includegraphics[width=\resultswidth]{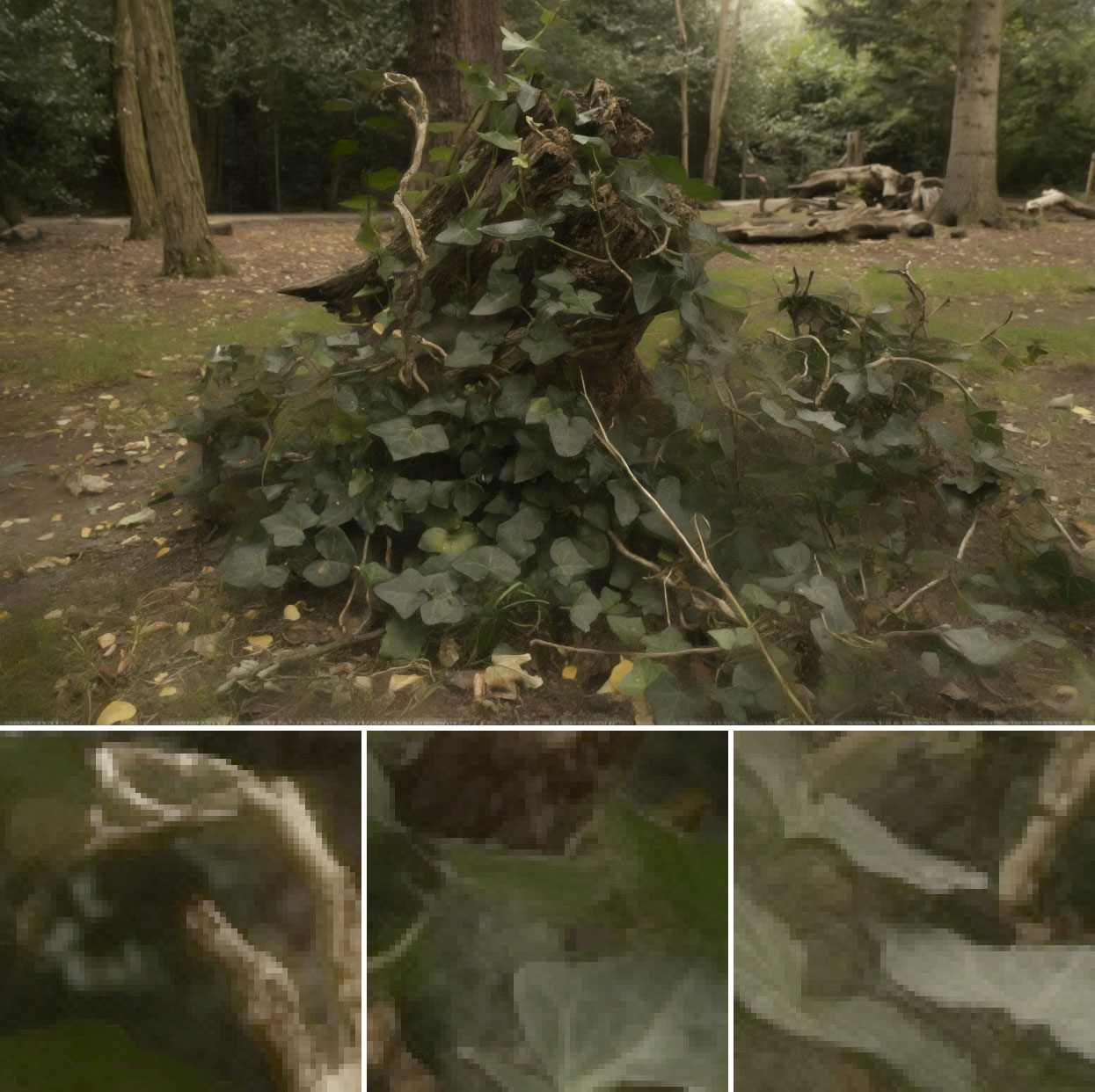} & 
        \includegraphics[width=\resultswidth]{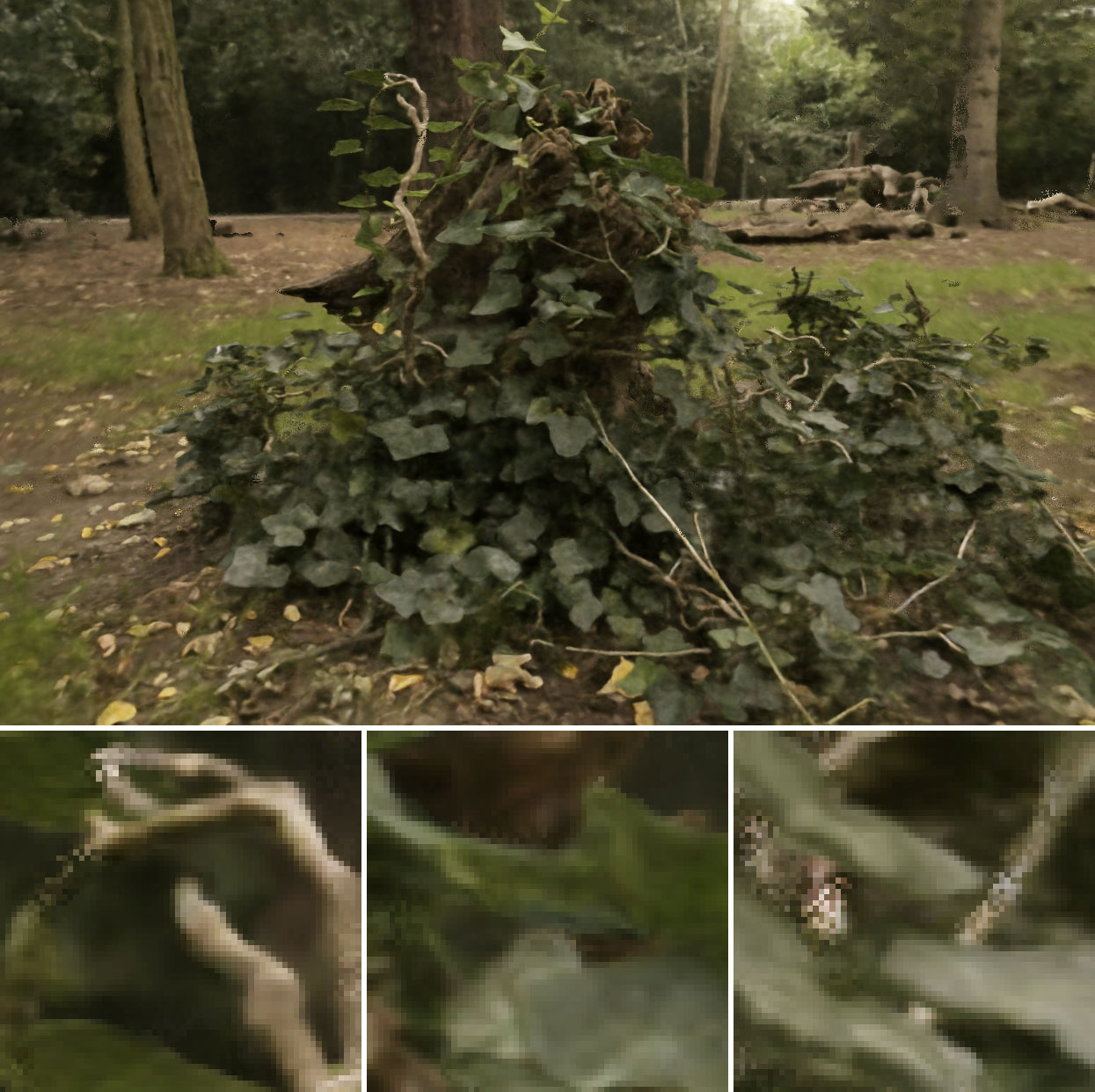}& 
        \includegraphics[width=\resultswidth]{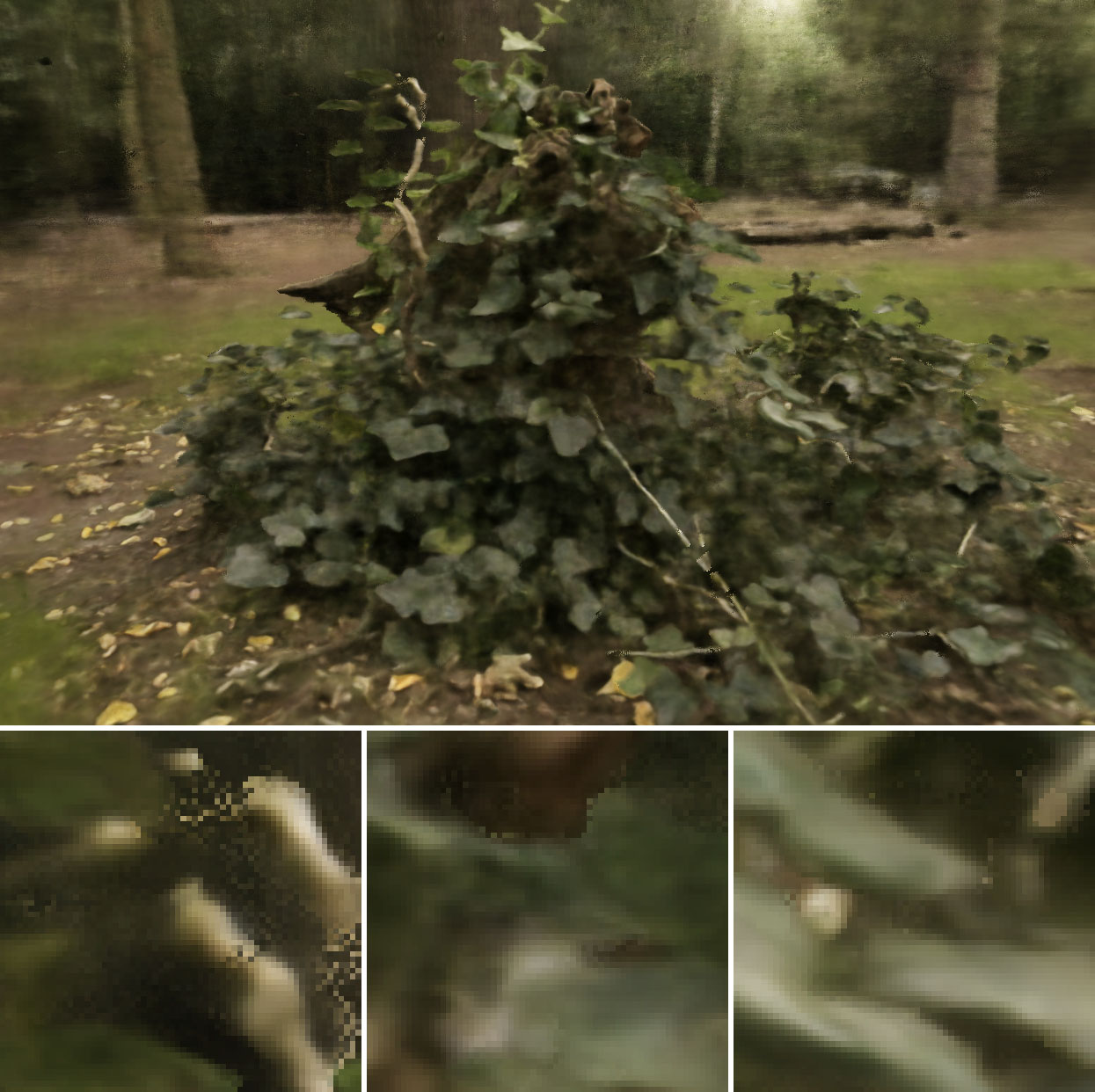} \\
        & \scriptsize (a) Ground-Truth Test-Set Image & \scriptsize (b) Our Model, SSIM=0.738 & \scriptsize (c) SVS~\cite{riegler2021svs, schoenberger2016sfm}, SSIM=0.680 & \scriptsize (d) NeRF++~\cite{kaizhang2020}, SSIM=0.622 & \scriptsize (e) mip-NeRF~\cite{barron2021mipnerf}, SSIM=0.522
    \end{tabular}
    \vspace{-0.1in}
    \caption{
    (a) A test-set image from our dataset's \textit{stump} scene, with (b) our model's rendered image and depth map (median ray termination distance~\cite{park2021nerfies}). Cropped patches are shown to highlight details. Compared to prior work (c-e) our renderings more closely resemble the ground-truth and our depths look more plausible (though no ground-truth depth is available). See the \suppname for more results.
    }
    \label{fig:results}
\end{figure*}

\begin{table}[b!]
    \centering
    \resizebox{\linewidth}{!}{
    \begin{tabular}{@{}l@{\,\,}|ccc|@{\,}c@{\,}|@{\,}r@{}}
    & \!PSNR $\uparrow$\! & \!SSIM $\uparrow$\! & \!LPIPS $\downarrow$\! & Time (hrs) & \# Params \\ \hline
    \input{tables/360.tex}
    \end{tabular}
    }
    \vspace{-0.1in}
    \caption{
    A quantitative comparison of our model with several prior works using the dataset presented in this paper.
    }
    \label{tab:avg_360_results}
\end{table}

We evaluate our model on a novel dataset: 9 scenes (5 outdoors and 4 indoors) each containing a complex central object or area and a detailed background. During capture we attempted to prevent photometric variation by fixing camera exposure settings, minimizing lighting variation, and avoiding moving objects --- we do not intend to probe all challenges presented by ``in the wild'' photo collections~\cite{martinbrualla2020nerfw}, only scale. Camera poses are estimated using COLMAP~\cite{schoenberger2016sfm}, as in NeRF. See the \suppname for details.

\paragraph{Compared methods.} We compare our model with NeRF~\cite{mildenhall2020} and mip-NeRF~\cite{barron2021mipnerf}, both using additional positional encoding frequencies so as to bound the entire scene inside the coordinate space used by both models. We evaluate against NeRF++~\cite{kaizhang2020}, which uses two MLPs to separately encode the ``inside'' and ``outside'' of each scene. We also evaluate against a version of NeRF that uses DONeRF's~\cite{donerf} scene parameterization, which uses logarithmically-spaced samples and a different contraction from our own. We also evaluate against mip-NeRF and NeRF++ variants in which the MLP(s) underlying each model have been scaled up to roughly match our own model in terms of number of parameter count (1024 hidden units for mip-NeRF, 512 hidden units for both MLPs in NeRF++). We evaluate against Stable View Synthesis~\cite{riegler2021svs}, a non-NeRF model that represents the state-of-the-art of a different view-synthesis paradigm in which neural networks are trained on external scenes and combined with a proxy geometry produced by structure-from-motion~\cite{schoenberger2016sfm}. We additionally compare with the publicly available SIBR implementations~\cite{sibr2020} of Deep Blending~\cite{hedman2018deep} and Point-Based Neural Rendering~\cite{kopanas2021point}, two real-time IBR-based view synthesis approaches that also depend on an external proxy geometry.
We also present a variant of our own model in which we use the latent appearance embedding (4 dimensions) presented in NeRF-W~\cite{martinbrualla2020nerfw, bojanowski2018optimizing} which ameliorates artifacts caused by inconsistent lighting conditions during scene capture (because our scenes do not contain transient objects, we do not benefit from NeRF-W's other components).

\paragraph{Comparative evaluation.} In Table~\ref{tab:avg_360_results} we report mean PSNR, SSIM~\cite{wang2004image}, and LPIPS~\cite{zhang2018unreasonable} across the test images in our dataset. For all NeRF-like models, we report train times from a TPU v2 with 32 cores~\cite{jouppi2017datacenter}, as well as model size (the train times and model sizes of SVS, Deep Blending, and Point-Based Neural Rendering are not presented, as this comparison would not be particularly meaningful). Our model outperforms all prior NeRF-like models by a significant margin, and we see a 57\% reduction in mean squared error relative to mip-NeRF with a 2.17$\times$ increase in train time. The mip-NeRF and NeRF++ baselines that use larger MLPs are more competitive, but are $\sim$3$\times$ slower to train than our model and still achieve significantly lower accuracies. 
Our model outperforms Deep Blending and Point-Based Neural Rendering across all error metrics. It also outperforms SVS for PSNR and SSIM, but not LPIPS. This may be due to SVS being supervised to directly minimize an LPIPS-like perceptual loss, while we minimize a per-pixel reconstruction loss. See the \suppname for renderings from SVS that achieve lower LPIPS scores than our model despite having reduced image quality~\cite{kettunen2019lpips}. Our model has several advantages over SVS and Deep Blending in addition to image quality: those models require external training data while our model does not, those models require the proxy geometry produced by a MVS package (and may fail when that geometry is incorrect) while we do not, and our model produces extremely detailed depth maps while SVS and Deep Blending do not (the ``SVS depths'' we show were produced by COLMAP~\cite{schoenberger2016sfm} and are used as input to the model).
Figure~\ref{fig:results} shows model outputs, though we encourage the reader to view our supplemental video.

\begin{table}[b!]
    \centering
    \resizebox{\linewidth}{!}{
    \begin{tabular}{@{}r@{\,\,}l@{\,\,}|ccc|@{\,}c@{\,}|@{\,}r@{}}
    & & \!PSNR $\uparrow$\! & \!SSIM $\uparrow$\! & \!LPIPS $\downarrow$\! & Time (hrs) & \# Params \\ \hline
    \input{tables/ablation.tex}
    \end{tabular}
    }
    \vspace{-0.1in}
    \caption{
    An ablation study  in which we remove or replace model components to measure their effect. See the text for details.
    }
    \label{tab:ablation}
\end{table}

\paragraph{Ablation study.} In Table~\ref{tab:ablation} we present an ablation study of our model on the \textit{bicycle} scene in our dataset, the findings of which we summarize here. A) Removing $\proploss$ significantly reduces performance, as the proposal MLP is not supervised during training. B) Removing $\distloss$ does not substantially affect our metrics but results in ``floater'' artifacts in scene geometry, as shown in Figure~\ref{fig:distortion_results}. C) the regularization proposed by Mildenhall \etal~\cite{mildenhall2020} of injecting Gaussian noise ($\sigma=1$) into density degrades performance (and as shown in Figure~\ref{fig:distortion_results} is less effective at eliminating floaters). D) Removing the proposal MLP and using a single MLP to model both the scene and the proposal weights does not degrade performance but increases training time by $\sim$3$\times$, hence our small proposal MLP. E) Removing the proposal MLP and training our model using mip-NeRF's approach (applying $\reconloss$ at all coarse scales instead of using our $\proploss$) worsens both speed \emph{and} accuracy, justifying our supervision strategy. F) Using a small NeRF MLP (256 hidden units instead of our 1024 hidden units) accelerates training but reduces quality, demonstrating the value of a high-capacity model when dealing with detailed scenes. G) Removing IPE completely and using NeRF's positional encoding~\cite{mildenhall2020} reduces performance, showing the value in building upon mip-NeRF instead of NeRF.
H) Ablating the contraction and instead adding positional encoding frequencies to bound the scene decreases accuracy and speed. I) Using the parameterization and logarithmic ray-spacing presented in DONeRF~\cite{donerf} reduces accuracy.

\paragraph{Limitations.} Though mip-NeRF 360 significantly outperforms mip-NeRF and other prior work, it is not perfect. Some thin structures and fine details may be missed, such as the tire spokes in the \textit{bicycle} scene (Figure~\ref{fig:distortion_results}), or the veins on the leaves in the \textit{stump} scene (Figure~\ref{fig:results}). View synthesis quality will likely degrade if the camera is moved far from the center of the scene. And, like most NeRF-like models, recovering a scene requires several hours of training on an accelerator, precluding on-device training.

\section{Conclusion}

We have presented mip-NeRF 360, a mip-NeRF extension designed for real-world scenes with unconstrained camera orientations. Using a novel Kalman-like scene parameterization, an efficient proposal-based coarse-to-fine distillation framework, and a regularizer designed for mip-NeRF ray intervals, we are able to synthesize realistic novel views and complex depth maps for challenging unbounded real-world scenes, with a 57\% reduction in mean-squared error compared to mip-NeRF.

\paragraph{Acknowledgements.} Our sincere thanks to David Salesin and Ricardo Martin-Brualla for their help in reviewing this paper before submission, and to George Drettakis and Georgios Kopanas for their help in evaluating baselines on our 360 dataset.

%% file: tables/360.tex
NeRF \cite{mildenhall2020, jaxnerf2020github}& 23.85 & 0.605 & 0.451 & 4.16 & 1.5M \\
NeRF w/ DONeRF~\cite{donerf} param. & 24.03 & 0.607 & 0.455 & 4.59 & 1.4M \\
mip-NeRF \cite{barron2021mipnerf}& 24.04 & 0.616 & 0.441 & 3.17 & 0.7M \\
NeRF++ \cite{kaizhang2020}& 25.11 & 0.676 & 0.375 & 9.45 & 2.4M \\
Deep Blending~\cite{hedman2018deep}& 23.70 & 0.666 & 0.318 & - & - \\
Point-Based Neural Rendering~\cite{kopanas2021point}& 23.71 & 0.735 & \cellcolor{yellow}0.252 & - & - \\
Stable View Synthesis~\cite{riegler2021svs}& 25.33 & \cellcolor{yellow}0.771 & \cellcolor{red}0.211 & - & - \\
mip-NeRF~\cite{barron2021mipnerf} w/bigger MLP & 26.19 & 0.748 & 0.285 & 22.71 & 9.0M \\
NeRF++~\cite{kaizhang2020} w/bigger MLPs & \cellcolor{orange}26.39 & 0.750 & 0.293 & 19.88 & 9.0M \\
\hline 
Our Model & \cellcolor{red}27.69 & \cellcolor{red}0.792 & \cellcolor{orange}0.237 & 6.89 & 9.9M \\
Our Model w/GLO & \cellcolor{yellow}26.26 & \cellcolor{orange}0.786 & \cellcolor{orange}0.237 & 6.90 & 9.9M \\

%% file: tables/ablation.tex
A) & No $\proploss$ & 20.49 & 0.406 & 0.573 & 6.21 & 9.0M \\
B) & No $\distloss$ & \cellcolor{red}24.41 & \cellcolor{red}0.687 & \cellcolor{red}0.300 & 7.08 & 9.0M \\
C) & No $\distloss$, w/Noise Injection & 24.00 & 0.655 & 0.328 & 7.08 & 9.0M \\
D) & No Proposal MLP & \cellcolor{yellow}24.26 & \cellcolor{orange}0.682 & \cellcolor{orange}0.307 & 18.89 & 8.7M \\
E) & No Prop. MLP w/\!\cite{barron2021mipnerf}'s Training & 23.45 & 0.659 & 0.328 & 18.89 & 8.7M \\
F) & Small NeRF MLP & 22.80 & 0.515 & 0.480 & 4.31 & 1.1M \\
G) & No IPE & 23.87 & \cellcolor{yellow}0.664 & \cellcolor{yellow}0.322 & 7.08 & 9.0M \\
H) & No Contraction & 23.77 & 0.642 & 0.347 & 8.79 & 10.9M \\
I) & w/DONeRFs Contraction~\cite{donerf} & 23.99 & 0.654 & 0.334 & 7.20 & 9.0M \\
\hline 
 & Our Complete Model & \cellcolor{orange}24.37 & \cellcolor{red}0.687 & \cellcolor{red}0.300 & 7.09 & 9.0M \\

%% file: supp_content.tex
\section{Additional Model Details}

Our model contains some small components not discussed in the main paper that improve performance slightly.

\paragraph{Off-Axis Positional Encoding.}

\begin{figure}[t!]
    \centering
    \begin{tabular}{@{}cc@{}}
    \begin{subfigure}{.41\linewidth}
      \centering
      \includegraphics[width=\textwidth,trim=0 0 1.5in 0, clip]{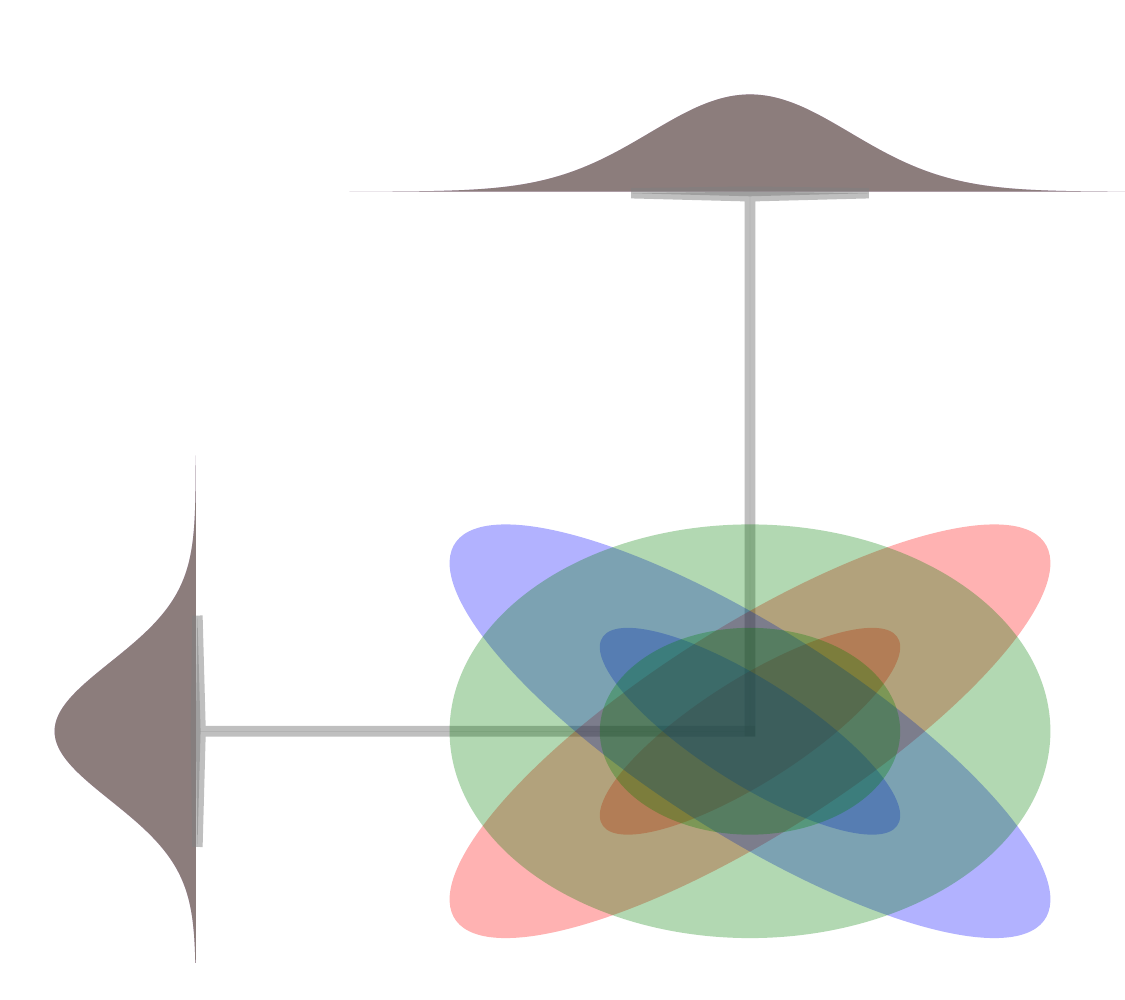}
      \caption{Axis-Aligned IPE~\cite{barron2021mipnerf}}
      \label{subfig:ipe}
    \end{subfigure} &
    \begin{subfigure}{.54\linewidth}
      \centering
      \includegraphics[width=\textwidth]{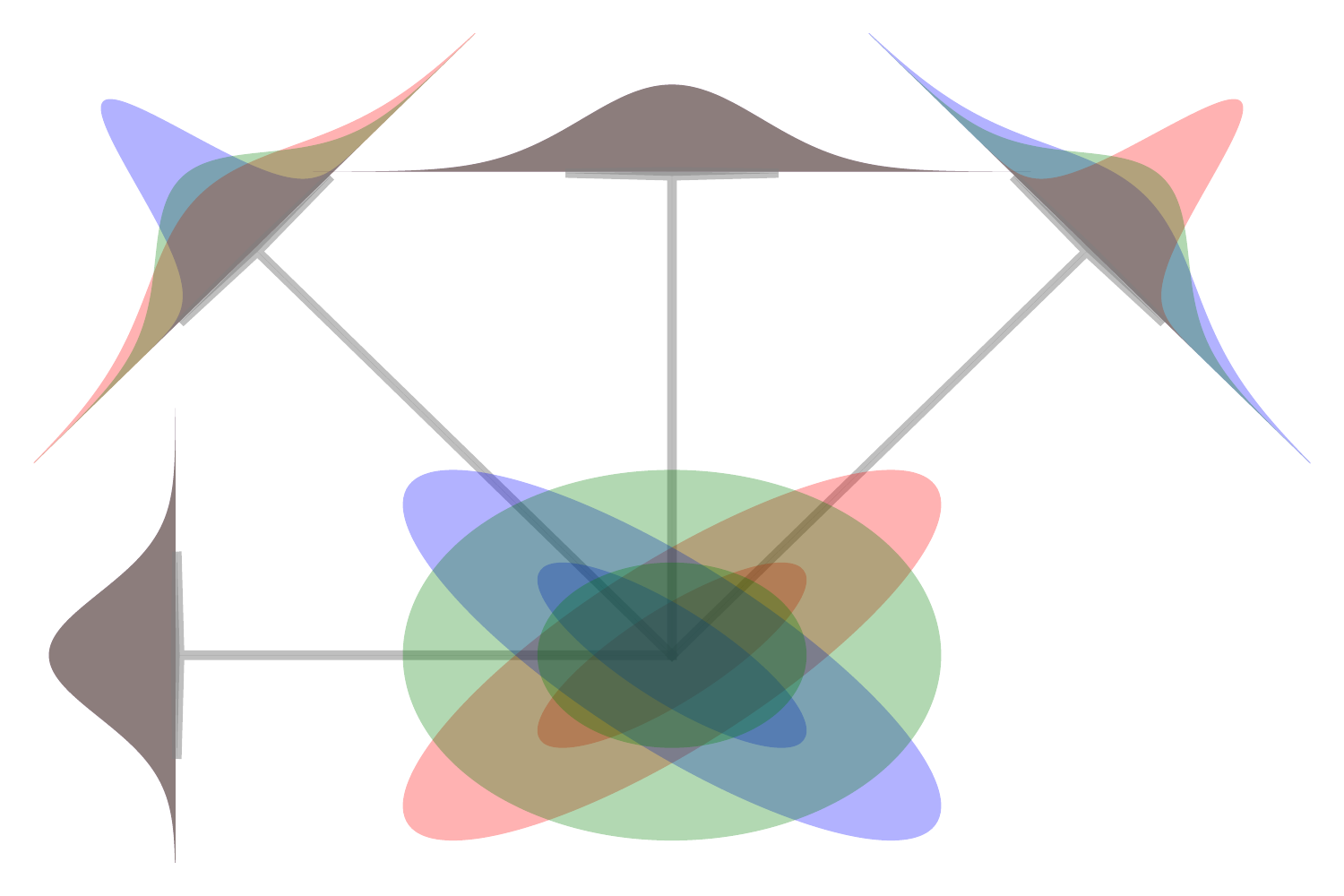}
      \caption{Off-Axis IPE}
      \label{subfig:ipe_aniso}
    \end{subfigure}
    \end{tabular}
    \caption{
    The axis-aligned positional encoding used by mip-NeRF~\cite{barron2021mipnerf} does not capture the covariance of the Gaussian being encoded. Here we plot three bivariate Gaussians colored red, green, and blue (a) with axis-aligned IPE and (b) with our off-axis IPE, and show the marginal distributions produced by projecting each Gaussians onto the basis used by each encoding. Because these Gaussians have identical marginal distributions, mip-NeRF's axis-aligned IPE produces identical features, while the off-axis projections of our approach allow them to be disambiguated.
    }
    \label{fig:off_axis}
\end{figure}

When constructing integrated positional encoding features, we must select a basis $\basis$. In mip-NeRF~\cite{barron2021mipnerf}, this basis is selected as the identity matrix. This is convenient, because it means that only the diagonal of the covariance matrix $\covmat$ is required to construct IPE features, and off-diagonal components need not be computed. However, the reparameterization used by our model requires access to a full covariance matrix, as otherwise the Kalman-like warping we use would be inaccurate in the presence of highly anisotropic Gaussians (which are frequent in distant parts of the scene). So given that we are required to construct a full $\covmat$ matrix, we take advantage of the extra information presented therein, and encode not just axis-aligned IPE features but off-axis IPE features as well. As our basis $\basis$, instead of an identity matrix we use a large skinny matrix that contains the unit-norm vertices of a twice-tessellated icosahedron, where redundant negative copies of vertices are removed. For reproducibility's sake this matrix is:
\begin{equation}
    \!\!\!\basis = \begin{bmatrix} 0.8506508 &  0 &  0.5257311 \\
             0.809017 &  0.5      &  0.309017 \\
             0.5257311 &  0.8506508 &  0     \\
             1       &  0       &  0     \\
             0.809017  &  0.5       & -0.309017 \\
             0.8506508 &  0       & -0.5257311 \\
             0.309017  &  0.809017  & -0.5     \\
             0       &  0.5257311 & -0.8506508 \\
             0.5       &  0.309017  & -0.809017 \\
             0       &  1       &  0     \\
            -0.5257311 &  0.8506508 &  0     \\
            -0.309017  &  0.809017  & -0.5     \\
             0       &  0.5257311 &  0.8506508 \\
            -0.309017  &  0.809017  &  0.5     \\
             0.309017  &  0.809017  &  0.5     \\
             0.5       &  0.309017  &  0.809017 \\
             0.5       & -0.309017  &  0.809017 \\
             0       &  0       &  1     \\
            -0.5       &  0.309017  &  0.809017 \\
            -0.809017  &  0.5       &  0.309017 \\
            -0.809017  &  0.5       & -0.309017
            \end{bmatrix}^\mathrm{T}\, .
\end{equation}
These off-axis features allow the model to encode the shape of anisotropic Gaussians (with a similar intuition as the random Fourier features explored by Tancik \etal \cite{tancik2020fourfeat}) which otherwise are indistinguishable using axis-aligned IPE features, as shown in Figure~\ref{fig:off_axis}.
Ablating these off-axis features reduces performance slightly, with SSIM for the \textit{bicycle} scene falling from 0.687 to 0.664. 

Computing IPE features with a large $\basis$ matrix using the procedure described in mip-NeRF ($\operatorname{diag}(\basis \covmat \basis^\transpose)$) is prohibitively expensive. A tractable alternative is to instead compute the equivalent expression $\operatorname{sum}(\basis^\mathrm{T} \circ (\covmat \basis^\mathrm{T}), 0)$ where $\circ$ is an element-wise product and $\operatorname{sum}(\cdot, 0)$ is summation over rows. With this small optimization, off-axis IPE features are only modestly more expensive to compute than the axis-aligned IPE features used in mip-NeRF. 

\paragraph{Annealing.}
Before resampling ray-intervals from proposal weights $\proposal{\mathbf{w}}$, we anneal those weights by raising them to a power. With $N$ training steps, at step $n$ we compute
\newcommand{\slope}{b}
\newcommand{\trainfrac}{\sfrac{n}{N}}
\begin{equation}
    \proposal{\mathbf{w}}_{n} \propto \proposal{\mathbf{w}}^{\frac{\slope \trainfrac}{(\slope - 1) \trainfrac + 1}}
\end{equation}
and use $\proposal{\mathbf{w}}_{n}$ when drawing samples. The exponent is Schlick's bias function~\cite{Schlick} applied to $\trainfrac \in [0, 1]$, which curves the exponent such that it quickly rises from 0 and saturates towards 1. We set the bias hyperparameter $\slope = 10$ in all experiments.
At the beginning of training the exponent is $0$, which yields a flat distribution ($\proposal{\mathbf{w}}_{0} \propto \mathbf{1}$), and at the end of training that power is $1$, which yields the proposal distribution ($\proposal{\mathbf{w}}_{N} = \proposal{\mathbf{w}}$). This annealing encourages ``exploration'' during training, by causing the NeRF MLP to be presented with a wider range of proposal intervals than it otherwise would towards the beginning of training. Annealing has a modest positive effect: ablating it causes SSIM on the \textit{bicycle} scene to decrease from 0.687 to 0.679.

\paragraph{Dilation.}
We slightly ``dilate'' each proposal histogram $(\proposal{\mathbf{t}}, \proposal{\mathbf{w}})$ before resampling it. This reduces aliasing artifacts, likely because the proposal MLP is supervised using only rays that correspond to input pixels, so its predictions may only hold for certain angles --- in a sense, the proposal network is rotationally aliased. By widening the intervals of the proposal MLP we help counteract this aliasing.
To dilate a histogram $(\proposal{\mathbf{s}}, \proposal{\mathbf{w}})$ we first compute $\proposal{\mathbf{p}}$ where $\proposal{p}_i = \proposal{w}_i / (\proposal{s}_{i+1} - \proposal{s}_{i})$, giving us a probability density that integrates to 1 rather than a histogram that sums to 1. We then dilate this by computing
\begin{equation}
  \max_{ s - \epsilon \leq s' < s + \epsilon}\proposal{\mathbf{p}}_{\proposal{\mathbf{s}}}(s') \label{eq:dilate}
\end{equation}
where $\proposal{\mathbf{p}}_{\proposal{\mathbf{s}}}(s)$ is interpolation into the step function defined by $\proposal{\mathbf{s}}, \proposal{\mathbf{p}}$ at $s$. Equation~\ref{eq:dilate} can be computed efficiently by constructing a new set of intervals whose endpoints are $\operatorname{sort}( \proposal{\mathbf{s}} \cup \proposal{\mathbf{s}} - \epsilon \cup \proposal{\mathbf{s}} + \epsilon \})$ and computing the max of all intervals in that expanded set. After dilation, we convert the dilated $\proposal{\mathbf{p}}$ back into a histogram by multiplying each element by the size of its interval, and then normalizing the resulting histogram to sum to 1.

When dilating each proposal histogram, we set the dilation factor $\epsilon$ to a function of the expected size of a histogram bin. That is, for each coarse-to-fine resampling level $k$ in which we resample $n_k$ fine intervals from the coarse histogram that preceded it, we compute the product of all sample counts that preceded level $k$ and set epsilon to an affine function of the inverse of that product:
\begin{equation}
    \epsilon_k = \frac{a}{\prod_{k'=1}^{k-1} n_k} + b
\end{equation}
where $a=0.5$ and $b = 0.0025$ are hyperparameters that respectively determine how much a histogram is dilated relative to the expected size of a histogram bin (which is the inverse of the product of prior sample counts) and in absolute terms.

\begin{figure}[t!]
    \centering
    \includegraphics[width=\linewidth]{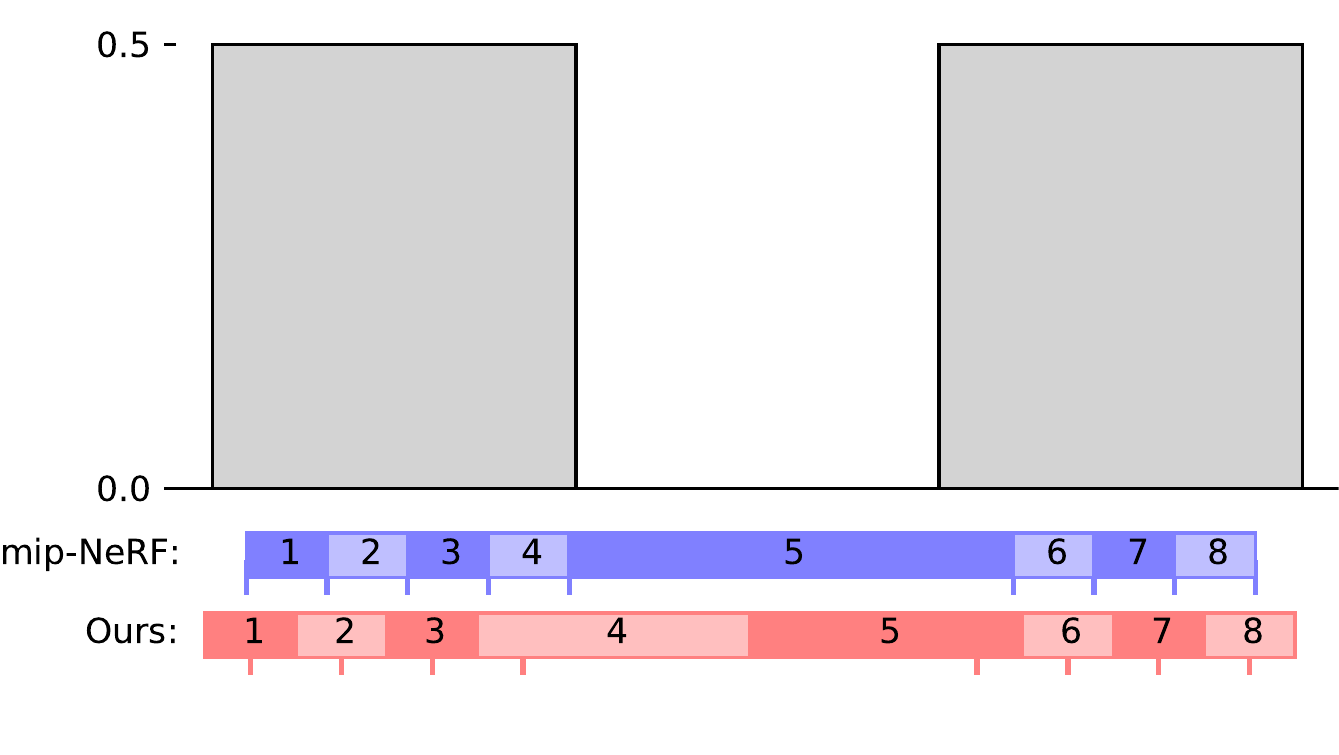}
    \caption{
    When resampling a histogram (such as the toy 3-bin histogram shown here in gray) into a set of $n$ intervals, mip-NeRF~\cite{barron2021mipnerf} samples $n+1$ sorted random values (blue ticks) from the histogram for use as the endpoints of intervals (blue boxes), which yields intervals that do not cover the front and back of histogram modes, and asymmetrically span gaps between modes. We instead draw $n+1$ sorted random values (red ticks) and use the midpoints of those samples as the endpoints of intervals (red boxes), resulting in less irregular resampling.
    }
    \label{fig:sample}
\end{figure}

\paragraph{Sampling.} Mip-NeRF generated $n$ ``fine'' intervals along each ray by sampling $n+1$ distances from the ``coarse'' histogram weights $\hat{\mathbf{t}}, \hat{ \mathbf{w}}$ and using those sorted sampled distances as the endpoints of a set of $n$ intervals. This approach of using sampled points for use as interval endpoints can produce unusual results, as evidenced by Figure~\ref{fig:sample} --- it effectively ``erodes'' the coarse histogram, as the samples are unlikely to span the extent of each coarse histogram bin. For this reason we use a slightly modified resampling procedure: We sample $n$ sorted values from the coarse histogram, and then use the midpoints of each adjacent sample pair as the endpoints of our new set of ``fine'' intervals (and we reflect the first and last sample around the first and last endpoint to deal with boundary conditions). This change has little quantitative effect on rendering quality, but we found that it qualitatively reduced aliasing.

\paragraph{Background Colors.} NeRF and mip-NeRF assume a known background color, which is usually set to black or white. This often results in scene reconstructions in which the background is incorrectly represented as semi-transparent instead of opaque. These semi-transparent backgrounds may still allow for realistic view synthesis, but they tend to yield less meaningful mean or median ray termination distances, which results in less accurate depth maps. For this reason, when compositing a pixel color during training we draw a random RGB background color from $[0, 1]^3$ which encourages training to reconstruct a fully-opaque background of the scene (at test-time we set the background color to $(0.5, 0.5, 0.5)$). We use randomized backgrounds for our 360 dataset and the LLFF dataset, but for the Blender dataset we use the same fixed/known white background as in prior work.

\begin{figure}[t]
    \centering
    \begin{subfigure}{.49\linewidth}
      \centering
      \includegraphics[width=\textwidth]{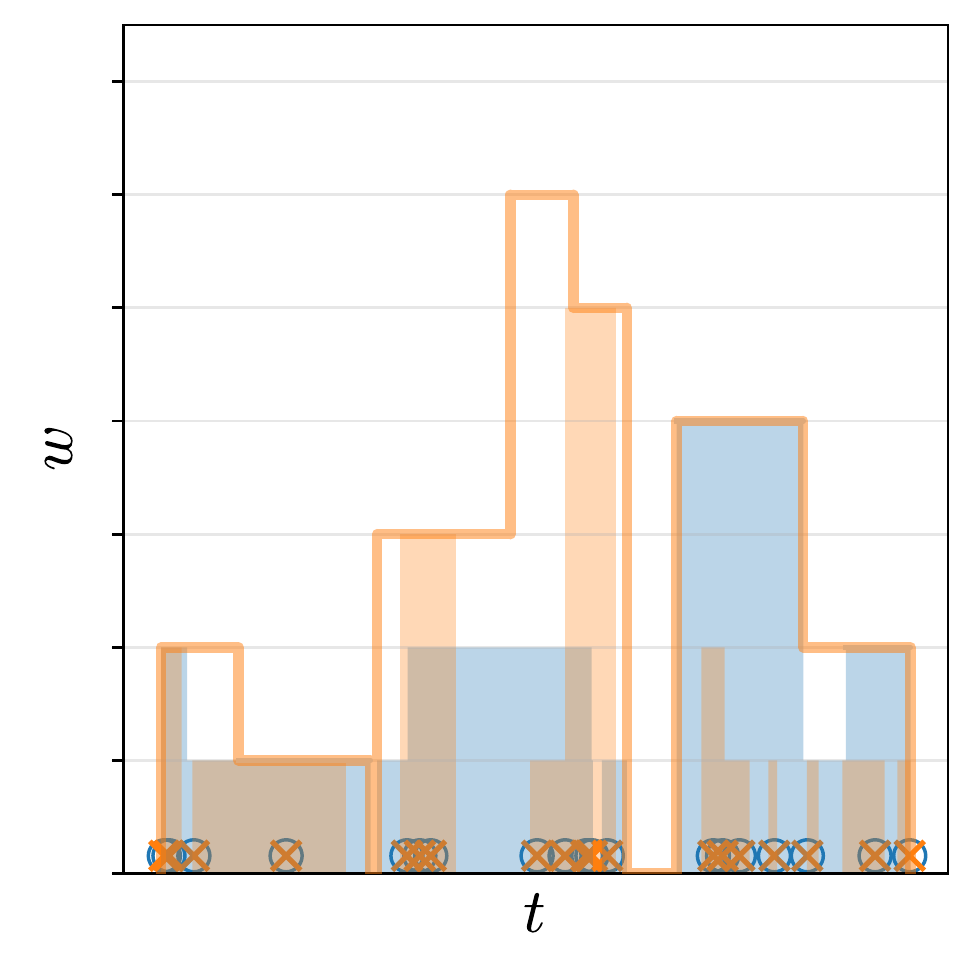}
      \caption{$\proploss\left(\mathbf{\distance}, \mathbf{w}, \proposal{\mathbf{\distance}}, \proposal{\mathbf{w}} \right) = 0$}
      \label{fig:inter1}
    \end{subfigure}
    \begin{subfigure}{.49\linewidth}
      \centering
      \includegraphics[width=\textwidth]{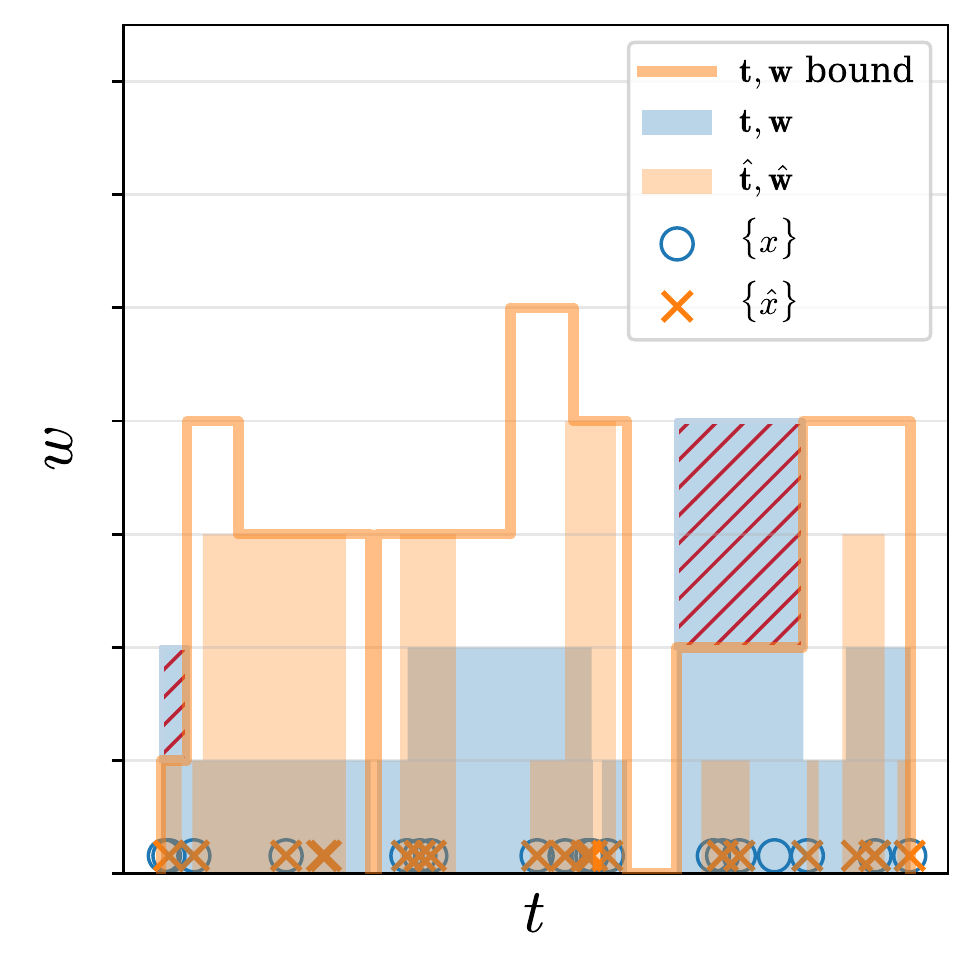}
      \caption{$\proploss\left(\mathbf{\distance}, \mathbf{w}, \proposal{\mathbf{\distance}}, \proposal{\mathbf{w}} \right) = 0.75$}
      \label{fig:inter2}
    \end{subfigure}
    \vspace{-0.05in}
    \caption{
    A visualization of the motivation behind $\proploss$, the loss used to train our proposal MLP to bound the weights emitted by our NeRF MLP.
    In both plots we have two different histograms $(\mathbf{t}, \mathbf{w})$ (shown in orange) and $(\hat{\mathbf{t}}, \hat{\mathbf{w}})$ (shown in blue) generated from points $\{ x \}$ and ${\{ \hat{x} \}}$ respectively, as well as a plot of the bound described in the paper. 
    (a) If $\{ x \} = {\{ \hat{x} \}}$, the bound implied by $(\hat{\mathbf{t}}, \hat{\mathbf{w}})$ is guaranteed to be an upper bound on $(\mathbf{t}, \mathbf{w})$, and our loss must be zero.
    (b) If $\{ x \} \neq {\{ \hat{x} \}}$ (in this case, only $16$ of $20$ points are shared between $\{ x \}$ and ${\{ \hat{x} \}}$) then $(\mathbf{t}, \mathbf{w})$ may exceed the upper bound implied by $(\hat{\mathbf{t}}, \hat{\mathbf{w}})$, and a loss may be incurred (shown in red). From this we see how minimizing $\proploss$ encourages the proposal weights $(\hat{\mathbf{t}}, \hat{\mathbf{w}})$ to describe the same distribution as the NeRF weights $(\mathbf{t}, \mathbf{w})$, despite their histogram bin endpoints being different.
    }
    \label{fig:interlevel}
\end{figure}

\section{Implementation Details}

The Jacobian $\mathbf{J}_{f}(\mathbf{\boldsymbol{\mu}})$ used by our Kalman-like reparameterization can be computed straightforwardly using most autodiff frameworks. A less expensive alternative (as it does not require the explicit construction of a Jacobian matrix) is to instead construct and apply a function whose application corresponds to matrix multiplication with $\mathbf{J}_{f}(\mathbf{\boldsymbol{\mu}})$. In Jax~\cite{jax2018github}, this can be accomplished using the \texttt{linearize} operator, and applying it twice in sequence to $\covmat$, with the dimensions of the covariance matrix transposed after each application.

\section{Proposal Supervision Visualization}

The loss used to supervise our proposal MLP is motivated by bounds that can be established between histograms of 1D data. The bound used by our loss is guaranteed to hold if two histograms are constructed from the same underlying ``true'' distribution of data. By minimizing any excess histogram mass that violates this bound, we can encourage two histograms with differently-spaced bin locations to be consistent with each other.  In Figure~\ref{fig:interlevel} we provide an illustration of this concept, and the supplemental video contains additional explanatory illustrations.

\section{Additional Results}

\paragraph{Our Dataset.}

We captured our dataset using two different mirrorless digital cameras. The outdoor scenes were captured with a Sony NEX C-3 equipped with a 18-55mm lens, using the widest possible zoom level. For the indoor scenes, we used a Fujifilm X100V camera with a fixed 22mm lens.
For each scene, we used the first camera location as a reference view, where we configured ISO, white balance, shutter speed, aperture size, and focus. We then kept these settings locked during capture, to limit the photometric variation between images of the same scene.
To further limit color harmonization issues, we captured the outdoor scene when the sky was overcast, making sure that the camera operator cast soft shadows that minimally affected the illumination in the scene. For the indoor scenes, we relied on large diffuse light sources (e.g. daylight reflecting off white walls) and avoided casting shadows onto the scene.

We captured between 100 and 330 images in each scene. This took between 1 and 20 minutes, depending on whether we used burst mode or not.
To obtain camera poses, we use the publicly available COLMAP software~\cite{schoenberger2016sfm}. We use shared intrinsics between all images in a scene, and calibrate using the OpenCV radial distortion model. Before training a NeRF, we use COLMAP to undistort the images, and downsample them to a resolution of 1.0 -- 1.6 megapixels using ImageMagick.
We use 1 in 8 of the input images as our test set, regularly subsampled to cover as many viewpoints as possible.

Post-capture, we apply a rigid transform and rescaling to COLMAP's reconstructed poses in order to better fit the captured scene content to our parameterization. In order to match the global coordinate frame to the capture pattern (assumed to be approximately circular rings orbiting a fixed point in space), we subtract the mean camera position and calculate the principal components of the recentered camera position vectors. We then use these three orthogonal vectors to form a new basis where the smallest principal component becomes the world-space ``up'' vector. After recentering all camera poses using this transformation, we rescale the camera positions such that they lie within the $[-1,1]^3$ cube. If the input poses lie approximately on a sphere, this usually causes them to lie within the uniformly parameterized region of space contained by the sphere of radius 1.

In Table~\ref{tab:scene_360_results} we show an expanded table of results for our dataset where we enumerate PSNRs, SSIMs, and LPIPS scores for each individual scene. Each technique's per-scene performance is roughly consistent with its average performance as reported in the main paper.

\begin{figure*}[t!]
    \centering
    \begin{tabular}{@{}c@{\,}|@{\,}c@{}}
        \includegraphics[width=0.49\linewidth]{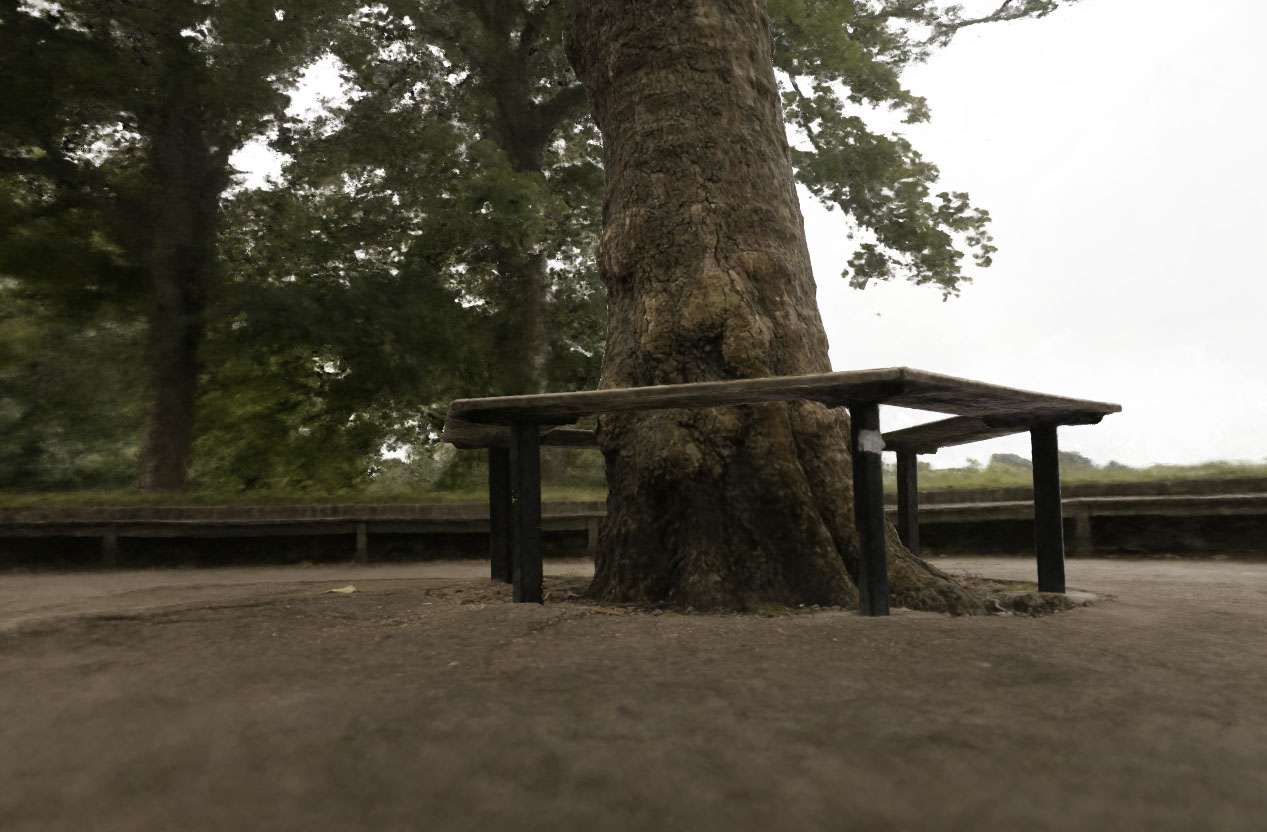} &
        \includegraphics[width=0.49\linewidth]{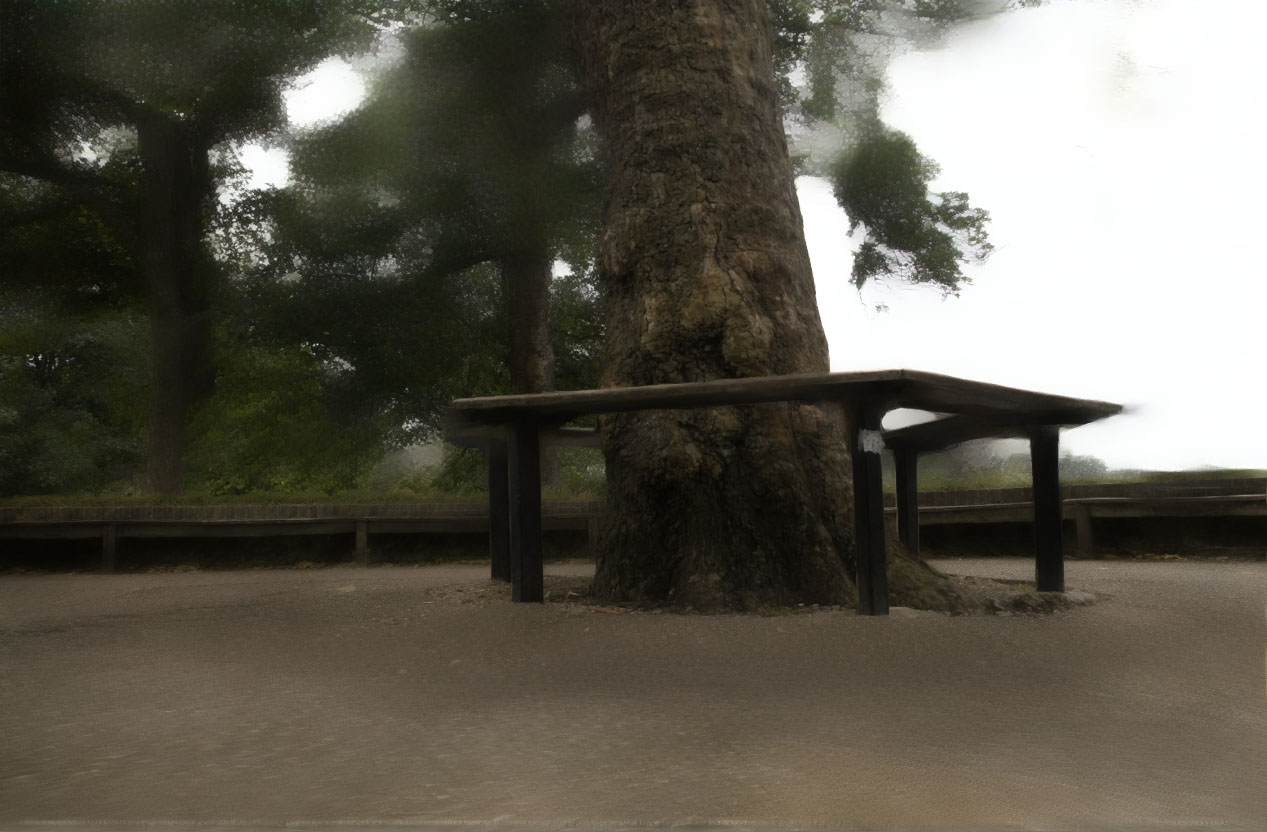} \\
        \includegraphics[width=0.49\linewidth]{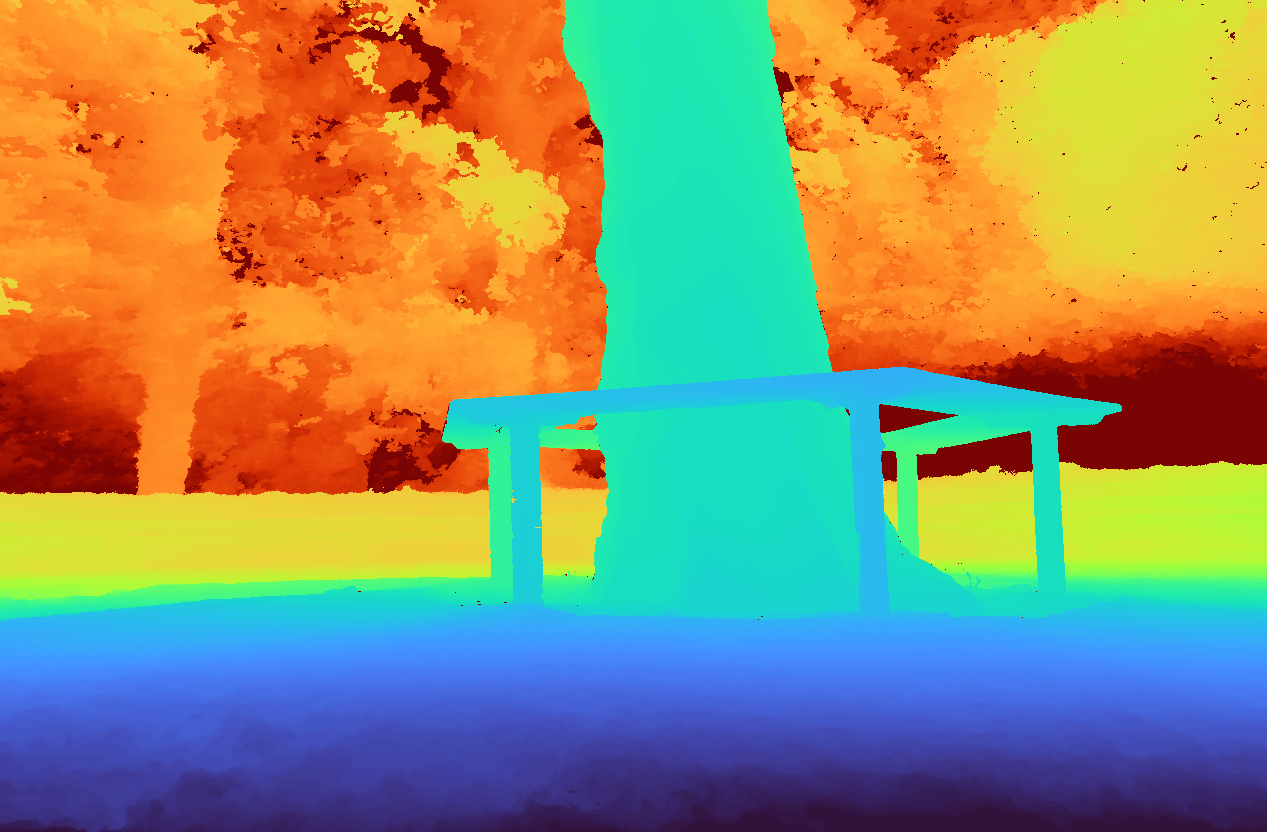} &
        \includegraphics[width=0.49\linewidth]{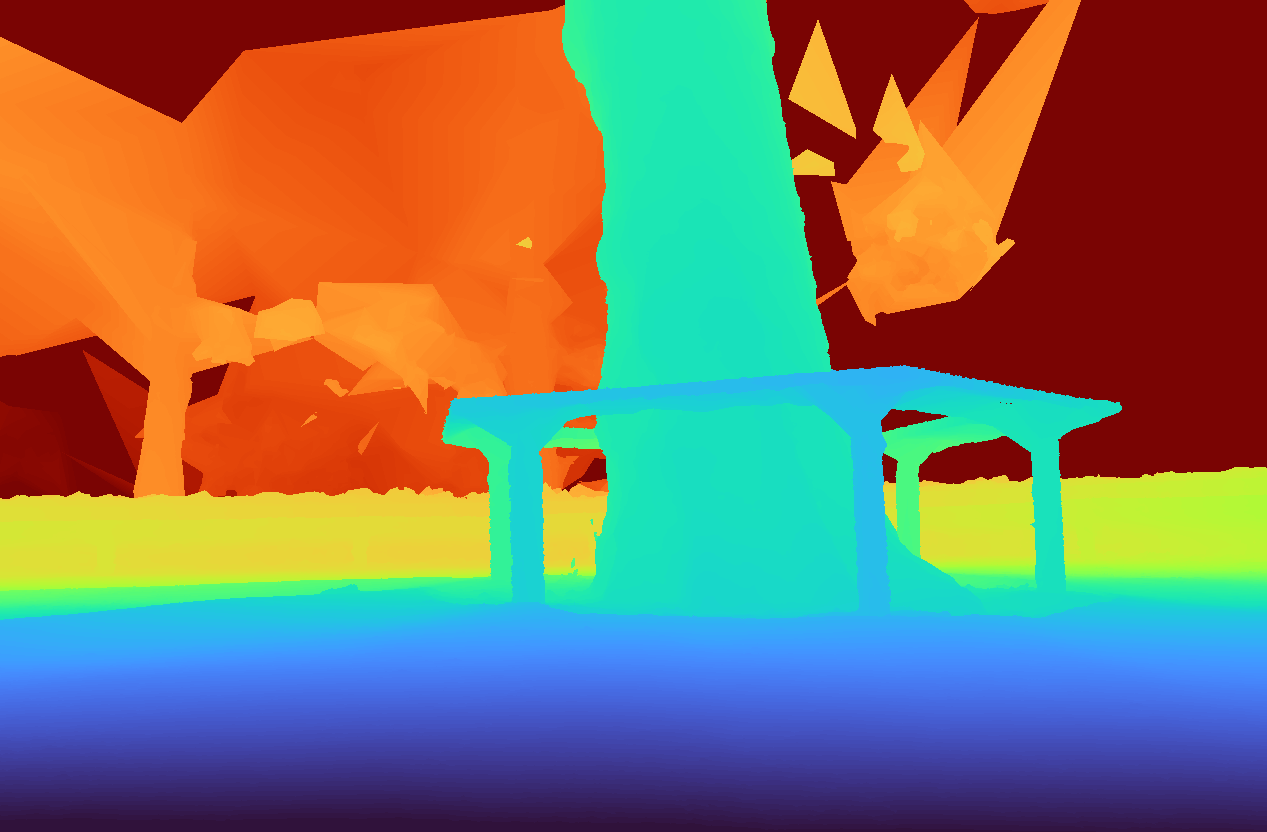} \\
        \small (a) Our Model, PSNR = \hl{16.67}, SSIM  = \hl{0.493}, LPIPS = 0.422 & \small (b) SVS \cite{riegler2021svs}, PSNR = 16.11, SSIM = 0.488, LPIPS = \hl{0.396} \\
    \end{tabular}
    \caption{
    A rendering from (a) our model, and (b) Stable View Synthesis~\cite{riegler2021svs} on a scene from our dataset. The PSNR, SSIM, and LPIPS metrics for \emph{this image} are shown in each subcaption. Despite SVS producing a blurry background, it achieves a lower LPIPS score, suggesting that this metric may be an unreliable signal in this setting. We also visualize (a) the depth map produced by our model alongside (b) the depth map produced by COLMAP~\cite{schoenberger2016sfm} which is used by SVS. The poor reconstruction quality of COLMAP in the distant trees may explain why SVS struggles with this scene.
    }
    \label{fig:lpips}
\end{figure*}

As discussed in the paper, Stable View Synthesis~\cite{riegler2021svs} is the only baseline model we evaluate against that outperforms our model on any metric, which is LPIPS~\cite{zhang2018unreasonable}. Upon visually inspecting the results of SVS on our dataset, we observed that LPIPS is often dramatically inconsistent with our own visual perception. See Figure~\ref{fig:lpips}, where we visualize the renderings (and depths) of our model versus SVS on one scene where SVS yielded a lower LPIPS metric than our model. Contrary to what the LPIPS scores indicate, our model's rendering is significantly more realistic and exhibits significantly fewer artifacts than SVS, particularly in the background of the scene. We believe this is due to SVS having been trained to minimize a perceptual loss that resembles LPIPS, causing it to produce results that are able to minimize LPIPS effectively despite being visually unsatisfying. This is consistent with recent work that has demonstrated vulnerabilities in LPIPS~\cite{kettunen2019lpips}.

\paragraph{NeRF's Blender Dataset.}
For completeness, in Table~\ref{tab:avg_blender_results} we evaluate our model on the Blender dataset from Mildenhall \etal~\cite{mildenhall2020}, for which mip-NeRF is the current state-of-the-art. This dataset consists entirely of small synthetic objects in front of a white background, unlike the large and unbounded scenes which motivated our model's design.
Because this task is easier than our 360 scenes, we use a simplified version of our model for the sake of speed: only one round of sampling, 128 samples for the proposal MLP, 32 samples for the NeRF MLP, a proposal MLP with 4 layers and 256 hidden units, a NeRF MLP with 8 layers and 256 (or 512) hidden units, axis-aligned IPE, MSE loss, and no distortion regularizer.
Our model is not designed to improve accuracy on these scenes, and as such our model's accuracy is comparable to mip-NeRF across all error metrics. However, we see that (due to our use of proposal networks) our model is significantly faster to train than mip-NeRF, and that this relative speedup increases as model capacity rises.

\begin{table}[h!]
    \centering
    \resizebox{\linewidth}{!}{
    \begin{tabular}{@{}lc|ccc|@{\,}c@{\,\,}|@{\,}r}
    & \# hidden & \!PSNR $\uparrow$\! & \!SSIM $\uparrow$\! & \!LPIPS $\downarrow$\! & Time (hrs) & \# Params \\ \hline
    mip-NeRF \cite{barron2021mipnerf} & \multirow{2}{*}{256} & \cellcolor{orange}33.09 & \cellcolor{yellow}0.961 & \cellcolor{yellow}0.043 & 2.89 & 0.61M \\
    Our Model & &                    32.96 &                    0.960 & \cellcolor{yellow}0.043 & 1.86 & 0.84M \\
    \hline 
    mip-NeRF \cite{barron2021mipnerf} & \multirow{2}{*}{512} & \cellcolor{yellow}33.03 & \cellcolor{red}0.964 & \cellcolor{red}0.037 & 7.03 & 2.27M \\
    Our Model & & \cellcolor{red}33.25 & \cellcolor{orange}0.962 & \cellcolor{orange}0.039 & 3.42 & 3.23M \\
    \end{tabular}
    }
    \vspace{-0.1in}
    \caption{
    Performance on the Blender dataset used in NeRF~\cite{mildenhall2020} as we vary the number of hidden units in the NeRF MLP.
    }
    \label{tab:avg_blender_results}
\end{table}

\paragraph{The LLFF Dataset.}
In Table~\ref{tab:avg_llff_results} we evaluate against mip-NeRF on the ``front-facing'' scenes presented in the LLFF paper~\cite{mildenhall2019llff}. We use the same model as in our Blender evaluation, except we do not disable our distortion regularizer and we use Charbonnier loss instead of MSE. Our model does not outperform mip-NeRF in terms of PSNR, but yields improved SSIM and LPIPs metrics and a significant speedup when the NeRF MLP is large.

\begin{table}[h!]
    \centering
    \resizebox{\linewidth}{!}{
    \begin{tabular}{@{}lc|ccc|@{\,}c@{\,\,}|@{\,}r}
    & \# hidden & \!PSNR $\uparrow$\! & \!SSIM $\uparrow$\! & \!LPIPS $\downarrow$\! & Time (hrs) & \# Params \\ \hline
mip-NeRF \cite{barron2021mipnerf} & \multirow{2}{*}{256}  & \cellcolor{orange}26.93 &                    0.830 &                    0.177 & 2.48 & 0.61M \\
Our Model &  &                    26.68 & \cellcolor{orange}0.847 & \cellcolor{yellow}0.150 & 2.39 & 0.84M \\ \hline
mip-NeRF \cite{barron2021mipnerf} & \multirow{2}{*}{512}  & \cellcolor{red}27.01 & \cellcolor{yellow}0.845 & \cellcolor{orange}0.148 & 6.15 & 2.27M \\
Our Model & & \cellcolor{yellow}26.86 & \cellcolor{red}0.858 & \cellcolor{red}0.128 & 3.84 & 3.23M
    \end{tabular}
    }
    \vspace{-0.1in}
    \caption{
    Performance on the dataset presented with LLFF~\cite{mildenhall2019llff} as we vary the number of hidden units in the NeRF MLP.
    }
    \label{tab:avg_llff_results}
\end{table}

\begin{figure}[b!]
    \centering
    \begin{tabular}{@{}c@{\,\,}c@{}}
        \includegraphics[width=0.49\linewidth]{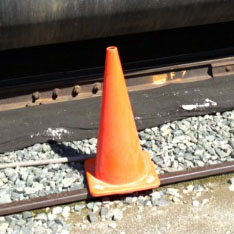} &
        \includegraphics[width=0.49\linewidth]{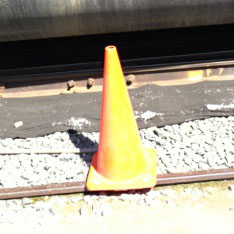}
    \end{tabular}
    \caption{
    Crops from two images taken from the ``Tanks and Temples'' dataset. The capture process used in acquiring this dataset seems to have allowed autoexposure and/or auto white balance to vary across images, which results in the same object having a different appearance across scenes. This issue partially motivated the construction of our own dataset, in which great care is taken to prevent such photometric variation.
    }
    \label{fig:tat_cone}
\end{figure}

\paragraph{The Tanks and Temples Dataset.}
The ``Tanks and Temples'' dataset is a popular dataset for 3D geometry and view synthesis tasks~\cite{Knapitsch2017}. It contains several scenes with a large central object with the camera moving around that object. At first glance this dataset may appear to be ideal for our purposes, but it has significant issues that motivated the construction of our own dataset. As shown in Figure~\ref{fig:tat_cone}, the photometric properties of the camera are not constant across each scene capture. We believe this is due to the camera's autoexposure or auto white balance being allowed to vary between images. Additionally, many images are overexposed, resulting in ``clipped'' RGB values (also visible in Figure~\ref{fig:tat_cone}). These issues make evaluation difficult, as measuring the accuracy of a view synthesis algorithm becomes an ill-posed task when faced with photometric variation --- which photometric condition should the model attempt to replicate? This challenge posed by ``in the wild'' images has been investigated by Martin-Brualla \etal~\cite{martinbrualla2020nerfw} who constructed specialized training and evaluation procedures for dealing with it. But we view this challenge as orthogonal to the challenges posed by the unbounded nature of a scene, hence the construction of our own dataset where our camera is photometrically fixed within each capture, and where scenes are chosen to minimize saturated pixels.

\begin{figure*}[t!]
    \centering
    \begin{tabular}{@{}l@{\,\,}|@{\,\,}c@{\,\,}c@{\,\,}c@{\,\,}c@{}}
        \rotatebox{90}{\scriptsize \quad\quad\quad\, SVS} & \includegraphics[width=0.24\linewidth]{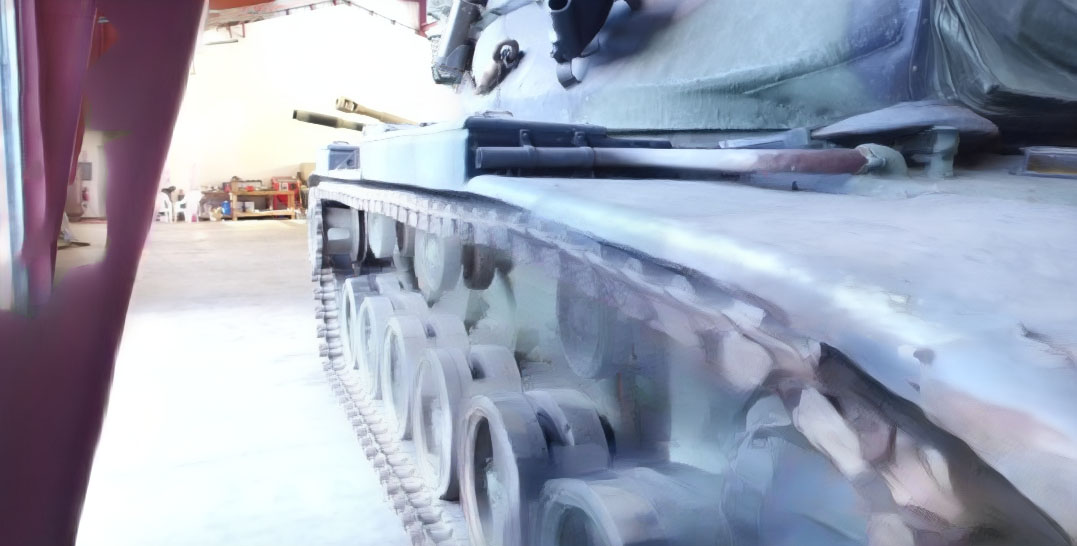} &
        \includegraphics[width=0.22\linewidth]{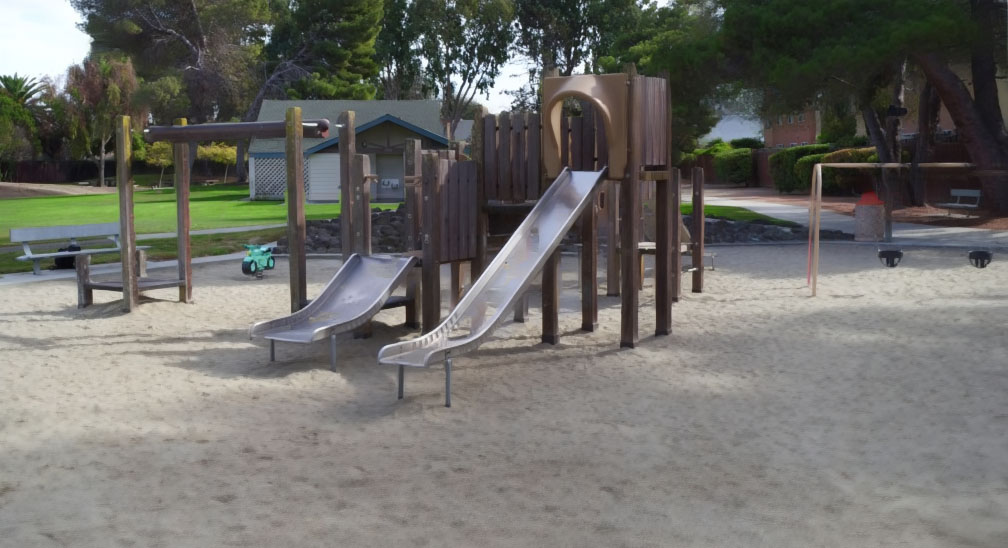} &
        \includegraphics[width=0.22\linewidth]{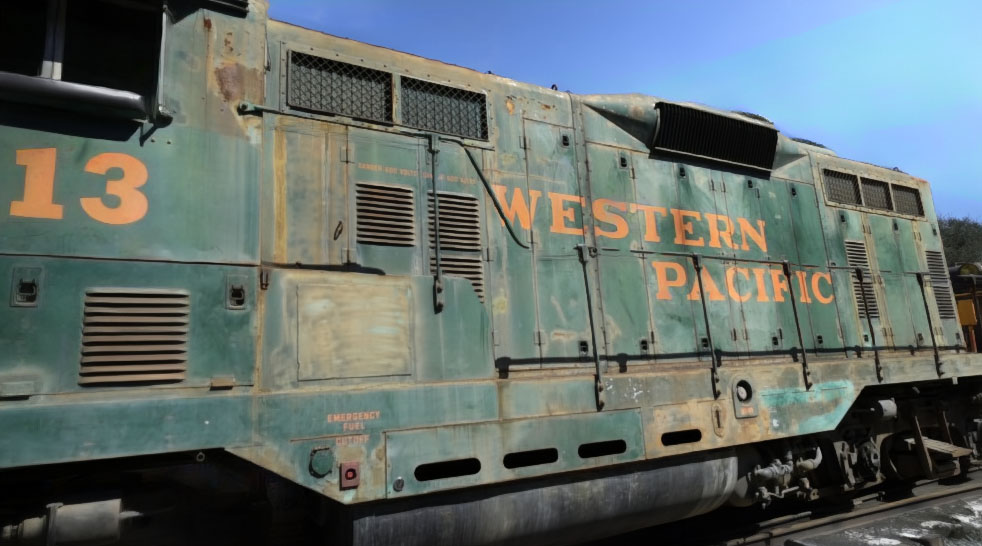} &
        \includegraphics[width=0.22\linewidth]{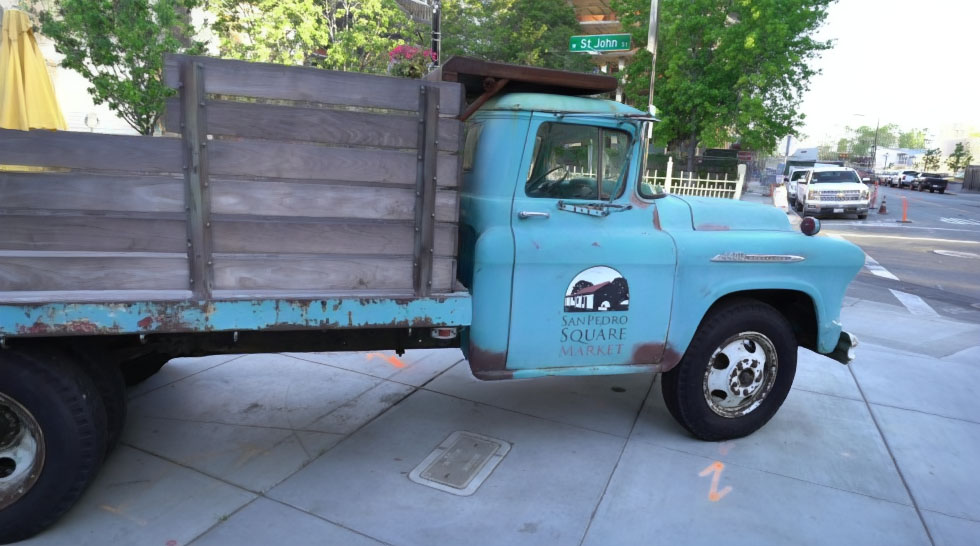} \\
        \rotatebox{90}{\scriptsize \, Our Model w/GLO } & \includegraphics[width=0.24\linewidth]{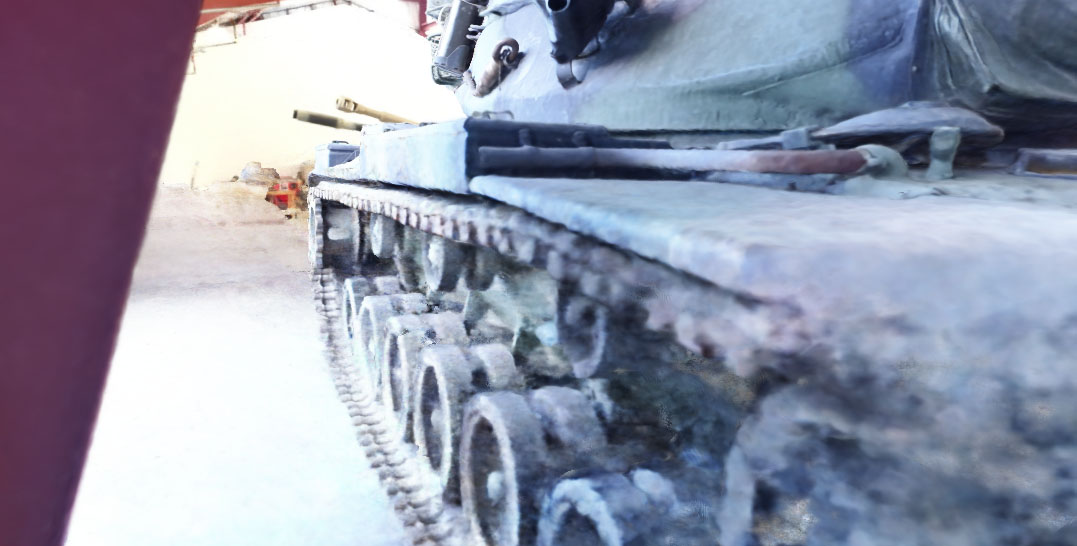} &
        \includegraphics[width=0.22\linewidth]{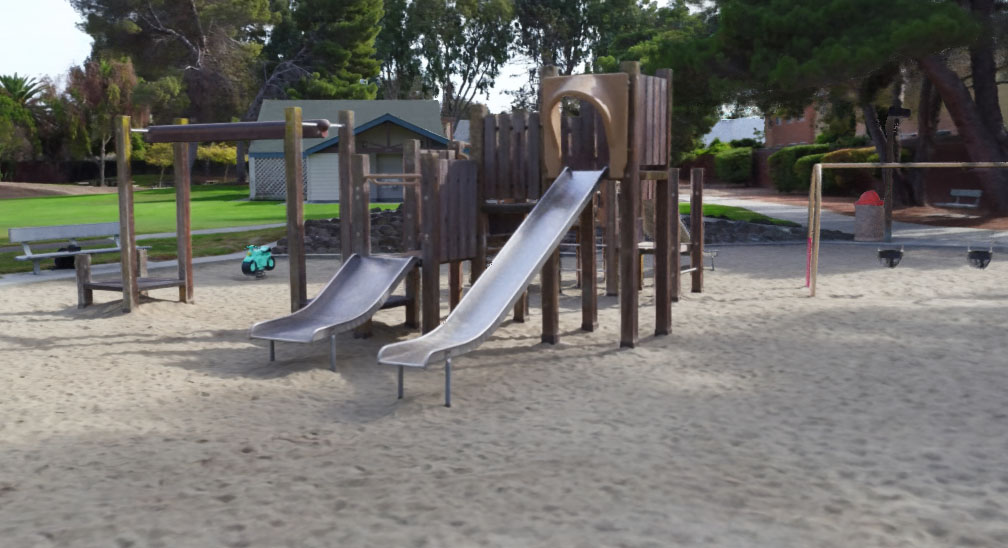} &
        \includegraphics[width=0.22\linewidth]{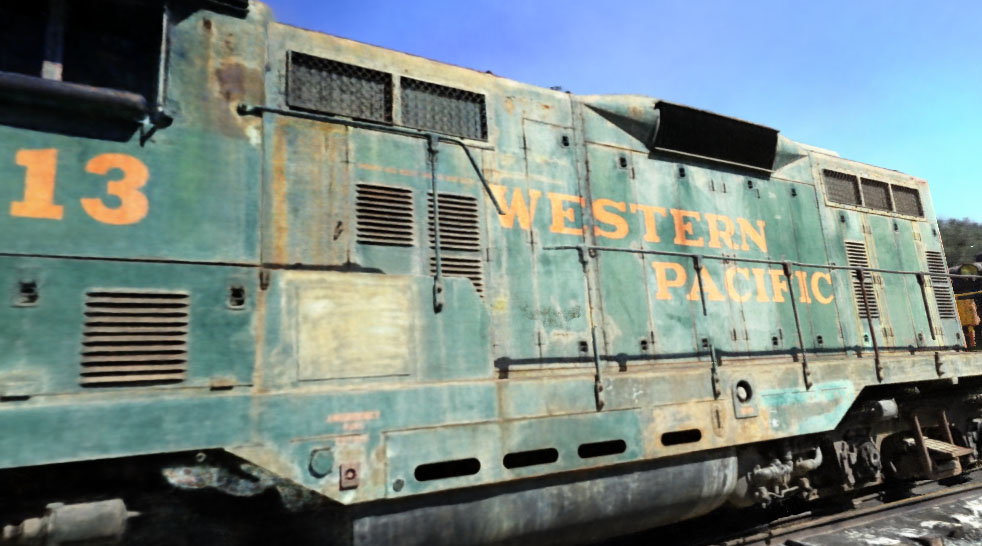} &
        \includegraphics[width=0.22\linewidth]{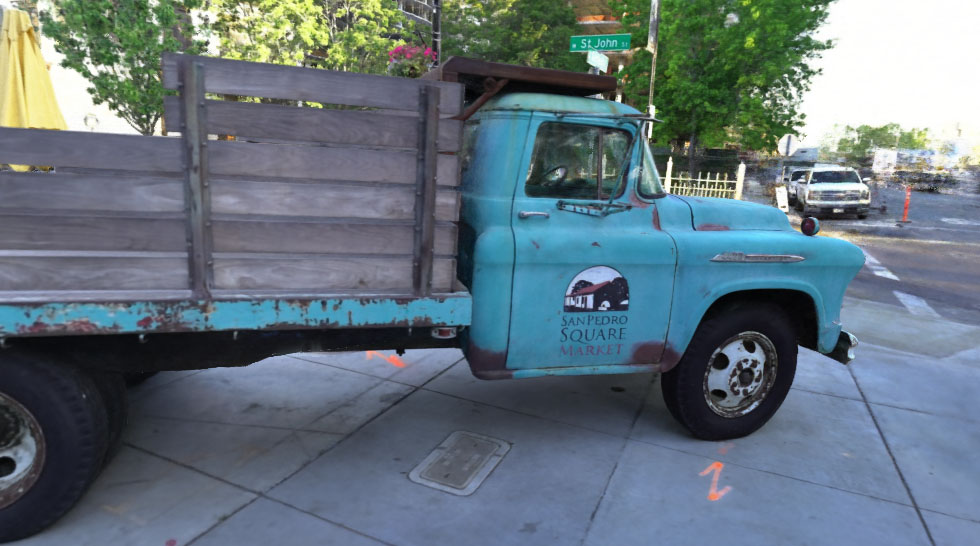} \\
        \hline
        & \scriptsize \textit{M60} & \scriptsize \textit{Playground} & \scriptsize \textit{Train} & \scriptsize \textit{Truck} \\
    \end{tabular}
    \caption{A visualization of our model with Stable View Synthesis~\cite{riegler2021svs} on scenes from the Tanks and Temples dataset~\cite{Knapitsch2017}. Image quality is roughly comparable across the two techniques, though our renderings exhibits different failure modes than SVS's in the absence of observations (as in \textit{M60}) and, because our model neutralizes most photometric variation during training, our renderings may have a different global brightness or color shift (as in \textit{Train}).
    }
    \label{fig:tat_results}
\end{figure*}

Despite the mismatch between this dataset and the goals of our work, we evaluated our model on this dataset against our NeRF-like baselines and against SVS (which is both the state of the art for this dataset as well as the most competitive baseline for our own dataset), the results of which are shown in Tables~\ref{tab:avg_tat_results} and \ref{tab:tat_perscene}, and visualized in Figure~\ref{fig:tat_results}. The metrics used elsewhere in this paper (PSNR, SSIM, and LPIPS) are difficult to draw meaningful conclusions from due to the aforementioned photometric variation. In particular, our top-performing ``w/GLO'' model variant performs quite poorly according to those metrics, because that model variant learns a per-image embedding for each scene and uses that embedding within the NeRF MLP when predicting color. When we evaluate this model variant at test-time, we set the embedding vector to $\mathbf{0}$. This gives us a pleasing looking reconstruction that roughly corresponds to the photometric average of all input cameras, and that is consistent across all images, which we believe to be a good goal for view synthesis. However, SVS (and to a lesser extent, the non-GLO NeRF baselines) do not behave this way, and instead attempt to ``explain away'' photometric variation due to the camera by modifying the brightness and color of the scene as a function of viewing direction. Effectively, SVS does not attempt to synthesize a view, it attempts to synthesize a view \emph{and} the most likely camera settings for that view. This motivated us to construct ``color corrected'' error metrics: before evaluating each metric we solve a per-image least squares problem that fits a quadratic polynomial expansion of the rendering's RGB values to the true image, while ignoring saturated pixels. This partially reduces the effect of photometric variation on this data, and yields results in which SVS and our model (with GLO) are roughly quantitatively comparable.

When using these color-corrected metrics, our model slightly outperforms SVS in terms of PSNR, but underperforms SVS on SSIM and LPIPS. With this in mind, it is worth reiterating the advantages that SVS has over our model on this benchmark: 1) SVS has been trained on the training set of this dataset, while our model does not use that external training data --- and indeed uses no external training data at all. 2) SVS relies on a proxy geometry produced by an external system (and may fail when that geometry is incorrect), while we use no proxy geometry and in fact produce high-quality depth maps ourselves. 3) SVS has been trained with a perceptual loss, while our model is trained using only a per-pixel loss on RGB. 4) Our model is extremely compact, and requires only 10 million parameters to perform view synthesis, while SVS requires multiple large CNNs and access to all training images (because it operates by blending training images together) to render novel views.

From Table~\ref{tab:tat_perscene} we see that our model's improvement over SVS is most significant on the \textit{playground} scene. Notably, this is the only test-set scene that mostly consists of natural content, while the other three scenes predominately feature large vehicles. We speculate that SVS may be better-suited to large piecewise planar objects (which makes sense, given SVS's reliance on a proxy geometry that is itself a piecewise planar mesh) while ours may be better suited to scenes that contain natural content (trees, grass, flowers, etc).

\section{Potential Negative Impact}

The broad use of neural rendering techniques carries with it several potential negative societal impacts. NeRF-like models have recently been incorporated into generative modeling approaches~\cite{gu2021stylenerf}, and generative modeling techniques can be used to synthesize ``deep fakes'' that could be used to mislead people. Though our work does not directly concern generative modeling and instead aims to reconstruct accurate physical models of a scene from which new views can be generated, our contributions may be useful for generative approaches that build on NeRF.

The ability to reconstruct accurate models of a scene from photographs may have modest potential negative impacts. Our technique could conceivably be used to construct a surveillance system, and such a system could have negative impact if used negligently or maliciously. Additionally, our system could be used to generate visual effects (a task that is currently labor intensive) and as such it may negatively affect job opportunities for artists.

Training a NeRF is computationally demanding, and requires multiple hours of optimization on an accelerator (though test-time rendering can be accelerated significantly~\cite{hedman2021snerg}). This expensive training requires energy, and this may be of concern if that energy was produced in a way that damages the climate.

\sethlcolor{yellow}

\begin{table*}[p]
    \centering
    \small
    \begin{tabular}{@{}l|ccccc|cccc}
    \multicolumn{1}{c}{\,} & \multicolumn{9}{c}{PSNR} \\
    & \multicolumn{5}{c|}{Outdoor} & \multicolumn{4}{c}{Indoor} \\
    \input{tables/360_per_psnr} \\
    \multicolumn{10}{c}{} \\
    \multicolumn{1}{c}{\,} & \multicolumn{9}{c}{SSIM} \\
    & \multicolumn{5}{c|}{Outdoor} & \multicolumn{4}{c}{Indoor} \\
    \input{tables/360_per_ssim} \\
    \multicolumn{10}{c}{} \\
    \multicolumn{1}{c}{\,} & \multicolumn{9}{c}{LPIPS} \\
     & \multicolumn{5}{c|}{Outdoor} & \multicolumn{4}{c}{Indoor} \\
    \input{tables/360_per_lpips}
    \end{tabular}
    \vspace{-0.1in}
    \caption{
    Here we present an expanded version of Table 1 from the main paper, where we evaluate our model and multiple NeRF and non-NeRF baselines on our new dataset, but where we report metrics for each scene separately. Though some scenes are more challenging than others, the overall ranking of all techniques on each scene is generally consistent with the ranking suggested by the average metrics.
    }
    \label{tab:scene_360_results}
\end{table*}

\begin{table*}[b!]
    \centering
    \small
    \begin{tabular}{@{}l|ccc|ccc|c@{\,\,}|@{\,}r}
    \multicolumn{4}{c|}{} & \multicolumn{3}{c|}{Color Corrected} \\
    & \!PSNR $\uparrow$\! & \!SSIM $\uparrow$\! & \!LPIPS $\downarrow$\! & \!PSNR $\uparrow$\! & \!SSIM $\uparrow$\! & \!LPIPS $\downarrow$\! & Time (hrs) & \# Params \\ \hline
    \input{tables/tat.tex}
    \end{tabular}
    \vspace{-0.1in}
    \caption{
    The average performance of our model and all NeRF baselines, as well as the top-performing non-NeRF baseline on our own dataset (Stable View Synthesis), on the ``Tanks and Temples'' dataset~\cite{Knapitsch2017}. This dataset exhibits significant photometric variation across images (see Figure~\ref{fig:tat_cone}), making it ill-suited to our goals. To partially ameliorate this we present additional ``color corrected'' metrics, in which this photometric variation has been ameliorated. Our model outperforms all NeRF baselines, but is slightly outperformed by SVS (which was designed for this dataset, and which was trained on the training set of this dataset), though this appears to be partially due to SVS being better able to predict the photometric variation of this dataset, while the ``w/ GLO'' variant of our model learns to be invariant to that photometric variation.
    \label{tab:avg_tat_results}
    }
    \centering
    \small
    \begin{tabular}{@{}l|cccc}
    \multicolumn{1}{c}{} & \multicolumn{4}{c}{Color Corrected PSNR} \\
    \input{tables/tat_per_psnr} \\
    \multicolumn{5}{c}{}  \\
    \multicolumn{1}{c}{} & \multicolumn{4}{c}{Color Corrected SSIM} \\
    \input{tables/tat_per_ssim} \\
    \multicolumn{5}{c}{}  \\
    \multicolumn{1}{c}{} & \multicolumn{4}{c}{Color Corrected LPIPS} \\
    \input{tables/tat_per_lpips}
    \end{tabular}
    \vspace{-0.1in}
    \caption{
    Performance on each individual test-set scene from the ``Tanks and Temples'' dataset, using our color corrected metrics.
    \label{tab:tat_perscene}
    }
\end{table*}

%% file: tables/360_per_psnr.tex
 & \textit{bicycle}& \textit{flowers}& \textit{garden}& \textit{stump}& \textit{treehill}& \textit{room}& \textit{counter}& \textit{kitchen}& \textit{bonsai}\\ \hline 
NeRF \cite{mildenhall2020, jaxnerf2020github}& 21.76 & 19.40 & 23.11 & 21.73 & 21.28 & 28.56 & 25.67 & 26.31 & 26.81 \\
NeRF w/ DONeRF~\cite{donerf} param. & 21.67 & 19.48 & 23.29 & 23.38 & 21.70 & 28.28 & 25.74 & 25.42 & 27.32 \\
mip-NeRF \cite{barron2021mipnerf}& 21.69 & 19.31 & 23.16 & 23.10 & 21.21 & 28.73 & 25.59 & 26.47 & 27.13 \\
NeRF++ \cite{kaizhang2020}& 22.64 & 20.31 & 24.32 & 24.34 & \cellcolor{yellow}22.20 & 28.87 & 26.38 & 27.80 & 29.15 \\
Deep Blending~\cite{hedman2018deep}& 21.09 & 18.13 & 23.61 & 24.08 & 20.80 & 27.20 & 26.28 & 25.02 & 27.08 \\
Point-Based Neural Rendering~\cite{kopanas2021point}& 21.64 & 19.28 & 22.50 & 23.90 & 20.98 & 26.99 & 25.23 & 24.47 & 28.42 \\
Stable View Synthesis~\cite{riegler2021svs}& 22.79 & 20.15 & \cellcolor{orange}25.99 & 24.39 & 21.72 & 28.93 & 26.40 & 28.49 & 29.07 \\
mip-NeRF~\cite{barron2021mipnerf} w/bigger MLP & 22.90 & 20.79 & 25.85 & 23.64 & 21.71 & \cellcolor{orange}30.67 & \cellcolor{orange}28.61 & \cellcolor{yellow}29.95 & \cellcolor{orange}31.59 \\
NeRF++~\cite{kaizhang2020} w/bigger MLPs & \cellcolor{yellow}23.75 & \cellcolor{yellow}21.11 & \cellcolor{yellow}25.91 & \cellcolor{yellow}25.48 & \cellcolor{orange}22.77 & \cellcolor{yellow}30.13 & 27.79 & 29.85 & \cellcolor{yellow}30.68 \\
\hline 
Our Model & \cellcolor{red}24.37 & \cellcolor{red}21.73 & \cellcolor{red}26.98 & \cellcolor{red}26.40 & \cellcolor{red}22.87 & \cellcolor{red}31.63 & \cellcolor{red}29.55 & \cellcolor{red}32.23 & \cellcolor{red}33.46 \\
Our Model w/GLO & \cellcolor{orange}23.95 & \cellcolor{orange}21.60 & 25.09 & \cellcolor{orange}25.98 & 21.99 & 28.24 & \cellcolor{yellow}28.40 & \cellcolor{orange}30.81 & 30.27 \\

%% file: tables/360_per_ssim.tex
 & \textit{bicycle}& \textit{flowers}& \textit{garden}& \textit{stump}& \textit{treehill}& \textit{room}& \textit{counter}& \textit{kitchen}& \textit{bonsai}\\ \hline 
NeRF \cite{mildenhall2020, jaxnerf2020github}& 0.455 & 0.376 & 0.546 & 0.453 & 0.459 & 0.843 & 0.775 & 0.749 & 0.792 \\
NeRF w/ DONeRF~\cite{donerf} param. & 0.454 & 0.379 & 0.542 & 0.522 & 0.461 & 0.841 & 0.776 & 0.678 & 0.813 \\
mip-NeRF \cite{barron2021mipnerf}& 0.454 & 0.373 & 0.543 & 0.517 & 0.466 & 0.851 & 0.779 & 0.745 & 0.818 \\
NeRF++ \cite{kaizhang2020}& 0.526 & 0.453 & 0.635 & 0.594 & 0.530 & 0.852 & 0.802 & 0.816 & 0.876 \\
Deep Blending~\cite{hedman2018deep}& 0.466 & 0.320 & 0.675 & 0.634 & 0.523 & 0.868 & 0.856 & 0.768 & 0.883 \\
Point-Based Neural Rendering~\cite{kopanas2021point}& 0.608 & 0.487 & 0.735 & 0.651 & 0.579 & 0.887 & 0.868 & 0.876 & 0.919 \\
Stable View Synthesis~\cite{riegler2021svs}& \cellcolor{yellow}0.663 & \cellcolor{yellow}0.541 & \cellcolor{red}0.818 & 0.683 & \cellcolor{yellow}0.606 & \cellcolor{yellow}0.905 & \cellcolor{yellow}0.886 & \cellcolor{yellow}0.910 & 0.925 \\
mip-NeRF~\cite{barron2021mipnerf} w/bigger MLP & 0.612 & 0.514 & 0.777 & 0.643 & 0.577 & 0.903 & 0.877 & 0.902 & \cellcolor{yellow}0.928 \\
NeRF++~\cite{kaizhang2020} w/bigger MLPs & 0.630 & 0.533 & 0.761 & \cellcolor{yellow}0.687 & 0.597 & 0.883 & 0.857 & 0.888 & 0.913 \\
\hline 
Our Model & \cellcolor{orange}0.685 & \cellcolor{red}0.583 & \cellcolor{orange}0.813 & \cellcolor{orange}0.744 & \cellcolor{red}0.632 & \cellcolor{red}0.913 & \cellcolor{red}0.894 & \cellcolor{red}0.920 & \cellcolor{red}0.941 \\
Our Model w/GLO & \cellcolor{red}0.687 & \cellcolor{orange}0.582 & \cellcolor{yellow}0.800 & \cellcolor{red}0.745 & \cellcolor{orange}0.619 & \cellcolor{orange}0.907 & \cellcolor{orange}0.890 & \cellcolor{orange}0.916 & \cellcolor{orange}0.932 \\

%% file: tables/360_per_lpips.tex
 & \textit{bicycle}& \textit{flowers}& \textit{garden}& \textit{stump}& \textit{treehill}& \textit{room}& \textit{counter}& \textit{kitchen}& \textit{bonsai}\\ \hline 
NeRF \cite{mildenhall2020, jaxnerf2020github}& 0.536 & 0.529 & 0.415 & 0.551 & 0.546 & 0.353 & 0.394 & 0.335 & 0.398 \\
NeRF w/ DONeRF~\cite{donerf} param. & 0.542 & 0.539 & 0.436 & 0.492 & 0.545 & 0.368 & 0.394 & 0.410 & 0.368 \\
mip-NeRF \cite{barron2021mipnerf}& 0.541 & 0.535 & 0.422 & 0.490 & 0.538 & 0.346 & 0.390 & 0.336 & 0.370 \\
NeRF++ \cite{kaizhang2020}& 0.455 & 0.466 & 0.331 & 0.416 & 0.466 & 0.335 & 0.351 & 0.260 & 0.291 \\
Deep Blending~\cite{hedman2018deep}& 0.377 & 0.476 & 0.231 & 0.351 & 0.383 & 0.266 & 0.258 & 0.246 & 0.275 \\
Point-Based Neural Rendering~\cite{kopanas2021point}& 0.313 & 0.372 & 0.197 & 0.303 & \cellcolor{orange}0.325 & 0.216 & 0.209 & 0.160 & \cellcolor{yellow}0.178 \\
Stable View Synthesis~\cite{riegler2021svs}& \cellcolor{red}0.243 & \cellcolor{red}0.317 & \cellcolor{red}0.137 & \cellcolor{yellow}0.281 & \cellcolor{red}0.286 & \cellcolor{red}0.182 & \cellcolor{red}0.168 & \cellcolor{red}0.125 & \cellcolor{red}0.164 \\
mip-NeRF~\cite{barron2021mipnerf} w/bigger MLP & 0.372 & 0.407 & 0.205 & 0.357 & 0.401 & 0.229 & 0.239 & 0.152 & 0.204 \\
NeRF++~\cite{kaizhang2020} w/bigger MLPs & 0.356 & 0.395 & 0.223 & 0.328 & 0.386 & 0.270 & 0.270 & 0.177 & 0.230 \\
\hline 
Our Model & \cellcolor{yellow}0.301 & \cellcolor{yellow}0.344 & \cellcolor{orange}0.170 & \cellcolor{orange}0.261 & 0.339 & \cellcolor{yellow}0.211 & \cellcolor{orange}0.204 & \cellcolor{orange}0.127 & \cellcolor{orange}0.176 \\
Our Model w/GLO & \cellcolor{orange}0.296 & \cellcolor{orange}0.343 & \cellcolor{yellow}0.173 & \cellcolor{red}0.258 & \cellcolor{yellow}0.338 & \cellcolor{orange}0.208 & \cellcolor{yellow}0.206 & \cellcolor{yellow}0.129 & 0.182 \\

%% file: tables/tat.tex
NeRF \cite{mildenhall2020, jaxnerf2020github}& 18.72 & 0.609 & 0.473 & 19.67 & 0.616 & 0.473 & 4.15 & 1.5M \\
NeRF w/ DONeRF~\cite{donerf} param. & 18.85 & 0.618 & 0.477 & 20.00 & 0.624 & 0.477 & 4.70 & 1.4M \\
mip-NeRF \cite{barron2021mipnerf}& 18.86 & 0.620 & 0.463 & 19.93 & 0.625 & 0.464 & 3.23 & 0.7M \\
NeRF++ \cite{kaizhang2020}& 19.32 & 0.647 & 0.425 & 20.52 & 0.652 & 0.427 & 9.71 & 2.4M \\
mip-NeRF~\cite{barron2021mipnerf} w/bigger MLP & \cellcolor{yellow}19.85 & 0.697 & 0.340 & 21.09 & 0.702 & 0.343 & 22.75 & 9.0M \\
NeRF++~\cite{kaizhang2020} w/bigger MLPs & 19.83 & 0.693 & 0.358 & 21.15 & 0.697 & 0.362 & 19.94 & 9.0M \\
Stable View Synthesis~\cite{riegler2021svs}& \cellcolor{red}21.13 & \cellcolor{red}0.777 & \cellcolor{red}0.209 & \cellcolor{orange}22.76 & \cellcolor{red}0.778 & \cellcolor{red}0.216 & - & - \\
\hline 
Our Model & \cellcolor{orange}20.52 & \cellcolor{orange}0.734 & \cellcolor{yellow}0.301 & \cellcolor{yellow}21.98 & \cellcolor{yellow}0.737 & \cellcolor{yellow}0.304 & 6.61 & 9.0M \\
Our Model w/GLO & 19.65 & \cellcolor{yellow}0.731 & \cellcolor{orange}0.280 & \cellcolor{red}22.78 & \cellcolor{orange}0.761 & \cellcolor{orange}0.272 & 7.09 & 9.0M \\

%% file: tables/tat_per_psnr.tex
 & \textit{M60}& \textit{Playground}& \textit{Train}& \textit{Truck}\\ \hline 
NeRF \cite{mildenhall2020, jaxnerf2020github}& 17.59 & 21.72 & 19.17 & 20.21 \\
NeRF w/ DONeRF~\cite{donerf} param. & 17.31 & 23.13 & 18.76 & 20.81 \\
mip-NeRF \cite{barron2021mipnerf}& 17.58 & 22.21 & 19.42 & 20.50 \\
NeRF++ \cite{kaizhang2020}& 18.09 & 23.05 & 19.50 & 21.44 \\
mip-NeRF~\cite{barron2021mipnerf} w/bigger MLP & 19.14 & 23.65 & 19.82 & 21.74 \\
NeRF++~\cite{kaizhang2020} w/bigger MLPs & 18.81 & 24.01 & \cellcolor{yellow}19.84 & 21.94 \\
Stable View Synthesis~\cite{riegler2021svs}& \cellcolor{red}19.94 & \cellcolor{yellow}25.50 & \cellcolor{orange}21.76 & \cellcolor{orange}23.85 \\
\hline 
Our Model & \cellcolor{yellow}19.28 & \cellcolor{orange}26.41 & 18.23 & \cellcolor{red}24.01 \\
Our Model w/GLO & \cellcolor{orange}19.50 & \cellcolor{red}27.00 & \cellcolor{red}22.15 & \cellcolor{yellow}22.48 \\

%% file: tables/tat_per_ssim.tex
 & \textit{M60}& \textit{Playground}& \textit{Train}& \textit{Truck}\\ \hline 
NeRF \cite{mildenhall2020, jaxnerf2020github}& 0.619 & 0.624 & 0.575 & 0.646 \\
NeRF w/ DONeRF~\cite{donerf} param. & 0.622 & 0.659 & 0.559 & 0.657 \\
mip-NeRF \cite{barron2021mipnerf}& 0.629 & 0.638 & 0.582 & 0.650 \\
NeRF++ \cite{kaizhang2020}& 0.644 & 0.676 & 0.586 & 0.704 \\
mip-NeRF~\cite{barron2021mipnerf} w/bigger MLP & 0.694 & 0.726 & \cellcolor{yellow}0.642 & 0.747 \\
NeRF++~\cite{kaizhang2020} w/bigger MLPs & 0.682 & 0.724 & 0.630 & 0.751 \\
Stable View Synthesis~\cite{riegler2021svs}& \cellcolor{red}0.756 & \cellcolor{orange}0.788 & \cellcolor{red}0.731 & \cellcolor{red}0.836 \\
\hline 
Our Model & \cellcolor{yellow}0.714 & \cellcolor{yellow}0.781 & 0.635 & \cellcolor{orange}0.818 \\
Our Model w/GLO & \cellcolor{orange}0.720 & \cellcolor{red}0.798 & \cellcolor{orange}0.723 & \cellcolor{yellow}0.804 \\

%% file: tables/tat_per_lpips.tex
 & \textit{M60}& \textit{Playground}& \textit{Train}& \textit{Truck}\\ \hline 
NeRF \cite{mildenhall2020, jaxnerf2020github}& 0.466 & 0.473 & 0.493 & 0.458 \\
NeRF w/ DONeRF~\cite{donerf} param. & 0.466 & 0.458 & 0.514 & 0.468 \\
mip-NeRF \cite{barron2021mipnerf}& 0.462 & 0.461 & 0.483 & 0.449 \\
NeRF++ \cite{kaizhang2020}& 0.432 & 0.418 & 0.473 & 0.387 \\
mip-NeRF~\cite{barron2021mipnerf} w/bigger MLP & 0.367 & 0.330 & \cellcolor{yellow}0.379 & 0.296 \\
NeRF++~\cite{kaizhang2020} w/bigger MLPs & 0.383 & 0.348 & 0.409 & 0.308 \\
Stable View Synthesis~\cite{riegler2021svs}& \cellcolor{red}0.251 & \cellcolor{red}0.212 & \cellcolor{red}0.247 & \cellcolor{red}0.152 \\
\hline 
Our Model & \cellcolor{yellow}0.341 & \cellcolor{yellow}0.264 & 0.389 & \cellcolor{orange}0.223 \\
Our Model w/GLO & \cellcolor{orange}0.330 & \cellcolor{orange}0.246 & \cellcolor{orange}0.284 & \cellcolor{yellow}0.228 \\

%% file: arxiv.bbl
\begin{thebibliography}{10}\itemsep=-1pt

\bibitem{arandjelovic2021nerf}
Relja Arandjelovi{\'c} and Andrew Zisserman.
\newblock {NeRF} in detail: Learning to sample for view synthesis.
\newblock {\em arXiv:2106.05264}, 2021.

\bibitem{attal2020}
Benjamin Attal, Selena Ling, Aaron Gokaslan, Christian Richardt, and James
  Tompkin.
\newblock {MatryODShka}: Real-time {6DoF} video view synthesis using
  multi-sphere images.
\newblock {\em ECCV}, 2020.

\bibitem{barron2021mipnerf}
Jonathan~T. Barron, Ben Mildenhall, Matthew Tancik, Peter Hedman, Ricardo
  Martin-Brualla, and Pratul~P. Srinivasan.
\newblock {Mip-NeRF: A Multiscale Representation for Anti-Aliasing Neural
  Radiance Fields}.
\newblock {\em ICCV}, 2021.

\bibitem{OmniPhotos}
Tobias Bertel, Mingze Yuan, Reuben Lindroos, and Christian Richardt.
\newblock {OmniPhotos}: Casual 360° {VR} photography.
\newblock {\em ACM Transactions on Graphics}, 2020.

\bibitem{Blinn1991}
J.F. Blinn.
\newblock A trip down the graphics pipeline: pixel coordinates.
\newblock {\em IEEE Computer Graphics and Applications}, 1991.

\bibitem{bojanowski2018optimizing}
Piotr Bojanowski, Armand Joulin, David Lopez-Pas, and Arthur Szlam.
\newblock Optimizing the latent space of generative networks.
\newblock {\em ICML}, 2018.

\bibitem{sibr2020}
Sebastien Bonopera, Peter Hedman, Jerome Esnault, Siddhant Prakash, Simon
  Rodriguez, Theo Thonat, Mehdi Benadel, Gaurav Chaurasia, Julien Philip, and
  George Drettakis.
\newblock sibr: A system for image based rendering, 2020.

\bibitem{jax2018github}
James Bradbury, Roy Frostig, Peter Hawkins, Matthew~James Johnson, Chris Leary,
  Dougal Maclaurin, George Necula, Adam Paszke, Jake Vander{P}las, Skye
  Wanderman-{M}ilne, and Qiao Zhang.
\newblock {JAX}: composable transformations of {P}ython+{N}um{P}y programs,
  2018.
\newblock \url{http://github.com/google/jax}.

\bibitem{broxton2020immersive}
Michael Broxton, John Flynn, Ryan Overbeck, Daniel Erickson, Peter Hedman,
  Matthew DuVall, Jason Dourgarian, Jay Busch, Matt Whalen, and Paul Debevec.
\newblock Immersive light field video with a layered mesh representation.
\newblock {\em SIGGRAPH}, 2020.

\bibitem{charbonnier1994two}
Pierre Charbonnier, Laure Blanc-Feraud, Gilles Aubert, and Michel Barlaud.
\newblock Two deterministic half-quadratic regularization algorithms for
  computed imaging.
\newblock {\em International Conference on Image Processing}, 1994.

\bibitem{crow1984summed}
Franklin~C Crow.
\newblock Summed-area tables for texture mapping.
\newblock {\em SIGGRAPH}, 1984.

\bibitem{dalal2005histograms}
Navneet Dalal and Bill Triggs.
\newblock Histograms of oriented gradients for human detection.
\newblock {\em CVPR}, 2005.

\bibitem{jaxnerf2020github}
Boyang Deng, Jonathan~T. Barron, and Pratul~P. Srinivasan.
\newblock {JaxNeRF}: an efficient {JAX} implementation of {NeRF}, 2020.
\newblock
  \url{http://github.com/google-research/google-research/tree/master/jaxnerf}.

\bibitem{evans2018measure}
Lawrence~C Evans and Ronald~F Garzepy.
\newblock {\em Measure theory and fine properties of functions}.
\newblock Routledge, 2018.

\bibitem{gu2021stylenerf}
Jiatao Gu, Lingjie Liu, Peng Wang, and Christian Theobalt.
\newblock Stylenerf: A style-based 3d-aware generator for high-resolution image
  synthesis, 2021.

\bibitem{hedman2017casual3d}
Peter Hedman, Suhib Alsisan, Richard Szeliski, and Johannes Kopf.
\newblock {Casual 3D Photography}.
\newblock {\em SIGGRAPH Asia}, 2017.

\bibitem{hedman2018deep}
Peter Hedman, Julien Philip, True Price, Jan-Michael Frahm, George Drettakis,
  and Gabriel Brostow.
\newblock Deep blending for free-viewpoint image-based rendering.
\newblock {\em SIGGRAPH Asia}, 2018.

\bibitem{hedman2021snerg}
Peter Hedman, Pratul~P. Srinivasan, Ben Mildenhall, Jonathan~T. Barron, and
  Paul Debevec.
\newblock Baking neural radiance fields for real-time view synthesis.
\newblock {\em ICCV}, 2021.

\bibitem{hinton2015distilling}
Geoffrey Hinton, Oriol Vinyals, and Jeff Dean.
\newblock Distilling the knowledge in a neural network.
\newblock {\em arXiv:1503.02531}, 2015.

\bibitem{jouppi2017datacenter}
Norman~P Jouppi, Cliff Young, Nishant Patil, David Patterson, Gaurav Agrawal,
  Raminder Bajwa, Sarah Bates, Suresh Bhatia, Nan Boden, Al Borchers, et~al.
\newblock In-datacenter performance analysis of a tensor processing unit.
\newblock {\em International Symposium on Computer Architecture}, 2017.

\bibitem{kalman1960new}
Rudolph~E. Kalman.
\newblock A new approach to linear filtering and prediction problems.
\newblock {\em Journal of Basic Engineering}, 1960.

\bibitem{kettunen2019lpips}
Markus Kettunen, Erik H{\"a}rk{\"o}nen, and Jaakko Lehtinen.
\newblock E-lpips: robust perceptual image similarity via random transformation
  ensembles.
\newblock {\em arXiv:1906.03973}, 2019.

\bibitem{Khademi2021}
Wesley Khademi and Jonathan Ventura.
\newblock View synthesis in casually captured scenes using a cylindrical neural
  radiance field with exposure compensation.
\newblock {\em ACM SIGGRAPH 2021 Posters}, 2021.

\bibitem{adam}
Diederik~P. Kingma and Jimmy Ba.
\newblock Adam: A method for stochastic optimization.
\newblock {\em ICLR}, 2015.

\bibitem{Knapitsch2017}
Arno Knapitsch, Jaesik Park, Qian-Yi Zhou, and Vladlen Koltun.
\newblock Tanks and temples: Benchmarking large-scale scene reconstruction.
\newblock {\em ACM Transactions on Graphics}, 36(4), 2017.

\bibitem{kopanas2021point}
Georgios Kopanas, Julien Philip, Thomas Leimk{\"u}hler, and George Drettakis.
\newblock Point-based neural rendering with per-view optimization.
\newblock {\em Computer Graphics Forum}, 2021.

\bibitem{lin2020deep}
Kai-En Lin, Zexiang Xu, Ben Mildenhall, Pratul~P Srinivasan, Yannick
  Hold-Geoffroy, Stephen DiVerdi, Qi Sun, Kalyan Sunkavalli, and Ravi
  Ramamoorthi.
\newblock Deep multi depth panoramas for view synthesis.
\newblock {\em ECCV}, 2020.

\bibitem{liu2020nsvf}
Lingjie Liu, Jiatao Gu, Kyaw~Zaw Lin, Tat-Seng Chua, and Christian Theobalt.
\newblock Neural sparse voxel fields.
\newblock {\em NeurIPS}, 2020.

\bibitem{maji2008classification}
Subhransu Maji, Alexander~C Berg, and Jitendra Malik.
\newblock Classification using intersection kernel support vector machines is
  efficient.
\newblock {\em CVPR}, 2008.

\bibitem{martinbrualla2020nerfw}
Ricardo Martin-Brualla, Noha Radwan, Mehdi S.~M. Sajjadi, Jonathan~T. Barron,
  Alexey Dosovitskiy, and Daniel Duckworth.
\newblock {NeRF in the Wild: Neural Radiance Fields for Unconstrained Photo
  Collections}.
\newblock {\em CVPR}, 2021.

\bibitem{max1995optical}
Nelson Max.
\newblock Optical models for direct volume rendering.
\newblock {\em IEEE TVCG}, 1995.

\bibitem{mildenhall2019llff}
Ben Mildenhall, Pratul~P. Srinivasan, Rodrigo Ortiz-Cayon, Nima~Khademi
  Kalantari, Ravi Ramamoorthi, Ren Ng, and Abhishek Kar.
\newblock {Local Light Field Fusion: Practical View Synthesis with Prescriptive
  Sampling Guidelines}.
\newblock {\em ACM Transactions on Graphics (TOG)}, 2019.

\bibitem{mildenhall2020}
Ben Mildenhall, Pratul~P. Srinivasan, Matthew Tancik, Jonathan~T. Barron, Ravi
  Ramamoorthi, and Ren Ng.
\newblock {NeRF}: Representing scenes as neural radiance fields for view
  synthesis.
\newblock {\em ECCV}, 2020.

\bibitem{donerf}
Thomas Neff, Pascal Stadlbauer, Mathias Parger, Andreas Kurz, Joerg~H. Mueller,
  Chakravarty R.~Alla Chaitanya, Anton~S. Kaplanyan, and Markus Steinberger.
\newblock {DONeRF: Towards Real-Time Rendering of Compact Neural Radiance
  Fields using Depth Oracle Networks}.
\newblock {\em Computer Graphics Forum}, 2021.

\bibitem{Oechsle2021ICCV}
Michael Oechsle, Songyou Peng, and Andreas Geiger.
\newblock Unisurf: Unifying neural implicit surfaces and radiance fields for
  multi-view reconstruction.
\newblock {\em ICCV}, 2021.

\bibitem{overbeck2018system}
Ryan~S Overbeck, Daniel Erickson, Daniel Evangelakos, Matt Pharr, and Paul
  Debevec.
\newblock A system for acquiring, processing, and rendering panoramic light
  field stills for virtual reality.
\newblock {\em ACM Transactions on Graphics}, 2018.

\bibitem{park2021nerfies}
Keunhong Park, Utkarsh Sinha, Jonathan~T. Barron, Sofien Bouaziz, Dan~B
  Goldman, Steven~M. Seitz, and Ricardo Martin-Brualla.
\newblock {Nerfies: Deformable Neural Radiance Fields}.
\newblock {\em ICCV}, 2021.

\bibitem{pele2010quadratic}
Ofir Pele and Michael Werman.
\newblock The quadratic-chi histogram distance family.
\newblock {\em ECCV}, 2010.

\bibitem{piala2021terminerf}
Martin Piala and Ronald Clark.
\newblock Terminerf: Ray termination prediction for efficient neural rendering.
\newblock {\em 3DV}, 2021.

\bibitem{reiser2021kilonerf}
Christian Reiser, Songyou Peng, Yiyi Liao, and Andreas Geiger.
\newblock Kilonerf: Speeding up neural radiance fields with thousands of tiny
  mlps.
\newblock {\em ICCV}, 2021.

\bibitem{riegler2021svs}
Gernot Riegler and Vladlen Koltun.
\newblock Stable view synthesis.
\newblock {\em CVPR}, 2021.

\bibitem{rubin1980bvh}
Steven~M. Rubin and Turner Whitted.
\newblock A 3-dimensional representation for fast rendering of complex scenes.
\newblock {\em SIGGRAPH}, 1980.

\bibitem{samet1990design}
Hanan Samet.
\newblock {\em The design and analysis of spatial data structures}.
\newblock Addison-Wesley, 1990.

\bibitem{Schlick}
Christophe Schlick.
\newblock Fast alternatives to perlin's bias and gain functions.
\newblock {\em Graphics Gems IV}, 1994.

\bibitem{schoenberger2016sfm}
Johannes~Lutz Sch\"{o}nberger and Jan-Michael Frahm.
\newblock Structure-from-motion revisited.
\newblock {\em CVPR}, 2016.

\bibitem{shum1999concentric}
Heung-Yeung Shum and Li-Wei He.
\newblock Rendering with concentric mosaics.
\newblock {\em SIGGRAPH}, 1999.

\bibitem{nerv2021}
Pratul~P. Srinivasan, Boyang Deng, Xiuming Zhang, Matthew Tancik, Ben
  Mildenhall, and Jonathan~T. Barron.
\newblock {NeRV}: Neural reflectance and visibility fields for relighting and
  view synthesis.
\newblock {\em CVPR}, 2021.

\bibitem{tancik2020fourfeat}
Matthew Tancik, Pratul~P. Srinivasan, Ben Mildenhall, Sara Fridovich-Keil,
  Nithin Raghavan, Utkarsh Singhal, Ravi Ramamoorthi, Jonathan~T. Barron, and
  Ren Ng.
\newblock Fourier features let networks learn high frequency functions in low
  dimensional domains.
\newblock {\em NeurIPS}, 2020.

\bibitem{wang2004image}
Zhou Wang, Alan~C Bovik, Hamid~R Sheikh, and Eero~P Simoncelli.
\newblock Image quality assessment: from error visibility to structural
  similarity.
\newblock {\em IEEE TIP}, 2004.

\bibitem{yu2021plenoctrees}
Alex Yu, Ruilong Li, Matthew Tancik, Hao Li, Ren Ng, and Angjoo Kanazawa.
\newblock {PlenOctrees} for real-time rendering of neural radiance fields.
\newblock {\em ICCV}, 2021.

\bibitem{kaizhang2020}
Kai Zhang, Gernot Riegler, Noah Snavely, and Vladlen Koltun.
\newblock {NeRF++: Analyzing and Improving Neural Radiance Fields}.
\newblock {\em arXiv:2010.07492}, 2020.

\bibitem{zhang2018unreasonable}
Richard Zhang, Phillip Isola, Alexei~A Efros, Eli Shechtman, and Oliver Wang.
\newblock The unreasonable effectiveness of deep features as a perceptual
  metric.
\newblock {\em CVPR}, 2018.

\bibitem{nerfactor}
Xiuming Zhang, Pratul~P. Srinivasan, Boyang Deng, Paul Debevec, William~T.
  Freeman, and Jonathan~T. Barron.
\newblock {NeRFactor: Neural Factorization of Shape and Reflectance Under an
  Unknown Illumination}.
\newblock {\em SIGGRAPH Asia}, 2021.

\bibitem{zheng2007layered}
Ke~Colin Zheng, Sing~Bing Kang, Michael~F Cohen, and Richard Szeliski.
\newblock Layered depth panoramas.
\newblock {\em CVPR}, 2007.

\end{thebibliography}
